\documentclass[journal]{IEEEtran}

\ifCLASSOPTIONcompsoc

  \usepackage[nocompress]{cite}
\else
  \usepackage{cite}
\fi

\ifCLASSINFOpdf

\else

\fi

\usepackage{bm}
\usepackage{booktabs}
\usepackage{multirow}
\usepackage{amssymb}
\usepackage{graphicx}
\usepackage{subfigure}
\usepackage{stfloats}
\usepackage{algorithm}  
\usepackage{algorithmic}
\usepackage{color}
\usepackage{hyperref}
\usepackage{url}

\begin{document}

\title{ROBY: Evaluating the Robustness of a Deep Model by its Decision Boundaries}
%
%

\author{Jinyin Chen,
        Zhen Wang,
        Haibin Zheng,
        Jun Xiao,
        Zhaoyan Ming
\thanks{J.~Chen is with the Institute of Cyberspace Security, College of Information Engineering, Zhejiang University of Technology, Hangzhou, 310023, China.}
\thanks{Z.~Wang is with the College of Software Technology at Zhejiang University, Hangzhou 310007, China.}
\thanks{H.~Zheng is with the College of Information Engineering at Zhejiang University of Technology, Hangzhou 310007, China.}
\thanks{J.~Xiao is with the College of Computer Science and Technology at Zhejiang University, Hangzhou 310007, China.}
\thanks{Z.~Ming is with the Institute of Computing Innovation, Zhejiang Univeristy, Hangzhou 310023, China.}
\thanks{
Corresponding author: Zhaoyan Ming, e-mail: mingzhaoyan@gmail.com}
}

\maketitle

\begin{abstract}
With the successful application of deep learning models in many real-world tasks, the model robustness becomes more and more critical. Often, we evaluate the robustness of the deep models by attacking them with purposely generated adversarial samples, which is computationally costly and dependent on the specific attackers and the model types. This work proposes a generic evaluation metric ROBY, a novel attack-independent robustness measure based on the model's decision boundaries. Independent of adversarial samples, ROBY uses the inter-class and intra-class statistics to capture the
features of the model’s decision boundaries. We experimented on ten state-of-the-art deep models and showed that ROBY matches the robustness gold standard of attack success rate (ASR) by a strong first-order generic attacker. With only 1\% of time cost. To the best of our knowledge, ROBY is the first lightweight attack-independent robustness evaluation metric that can be applied to a wide range of deep models. The code of ROBY is open-sourced at  
\href{https://github.com/baaaad/ROBY-Evaluating-the-Robustness-of-a-Deep-Model-by-its-Decision-Boundaries}{https://github.com/baaaad/ROBY-Evaluating-the-Robustness-of-a-Deep-Model-by-its-Decision-Boundaries}.
\end{abstract}


\IEEEpeerreviewmaketitle

\section{Introduction}

\IEEEPARstart{D}{eep} learning has solved many complex pattern recognition problems through neural networks in recent years \cite{kamilaris2018deep}. Correspondingly, a variety of deep learning models have emerged, which have achieved great success in data mining, natural language processing, multimedia learning, speech recognition, recommendation, and other related fields \cite{nguyen2019machine}, \cite{redmon2016you}, \cite{shi2018pairwise}. The progress has also promoted the application of deep learning models in the safety-critical area, including autonomous driving, healthcare, face recognition, and malware detection \cite{bojarski2016end}, \cite{Adrian2017A}.

At the same time, studies have proved that deep learning models are vulnerable to adversarial examples \cite{szegedy2013intriguing}. In related research work, Goodfellow et al. \cite{Goodfellow2015Explaining} proved that a simple gradient method could obtain adversarial perturbations and generate adversarial samples so that the deep learning model changes the output with a high degree of confidence.
Therefore, the robustness of neural networks against adversarial samples has received increasing attention, research of adversarial samples and the robustness of deep models have rapidly developed. In existing research, there’re two main directions: (i) robustness evaluation method for deep learning models \cite{kurakin2016adversarial}, \cite{hein2017formal}, and (ii) effective attack algorithm \cite{Goodfellow2015Explaining}, \cite{carlini2018ground}, \cite{8842604} and defense method \cite{Madry2017Towards}, \cite{8611298}, \cite{8698453}.

In this paper, we focus on the adversarial robustness evaluation of deep models. It has become a vital issue in deep learning and directly affects the application of deep models. However, existing robustness evaluation methods generally use different attack results to evaluate the robustness of deep models. That is, using the attack success rate as robustness metrics \cite{moosavi2016deepfool}. Consequently, the model robustness is entangled with the attack algorithms, leading to high computational cost, and the attack capabilities limit the robustness analysis. More importantly, the dependency between robustness evaluation and attack approaches can cause biased analysis. The existing certified robustness evaluation methods proposed various model certifiers and metrics \cite{weng2018evaluating}, \cite{gehr2018ai2:}. These methods mainly prove the model robustness by analyzing specific layers and activation functions of the network, which require detailed information about model structure. The certified methods usually have high computational cost, and the white-box property also limits their practicability.

The decision boundaries are essential constructs of deep learning classifiers \cite{spangher2018actionable}. Meanwhile, there have been robustness researches from the perspective of model decision boundaries \cite{tanay2016boundary}, \cite{he2018decision}, \cite{ding2019mma}. As pointed out, vulnerable samples tend to lie closer to the decision boundary \cite{cao2017mitigating}, \cite{kim2019bridging}, \cite{zha2019deep}. Recent studies showed that decision boundaries keep moving closer to natural samples across training, while adversarial training appeared to have the potential to prevent this kind of convergence \cite{mickisch2020understanding}. To a certain extent, it implies the connection between the model robustness and the distance between the samples and the decision boundary.

\begin{figure*}[ht]
\centering
\subfigure[Natural model with original samples]{
\includegraphics[width=0.3\textwidth]{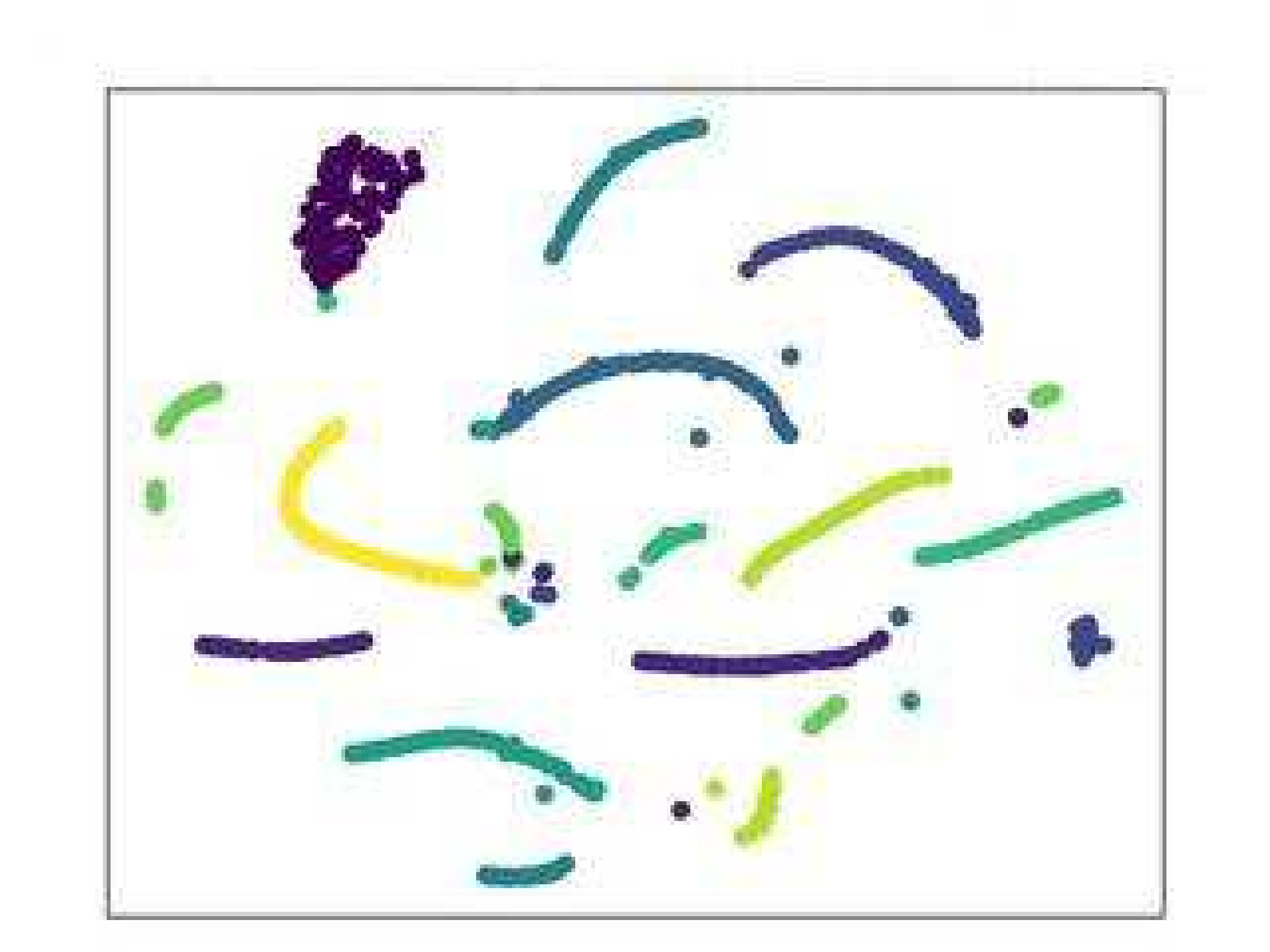} 
}
\subfigure[Natural model with adversarial samples]{
\includegraphics[width=0.3\textwidth]{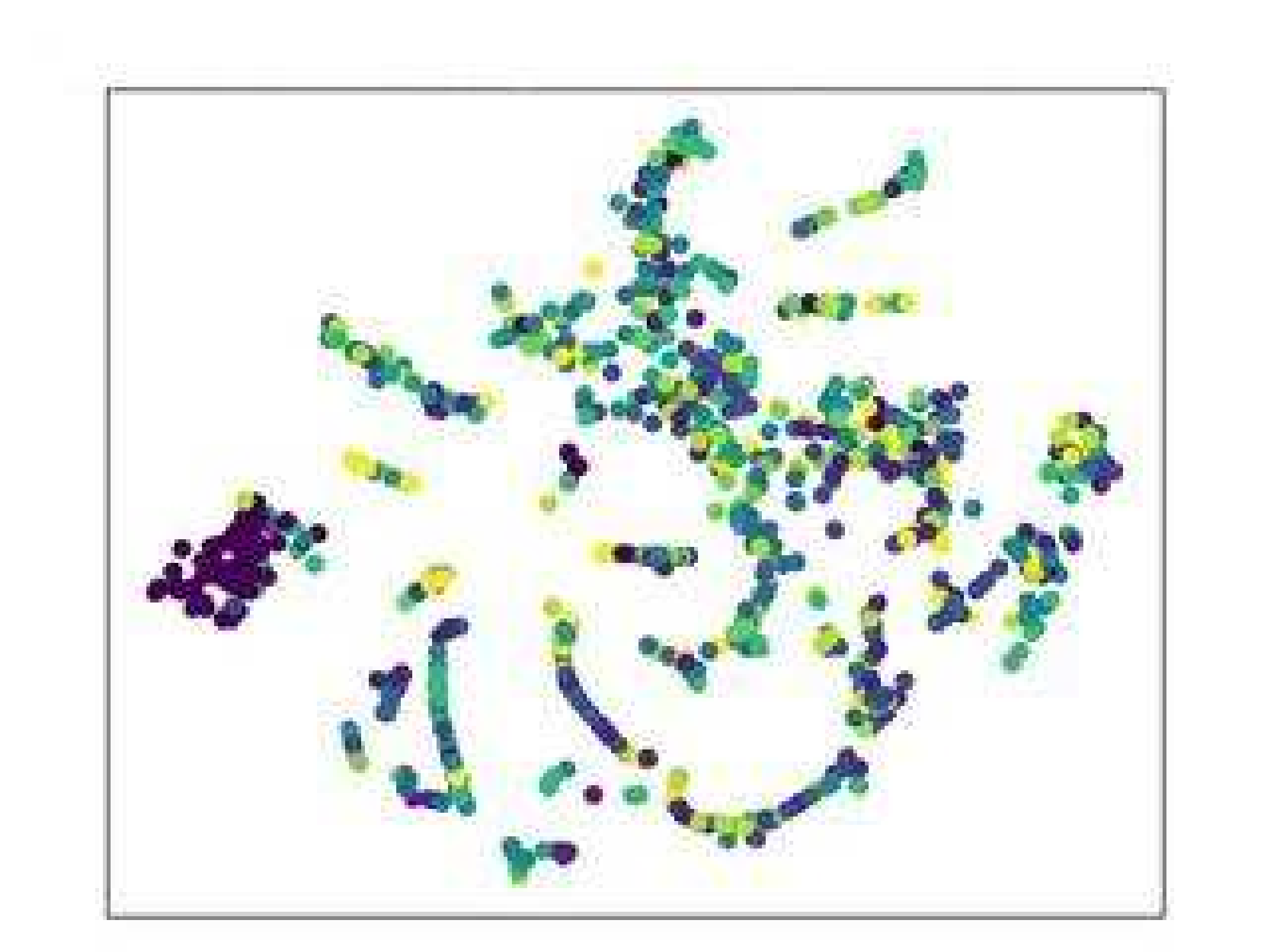}
}
\subfigure[Adversarial trained robust model with adversarial samples]{
\includegraphics[width=0.3\textwidth]{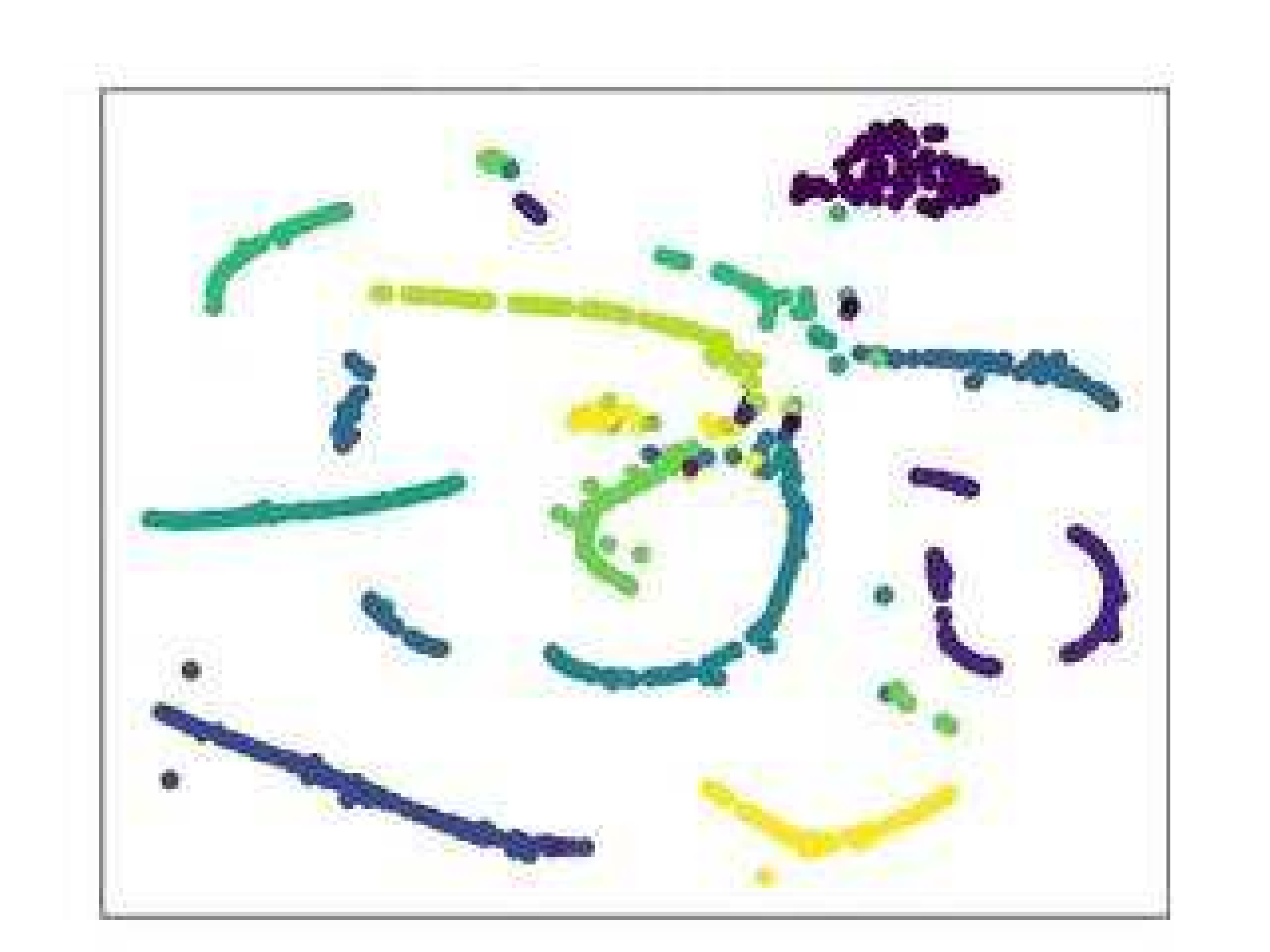}
}
\caption{Visualization of t-SNE embeddings of RestNet101 decision boundaries on MINIST. We used the PGD attack \cite{Madry2017Towards} for adversarial training and generating adversarial samples with perturbations of size $\varepsilon= 0.3$. The classification accuracy on original samples of the natural and robust model reaches 99\%, and 10000 test images are used for visualization.}
\label{tsne_intro}
\end{figure*}

As a concrete example, we use the ``t-SNE'' technique \cite{maaten2008visualizing} to visualize a RestNet101 model trained on the MNIST dataset.  
As illustrated by Fig.~\ref{tsne_intro}, we can see that the natural model gives proper grouping of classes and decision boundaries with normal samples. Samples belonging to the same class tend to locate closer, and the distance between different classes is large and clear. However, with adversarial samples, its robustness is compromised, where no clear class grouping and decision boundaries can be found visually. On the contrary, with an adversarial trained robust model, even the adversarial samples produce clear decision boundaries and grouping, demonstrating its robustness. 

Motivated by previous studies on the connection between adversarial training and decision boundaries, and the visualization of decision boundaries in models of different robustness levels, we propose ROBY, an adversarial robustness evaluation metric of deep models based on the deep learning model's decision boundary and feature space. It integrates the ROBY metrics about the model's decision boundary for robustness evaluation. 

Specifically, we quantitatively capture the decision boundary of a deep classifier by the distribution of the data in relation to the decision boundaries. It needs less in¬formation about model architecture and network composition than certified robustness evaluation methods. Compared to the attack-based robustness evaluation, it is computationally more efficient.
Given that any model can compute its ROBY metric once it's trained, ROBY is not dependent on adversarial samples, sample generating algorithms, nor attack algorithms. Our experiments showed that ROBY could achieve similar performance as ASR on adversarial robustness evaluation on many state-of-the-art deep models (DenseNet, MobileNetV, MobileNetV2, ResNet50, ResNet101, InceptionV1, InceptionV2, AlexNet, SqueezeNet, LeNet) and deep models of different complexity and capacity.

We have made the following contributions in this paper:

\begin{itemize}
\item We proposed ROBY, a generic model robustness evaluation metric based on the decision boundaries. It is efficient, lightweight, attack-independent, and adversarial sample free.

\item ROBY achieves a strong positive correlation with attack success rate, the most adopted robustness evaluation metric.  

\item We applied ROBY on ten state-of-the-art neural network models, models with different complexity, and networks defended against projected gradient descent (PGD) \cite{Madry2017Towards} adversarial samples to verify its practicability and effectiveness. Our study is the first robustness metric that is generic across deep models to the best of our knowledge.
\end{itemize}

The rest of the paper is organized as follows. We compare and analyze existing robustness evaluation methods and prior researches on decision boundaries and adversarial samples in RELATED WORK. The following section, we describe our metric ROBY: Robustness Evaluation based on Decision Boundaries. In EXPERIMENTS, we demonstrate the effectiveness and efficiency of ROBY on image classification tasks.

\section{RELATED WORK}

We studied the problem of evaluating the robustness of a deep network against adversarial attacks and discussed prior works on the same. In particular, we reviewed algorithms that explored the decision boundaries and adversarial samples for the inspiration they bring.  

\subsection{Robustness Evaluation of Deep Models}

The general robustness evaluation method measures the upper bound on the minimum adversarial distance demonstrated through adversarial examples created by adversarial attacks \cite{Goodfellow2015Explaining}, \cite{kurakin2016adversarial}. Generally, the easier it is to build an adversarial sample on a deep model, and the less perturbation added, the less robust the model is \cite{moosavi2016deepfool}. Using the attack algorithm to attack deep models, a robust model tends to have a low attack success rate. Such an attack-based evaluation is generally conducted on state-of-the-art attack algorithms  \cite{carlini2017towards}, \cite{Madry2017Towards}. However, this kind of robustness evaluation method still has a high computational cost and biased evaluation conclusions.

In addition to adversarial attacks, the deep model's robustness can also be evaluated based on the deep model's structure and parameters. The work by Wang et al. \cite{wang2016theoretical}  of the University of Virginia used topological terms to study the robustness of deep models but did not provide robustness boundaries or estimates for neural networks. Gopinath \cite{gopinath2018deepsafe} of Carnegie Mellon University and his team proposed DeepSafe, an automatic, data-driven approach. DeepSafe clustered the known labeled data and used the existing constraint solver to automatically identify the network's robust area where all inputs in this area were correctly classified.

Besides, certified robustness evaluation methods measure provable lower bound on the minimum adversarial distance necessary to cause misclassification, mainly through well-designed metrics or model certifiers. The research team led by Google artificial intelligence researcher Christian Szegedy \cite{szegedy2013intriguing} calculated the global Lipschitz constant of each layer of the neural network and uses their product to explain the robustness of the deep model, but the global Lipschitz constant usually corresponds to loose bound.  Hein et al. \cite{hein2017formal} at the University of Tübingen used local Lipschitz continuity conditions to give a lower bound on robustness and derived a closed-form bound for a multilayer perceptron (MLP) with a single hidden layer and soft plus activation. However, for neural networks with multiple hidden layers, it was difficult to derive closed-form bounds. MIT's Weng \cite{weng2018evaluating} and his team proposed a new type of robustness metric called CLEVER (Cross Lipschitz Extreme Value for network Robustness), which provided a theoretical basis for transforming the robustness analysis into a local Lipschitz constant estimation problem. The CLEVER indicator could be applied to any deep neural network classifier, which used the extreme value theory to estimate the scale of the minimum perturbation required to generate effective adversarial samples.

Katz et al. \cite{katz2017reluplex} of Stanford University proposed the Reluplex analyzer, a new algorithm for neural network error detection based on model verification.  
Reluplex evaluated the model robustness by measuring the minimum threshold or adversarial signal that can produce false results. The algorithm allowed the testing of neural networks of orders of magnitude larger than before, for example, a fully connected neural network with eight layers and 300 nodes. Based on Reluplex, Katz et al.\cite{katz2017reluplex} defined local adversarial robustness, global adversarial robustness, and mixed definition metrics in their subsequent works \cite{katz2017towards} on the robustness evaluation of deep models. Huang et al. \cite{huang2017safety}  of Oxford University developed a novel automatic verification framework for feed-forward multilayer neural networks based on satisfaction model theory (SMT) to prove the local adversarial robustness of neural networks. 

The end-to-end analyzer AI2 proposed by Gehr et al. \cite{gehr2018ai2:} of the Federal Institute of Technology in Zurich can automatically prove the robustness of the neural network by analyzing the behavior of the maximum connection layer and the convolutional neural network layer with the ReLU activation function and the maximum pooling layer.  AI2 was much faster than existing symbol-based analyzers. It could evaluate large networks' robustness with more than 1200 neurons in minutes, while Reluplex required hours on the same scale. Guo et al. \cite{guo2020coverage} proposed DLFuzz, a coverage-guided differential adversarial testing framework to expose deep learning systems' possible incorrect behaviors. DLFuzz performed better than the state-of-the-art testing framework DeepXplore \cite{pei2017deepxplore},  the first white-box framework for systematically testing real-world deep learning systems.

\subsection{Decision Boundary and Adversarial Examples}

Regarding the vulnerability of trained neural networks towards adversarial samples, there are studies about decision boundaries and adversarial samples. As one of the essential components of a classifier, researchers have highlighted the importance of the decision boundary and related properties \cite{vemuri1988artificial}, \cite{spangher2018actionable}. A particular case is in a classification problem; the deep learning model's decision boundary reflects the model's fitting situation and generalization ability to the data. Cortes and Vapnik \cite{cortes1995support} proved that samples near the decision boundary of a classifier have a more considerable impact on the performance than those far away from it. 

Meanwhile, consistent success of strong targeted adversarial attacks like PGD \cite{Madry2017Towards} and C\&W \cite{carlini2017towards} suggested that an adversary can create any desired classification output by adding a suitable adversarial perturbation to the natural image. Tanay and Griffin \cite{tanay2016boundary} argued that adversarial examples exist when the classification boundary lies close to the sub-manifold of sampled data, but their analysis was limited to linear classifiers. Cao and Gong \cite{cao2017mitigating} found that an adversarial sample was located near the decision boundary and used this property to defense against adversarial attacks. He et al. \cite{he2018decision} introduced a technique to examine the properties of decision boundaries around an input instance, namely the distances to the boundaries and the adjacent classes, which helps to investigate the effectiveness of adversarial attacks. Shamir et al. \cite{shamir2019simple} explained the adversarial examples via the partitions' geometric structure defined by decision boundaries. However, their work was limited to linear decision boundaries, and they did not consider the actual distance to the decision boundaries.

Typically, neural networks' decision boundaries were mostly investigated through simplifying assumptions or approximation methods, leading to unreliable results. Recently, more numerical researches were carried out. Yousefzadeh et al. \cite{yousefzadeh2019investigating} computed exact points called flip points on the decision boundaries of deep models and provide mathematical tools to investigate the surfaces that define the decision boundaries, which show better insight into models robustness and weakness against adversarial attack. Mickisch et al. \cite{mickisch2020understanding} studied the minimum distance of data points to the decision boundary and how this margin evolves over the training of a deep neural network. Experimental results showed that the decision boundary keeps moving closer to natural images over the training. In contrast, adversarial training appeared to have the potential to prevent this convergence of the decision boundary. However, this phenomenon lacks explanation and further study.

\subsection{Robustness based on Adversarial Attack and Defense}

The adversarial attack is a crucial way to measure the robustness of a machine learning model; thereby, the way the adversarial samples are generated is critical. Image classification models are typically gauged in terms of \emph{p-norm} distortions in the pixel feature space. Adversarial samples are generally generated by restricting $l_0$, $l_2$, and $l_\infty$ norms of the perturbations to make them indistinguishable by humans. This paper mainly focuses on the model robustness against $l_2$ and $l_\infty$ attacks, instead of $l_0$ attacks, given the discussion below.

The $l_0$ norm attacks find the smallest number of pixels that need to be changed to fool the classifier. There are currently gradient-based white-box attacks \cite{papernot2016limitations}, \cite{carlini2017towards}, \cite{modas2019sparsefool}, and mainly black-box attacks use either local search or evolutionary algorithms \cite{narodytska2016simple}, \cite{schott2019towards}, \cite{croce2019sparse}, \cite{su2019one}. The $l_0$ norm attacks have a general application limitation of model type and dataset scale, leads to limited research on this topic.

Meanwhile, $l_2$ norm and $l_\infty$ norm attacks gain more research interests. The $l_\infty$ norm attacks pay attention to the maximum absolute value of perturbations. Typically, Ian Goodfellow et al. \cite{Goodfellow2015Explaining} proposed the Fast Gradient Sign Method (FGSM). More works in this line include  BIM \cite{dong2018boosting}, the momentum-based iterative algorithm MI-FGS \cite{dong2018boosting}, the distributed adversarial attack (DAA) \cite{zheng2019distributionally}, and a transfer-based targeted adversarial attack \cite{inkawhich2019feature}.

The $l_2$ norm attack measures the standard Euclidean distance of perturbations \cite{szegedy2013intriguing}. As there have been methods able to generate $l_\infty$, $l_2$, and $l_0$ attacks together or their combinations. The $l_2$ norm attack usually come together with $l_\infty$ attack \cite{moosavi2016deepfool}, \cite{carlini2017towards}, \cite{chen2017zoo}, \cite{Madry2017Towards}.

To improve the robustness of deep neural networks against adversarial attacks, existing defense methods are developed along with three main directions: modifying network parameters or structures \cite{papernot2016distillation}, \cite{dhillon2018stochastic}, \cite{sun2019adversarial}, \cite{mustafa2019adversarial}, \cite{8758400}, \cite{7407373}; adding additional plugins to the model and using external models \cite{samangouei2018defense}, \cite{liao2018defense}, \cite{10.1145/3394486.3403241}, \cite{251574}, \cite{goel2020dndnet}, \cite{7533509}; modifying inputs for training or testing.

 Studies also show that adversarial training is one of the most effective defenses against adversarial attacks \cite{papernot2016technical}, \cite{carlini2018ground}, \cite{xie2019feature}. Goodfellow et al. \cite{Goodfellow2015Explaining} and Huang et al. \cite{huang2015learning} first proposed adversarial training by adding FGSM generated adversarial samples into the training set. Further studies deepen into adversarial training with stronger adversarial attacks \cite{Madry2017Towards}, \cite{kannan2018adversarial}, as well as providing robustness guarantee for multiple perturbation attack \cite{tramer2019adversarial} and optimizing the computation and time cost of adversarial training \cite{shafahi2019adversarial}, \cite{zhang2019you}.
 
This paper pays attention to the robustness change of deep models after adversarial training and whether ROBY can measure it.

\section{ROBY: Robustness Evaluation based on Decision Boundaries}
\subsection{Preliminaries}
The robustness of a deep classifier's ability to produce accurate results is generally not directly reflected by the structure of the model, such as the number of hidden layers of the neural network. An exception is in a classification problem; the deep learning model's decision boundary reflects the model's fitting situation and generalization ability to the data. For a deep learning model based on a backpropagation neural network, the model structure determines the type (linear or non-linear) of the decision boundary that the model can learn.

For the same data set, different deep learning models learn different decision boundaries. For adversarial samples, models require a significantly more complicated decision boundary. It is difficult to directly define the decision boundary numerically and explain it theoretically. However, we find that the data distribution results obtained by different decision boundaries in the feature space reflect the differences in the model's mapping of data features, which to a certain extent show the differences of decision boundaries. Thus, in this work, instead of calculating the actual distance between the samples and the decision boundaries, we capture the decision boundary of a deep model by measuring the distribution of the data relating to the decision boundaries, in particular the inter-class and intra-class statistic features.

The decision boundary is defined in the embedding space. Consisting of the major component of a deep classifier except the output layer, the embedding function $f:\mathbb{R}^{N}\rightarrow \mathbb{R}^{M}$, maps the $N$-dimensions input $x$ into the $M$-dimension embedding space representation $e$. 
Generally, for a K-class  $y\in \left \{ 1,...,K\right \}$ classification task $c(f(x))\rightarrow y$, the classifier's output $y$ can be obtained by the softmax value of its M-dimensional embedding result $f(x):(e_{1},...,e_{M})$. Thus, we have the classification result as:

\begin{equation}
\label{eqn_y}
y =\mathop{\arg\max}_{k}\left ( \frac{\exp( c_k(f(x)))}{\sum_{k=1}^{K}\exp(c_k(f(x)))}\right )
\end{equation}

\textbf{Adversarial Robustness}. As for the adversarial attack, with well-designed perturbations, the adversarial sample $x^{adv}$ can easily fool the classifier to make a wrong decision from  $c(f(x))\rightarrow y$ to $c(f(x^{adv}))\rightarrow {y}'$. Here we give a general definition of adversarial robustness for deep classifiers against adversarial attacks. For a set of allowed perturbations $S\in \mathbb{R}^{N}$, the deep classifier which under adversarial attack keeps the decision result as original input denoted as:

\begin{equation}
\label{eqn_robustness}
c(f(x+S))\rightarrow y, \quad  while  \ c(f(x))\rightarrow y
\end{equation}

 \textbf{Feature Vector}. For the input $x$, we use its M-dimensional embedding $f(x):(e_{1},...,e_{M})$ to represent it in the feature space. Suppose the classifier gives the decision result as $y= k\in\mathbb{R}^{K}$; we denote this sample's feature vector as $f_{x,k}$. For the set of examples belonging to class $k$ denote as $N_k$, we calculate the center of this class as the mean vector of the embedded samples in this class, denote as

\begin{equation}
\label{eqn_center}
f_{c_{k}} = \frac{1}{\left | N_k\right |} \sum_{x_i \in N_k} f_{x_i,k}
\end{equation}

\textbf{Distance Metric}.
We use the Minkowski Distance \cite{van1995some} to measure the distance between a pair of samples represented as two M-dimensional vectors $f_{x_{1}}:(e_{11},...,e_{1M})$ and $f_{x_{2}}:(e_{21},...,e_{2M})$ in the feature space, written as:

\begin{equation}
\label{eqn_dist}
dist(f_{x_{1}},f_{x_{2}})=\left ( \sum_{a}^{M}\left | e_{1a}-e_{2a}\right |^{p}\right )^{1/p}
\end{equation}

\subsection{The Proposed ROBY Metrics for Robustness  Evaluation}

Specifically, we propose to evaluate the model's robustness by integrating the statistic features based on decision boundaries (ROBY) from two aspects: the feature subspace aggregation (FSA) of the same class, and the feature subspace distance (FSD) of different classes. 

The feature subspace aggregation (FSA) depicts the compactness of data in the same class. The smaller the distance within the same class of samples in the feature space and the center of the same class's feature, the higher the data aggregation degree, and the model is more robust.

\begin{equation}
\label{eqn_FSAK}
FSA_{k}= 1-\frac{norm(\sum_{j=1}^{n_{k}}dist(f_{x_{j},k},f_{c_{k}}))}{n_{k}}
\end{equation}

\begin{equation}
\label{eqn_FSA}
FSA =  \frac{\sum_{i=1}^{K}FSA_{i}}{K}
\end{equation}

\noindent where $n_{k}$ represents the number of samples of the $k^{th}$ class in the data set. $norm(\cdot)$ denotes the standardized function.  $f_{x_{j},k}$ represents the feature vector of the sample $x_{j}$ belonging to the $k^{th}$ class in the high-dimensional feature space. An $f_{c_{k}}$ represents the $k^{th}$ class center. $K$ denotes the number of classes. We calculate the final FSA by averaging the sum of all $FSA_{k}$ belongs to $K$ classes.

The feature subspace distance (FSD) depicts the distinction of different classes: the more significant the average distance between all classes, the more robust the model.

\begin{equation}
\label{eqn_FSDK}
FSD_{k,k+1}= dist(f_{c_{k}},f_{c_{k+1}})
\end{equation}

\begin{equation}
\label{eqn_FSD}
FSD= \frac{ \sum_{i=1}^{K-1}\sum_{j=i+1}^{K}FSD_{i,j}}{K(K-1)/2}
\end{equation}

\noindent where $f_{c_{k}}$ and $f_{c_{k+1}}$ denote the center of the feature subspace of the $k^{th}$ and $k+1^{th}$ data. We calculate the final FSD by averaging the sum of all $FSD_{i,j}$ between each class.

ROBY integrates FSA and FSD into a single metric. The smaller the ROBY value is, the lower the overlap of different feature subspaces, the greater the decision boundary distance, and the more robust of the deep model. 

\begin{equation}
\label{eqn_ROBYK}
ROBY_{k,k+1}=FSA_{k}+FSA_{k+1}- FSD_{k,k+1}
\end{equation}

\begin{equation}
\label{eqn_ROBY}
ROBY= \frac{ \sum_{i=1}^{K-1}\sum_{j=i+1}^{K}ROBY_{i,j}}{K(K-1)/2}
\end{equation}

\noindent where $FSA_{k}$ and $FSA_{k+1}$ represent the feature subspace aggregation of the $k^{th}$ and $k+1^{th}$ data, and $FSD_{k,k+1}$ represents the feature subspace distance of the $k^{th}$ and $k+1^{th}$ data. Finally, we calculate ROBY by averaging over all $ROBY_{i,j}$ between classes.

Since we used the $l_p$-norm distance in our metric calculation, the corresponding ROBY metrics can have $l_p$ form due to the specific norm choice. And the ROBY metrics have a time complexity of $O(n^{2})$. We summarize the flow of computing feature statistical metrics ROBY, FSA, FSD in Algorithm 1.

\begin{algorithm}[!h]
\caption{Compute the robustness metrics of FSA, FSD, ROBY}
\textbf{Input:} Samples with $K$ classes and their feature vector $f$, \\ 
\leftline{\textbf{Output:} FSA, FSD, ROBY value. }
	\begin{algorithmic}[1]
		\STATE $FSA\_list \gets \{{\emptyset}\}$, $center\_list \gets \{{\emptyset}\}$, $ROBY\_list \gets \{{\emptyset}\}$
		\FOR{$k \gets 1$ to $K$}
		\FOR{$i \gets 1$ to $n_{k}$}
		\STATE  $f_{c_{k}} \gets f_{c_{k}} + f_{x_{j},k}$
		\ENDFOR
		\STATE $f_{c_{k}} \gets f_{c_{k}} / n_{k}$
		\STATE $center\_list \gets center\_list \bigcup f_{c_{k}}$
		\ENDFOR
		\FOR{$k \gets 1$ to $K$}
		\FOR{$i \gets 1$ to $n_{k}$}
		\STATE  $d_{k} \gets d_{k} + dist(f_{x_{j},k},f_{c_{k}})$
		\ENDFOR
		\STATE $FSA_{k} \gets d_{k}/n_{k}$
		\STATE $FSA\_list \gets FSA\_list \bigcup FSA_{k}$
		\ENDFOR
		\STATE $FSA \gets 1-avg(norm(FSA\_list))$
		\FOR{$i \gets 1$ to $K-1$}
		\FOR{$j \gets i+1$ to $K$}
		\STATE  $d_{i,j} \gets  dist(f_{c_{i}},f_{c_{j}})$
		\STATE  $d \gets  d+d_{i,j}$
		\ENDFOR
		\ENDFOR
		\STATE $FSD \gets norm(d/(K*(K-1)/2))$
		\FOR{$i \gets 1$ to $K$}
		\FOR{$j \gets i+1$ to $K$}
		\STATE  $ROBY_{i,j} \gets  FSA_{i}+FSA_{j}- dist(f_{c_{i}},f_{c_{j}})$
		\STATE  $ROBY\_list \gets ROBY\_list \bigcup ROBY_{i,j}$
		\ENDFOR
		\ENDFOR
		\STATE $ROBY \gets avg(norm(ROBY\_list))$
	\end{algorithmic}
\end{algorithm}

\subsection{The Key Properties of ROBY}
With ROBY's implementation algorithm above, we highlight ROBY's key properties are as follows.

 \textbf{Property 1: The smaller ROBY is, the more robust is the model. } According to Equation ~\ref{eqn_ROBYK} and~\ref{eqn_ROBY}, a smaller ROBY generally represents a higher data aggregation degree of each class; and a greater distance between different classes thus leads to smaller overlapping among the classes. Correspondingly, each sample generally locates farther from the decision boundary and other classes, which requires more substantial perturbations to generate adversarial samples.

\textbf{Property 2: ROBY can be calculated with natural samples without the need of generating adversarial samples.} We calculate the data distribution, using the M-dimensional embeddings of each input as their representation in the feature space, which is attack-independent.

\textbf{Property 3: ROBY is model-agnostic, applicable to natural models, defensive models, and models with varying complexity. }
According to Equation ~\ref{eqn_y}, a deep classifier can generally be represented as an embedding function followed by a softmax function. The embedding output and classification result of the model are all we need to calculate ROBY, which is easy to obtain and model-agnostic.

\section{Experiments}
This section tests the robustness evaluation metrics on several deep models under different settings and model capacity. The main research questions are:

\noindent \textbf{Q1}: How accurate can ROBY quantify the model robustness measured by attack success rate?

\noindent \textbf{Q2}: Can ROBY evaluate robust models of defended networks against adversarial samples?

\noindent \textbf{Q3}: How accurate can ROBY quantify the robustness of models with different complexity and model capacity?

\subsection{Datasets and Setups}

We conduct experiments on CIFAR-10, MNIST, and Fashion-MNIST datasets. We evaluate the robustness metrics on state-of-the-art neural network models, including DenseNet, MobileNetV, MobileNetV2, ResNet50, ResNet101, InceptionV1, InceptionV2, AlexNet, SqueezeNet, LeNet, simple fully connected network and convolutional network.  We use the same training set of each dataset for training and use 10000 test images to calculate the ROBY metrics.

All neural network models converge after training as natural models. The ROBY metric and the attack success rate are calculated under the same settings of the models.

We apply both $l_2$ and $l_\infty$ norm PGD attacks to generate adversarial samples for attack success rate. In the case of $l_\infty$-bounded PGD, we take gradient steps in the $l_\infty$ norm, adding the gradient sign. In the case of $l_2$-bounded PGD, we take steps in the gradient direction instead of its sign. For MNIST and Fashion-MNIST, we run 40 iterations of PGD to generate the adversary samples, with a step size of 0.01 and perturbations of size $\varepsilon= 0.3$. For the CIFAR10 data set, we generate PGD adversaries with step size 2 and perturbation size $\varepsilon= 1.0$. 
 
\subsection{Choice of Robustness Gold Standard}

We choose the attack success rate (ASR) as the robustness gold standard. Following the trend in the research community, we pay attention to the $l_2$ norm and $l_\infty$ norm attacks during the bench-marking. The $l_\infty$ norm attacks optimize the maximum absolute value of perturbations, while the $l_2$ norm attacks measure the standard Euclidean distance of perturbations \cite{szegedy2013intriguing}. Specifically, we apply state-of-the-art first-order attack methods, the sign-based PGD, to find adversarial examples for networks
 and calculate the attack success rate.

We applied the min-max robust optimization-based adversarial training \cite{Madry2017Towards} as a defense method, which has been widely used to improving adversarial robustness in model compression tasks \cite{ye2019adversarial}, \cite{gui2019model}. 

We use the same number of original samples and adversarial samples for $l_\infty$ adversarial training, with the adversarial samples  generated under the same settings above.

All neural network models converged after adversarial training are used as robust models. We then calculate ROBY metrics on these robust models to investigate defended networks against adversarial examples.

\subsection{Benchmark ROBY Measured Robustness on Deep Models}
In this subsection, we conduct experiments to answer \textbf{Q1,} the critical question of our study: how accurate can ROBY quantify the model robustness measured by attack success rate?

According to Equation ~\ref{eqn_dist}, ROBY can have two forms $l_\infty$, and $l_2 $. For the comprehensive evaluation of ROBY's consistency with the attack success rate, we compute the two forms of ROBY metrics and the $l_\infty$ and $l_2 $ norm PGD attack success rate of each natural model. We then investigate the effectiveness of ROBY metrics by comparing them with the attack success rate. TABLE~\ref{effectiveness} compares the attack success rate of PGD and the ROBY metrics.

\begin{table*}[htbp]
  \centering
  \caption{Robustness evaluation based on ASR and the proposed methods}
  \setlength{\tabcolsep}{2mm}{
  \scriptsize 
    \begin{tabular}{ccccccccccc}
    \hline\hline
    \textbf{Dataset} & \textbf{Model} & \textbf{ACC} & \textbf{$ASR_\infty$} & \textbf{$FSA_\infty$} & \textbf{$FSD_\infty$} & \textbf{$ROBY_\infty$} & \textbf{$ASR_2$} & \textbf{$FSA_2$} & \textbf{$FSD_2$} & \textbf{$ROBY_2$} \\
    \hline\hline
      & \textbf{ResNet101} & \textbf{0.7548 } & \textbf{0.3854 } & \textbf{0.8327 } & \textbf{0.7435 } & \textbf{0.2876 } & \textbf{0.8792 } & \textbf{0.8422 } & \textbf{0.7614 } & \textbf{0.2088 } \\
      & \textbf{DenseNet} & \textbf{0.6928 } & \textbf{0.4301 } & \textbf{0.7294 } & \textbf{0.6841 } & \textbf{0.3152 } & \textbf{0.8947 } & \textbf{0.7462 } & \textbf{0.6729 } & \textbf{0.3148 } \\
      & \textbf{MobileNetV2} & \textbf{0.7039 } & \textbf{0.4961 } & \textbf{0.6480 } & \textbf{0.6342 } & \textbf{0.3379 } & \textbf{0.9189 } & \textbf{0.7248 } & \textbf{0.6670 } & \textbf{0.3480 } \\
    \textbf{CIFAR-10} & \textbf{MobileNetV1} & \textbf{0.6913 } & \textbf{0.5497 } & \textbf{0.6301 } & \textbf{0.5739 } & \textbf{0.3509 } & \textbf{0.9205 } & \textbf{0.6621 } & \textbf{0.6052 } & \textbf{0.3796 } \\
      & \textbf{ResNet50} & \textbf{0.7015 } & \textbf{0.5504 } & \textbf{0.6155 } & \textbf{0.5124 } & \textbf{0.3697 } & \textbf{0.9362 } & \textbf{0.6548 } & \textbf{0.5431 } & \textbf{0.4138 } \\
      & \textbf{Inception V2} & \textbf{0.7485 } & \textbf{0.5508 } & \textbf{0.6129 } & \textbf{0.5002 } & \textbf{0.3889 } & \textbf{0.9410 } & \textbf{0.5436 } & \textbf{0.5081 } & \textbf{0.4257 } \\
      & \textbf{AlexNet} & \textbf{0.7366 } & \textbf{0.5590 } & \textbf{0.5483 } & \textbf{0.4752 } & \textbf{0.3952 } & \textbf{0.9518 } & \textbf{0.5192 } & \textbf{0.5011 } & \textbf{0.4853 } \\
      & \textbf{LeNet} & \textbf{0.6809 } & \textbf{0.6216 } & \textbf{0.5051 } & \textbf{0.4017 } & \textbf{0.4176 } & \textbf{0.9663 } & \textbf{0.4799 } & \textbf{0.4728 } & \textbf{0.5171 } \\
      & \textbf{Inception V1} & \textbf{0.6918 } & \textbf{0.6793 } & \textbf{0.4990 } & \textbf{0.3688 } & \textbf{0.4307 } & \textbf{0.9774 } & \textbf{0.4558 } & \textbf{0.4427 } & \textbf{0.5516 } \\
      & \textbf{SqueezeNet} & \textbf{0.6923 } & \textbf{0.8930 } & \textbf{0.4902 } & \textbf{0.3619 } & \textbf{0.4431 } & \textbf{0.9814 } & \textbf{0.4362 } & \textbf{0.3097 } & \textbf{0.6274 } \\
    \midrule
      & \textbf{MobileNetV2} & \textbf{0.9936 } & \textbf{0.6559 } & \textbf{0.8553 } & \textbf{0.6222 } & \textbf{0.2557 } & \textbf{0.8912 } & \textbf{0.8674 } & \textbf{0.7382 } & \textbf{0.2948 } \\
      & \textbf{MobileNetV1} & \textbf{0.9802 } & \textbf{0.7843 } & \textbf{0.7816 } & \textbf{0.6014 } & \textbf{0.3899 } & \textbf{0.9003 } & \textbf{0.8542 } & \textbf{0.5584 } & \textbf{0.4062 } \\
      & \textbf{ResNet101} & \textbf{0.9931 } & \textbf{0.8221 } & \textbf{0.6706 } & \textbf{0.5754 } & \textbf{0.4002 } & \textbf{0.9137 } & \textbf{0.7452 } & \textbf{0.4680 } & \textbf{0.4525 } \\
    \textbf{MNIST} & \textbf{LeNet} & \textbf{0.9892 } & \textbf{0.8540 } & \textbf{0.5946 } & \textbf{0.5493 } & \textbf{0.4304 } & \textbf{0.9245 } & \textbf{0.6832 } & \textbf{0.4538 } & \textbf{0.5237 } \\
      & \textbf{DenseNet} & \textbf{0.9906 } & \textbf{0.8794 } & \textbf{0.5411 } & \textbf{0.5207 } & \textbf{0.4859 } & \textbf{0.9371 } & \textbf{0.6101 } & \textbf{0.4205 } & \textbf{0.5671 } \\
      & \textbf{ResNet50} & \textbf{0.9842 } & \textbf{0.8996 } & \textbf{0.4737 } & \textbf{0.4865 } & \textbf{0.5006 } & \textbf{0.9459 } & \textbf{0.5881 } & \textbf{0.4169 } & \textbf{0.5986 } \\
      & \textbf{AlexNet} & \textbf{0.9880 } & \textbf{0.9217 } & \textbf{0.4703 } & \textbf{0.4486 } & \textbf{0.5512 } & \textbf{0.9586 } & \textbf{0.5623 } & \textbf{0.4117 } & \textbf{0.6054 } \\
      & \textbf{SqueezeNet} & \textbf{0.9868 } & \textbf{0.9254 } & \textbf{0.3879 } & \textbf{0.4133 } & \textbf{0.5705 } & \textbf{0.9603 } & \textbf{0.4523 } & \textbf{0.3559 } & \textbf{0.6097 } \\
      & \textbf{Inception V2} & \textbf{0.9910 } & \textbf{0.9377 } & \textbf{0.2573 } & \textbf{0.3835 } & \textbf{0.6257 } & \textbf{0.9662 } & \textbf{0.4509 } & \textbf{0.2736 } & \textbf{0.6485 } \\
      & \textbf{Inception V1} & \textbf{0.9844 } & \textbf{0.9859 } & \textbf{0.2217 } & \textbf{0.3679 } & \textbf{0.7209 } & \textbf{0.9722 } & \textbf{0.4257 } & \textbf{0.2524 } & \textbf{0.7284 } \\
    \midrule
      & \textbf{MobileNetV1} & \textbf{0.9068 } & \textbf{0.7985 } & \textbf{0.8291 } & \textbf{0.8198 } & \textbf{0.2060 } & \textbf{0.9063 } & \textbf{0.8394 } & \textbf{0.7100 } & \textbf{0.1543 } \\
      & \textbf{MobileNetV2} & \textbf{0.9137 } & \textbf{0.8071 } & \textbf{0.8113 } & \textbf{0.8004 } & \textbf{0.2573 } & \textbf{0.9124 } & \textbf{0.7872 } & \textbf{0.5457 } & \textbf{0.2721 } \\
      & \textbf{DenseNet} & \textbf{0.8950 } & \textbf{0.8486 } & \textbf{0.7757 } & \textbf{0.6819 } & \textbf{0.2954 } & \textbf{0.9397 } & \textbf{0.7815 } & \textbf{0.5248 } & \textbf{0.3763 } \\
    \textbf{Fashion-MNIST} & \textbf{LeNet} & \textbf{0.8985 } & \textbf{0.8653 } & \textbf{0.7261 } & \textbf{0.5274 } & \textbf{0.3480 } & \textbf{0.9463 } & \textbf{0.7778 } & \textbf{0.4932 } & \textbf{0.3961 } \\
      & \textbf{ResNet50} & \textbf{0.9164 } & \textbf{0.8763 } & \textbf{0.7001 } & \textbf{0.4717 } & \textbf{0.3907 } & \textbf{0.9486 } & \textbf{0.7442 } & \textbf{0.4707 } & \textbf{0.4295 } \\
      & \textbf{ResNet101} & \textbf{0.9189 } & \textbf{0.8816 } & \textbf{0.5886 } & \textbf{0.4375 } & \textbf{0.4513 } & \textbf{0.9518 } & \textbf{0.6864 } & \textbf{0.4493 } & \textbf{0.4842 } \\
      & \textbf{Inception V2} & \textbf{0.9159 } & \textbf{0.9030 } & \textbf{0.5392 } & \textbf{0.3804 } & \textbf{0.4991 } & \textbf{0.9657 } & \textbf{0.5765 } & \textbf{0.4485 } & \textbf{0.5001 } \\
      & \textbf{AlexNet} & \textbf{0.9140 } & \textbf{0.9071 } & \textbf{0.5007 } & \textbf{0.3521 } & \textbf{0.5520 } & \textbf{0.9690 } & \textbf{0.5523 } & \textbf{0.4482 } & \textbf{0.5772 } \\
      & \textbf{SqueezeNet} & \textbf{0.8920 } & \textbf{0.9409 } & \textbf{0.4472 } & \textbf{0.3326 } & \textbf{0.6029 } & \textbf{0.9704 } & \textbf{0.5004 } & \textbf{0.4412 } & \textbf{0.6260 } \\
      & \textbf{Inception V1} & \textbf{0.8869 } & \textbf{1.0000 } & \textbf{0.4395 } & \textbf{0.3118 } & \textbf{0.6716 } & \textbf{0.9882 } & \textbf{0.4284 } & \textbf{0.3657 } & \textbf{0.6853 } \\
    \hline\hline
    \end{tabular}}%
  \label{effectiveness}%
\end{table*}%

We denote the $l_\infty$ and $l_2$ form of ROBY metrics as $FSA_\infty$, $FSD_\infty$, $ROBY_\infty$, $FSA_2$, $FSD_2$, $ROBY_2$. And denote the $l_\infty$ and $l_2$ norm PGD attack success rate as $ASR_\infty$ and $ASR_2$.
Each natural model converges after training on the original samples. The classification accuracy (ACC) of the natural model reaches approximately 70\%, 99\%, and 90\% on the CIFAR-10, MNIST, and Fashion-MNIST dataset. It sets the ground for the robustness evaluation of the models with comparable classification performance under the same training settings.

To overview the robustness of models on each dataset, we rank the models in descending order of attack success rate. We find that the ranking of model robustness based on $ASR_\infty$ or $ASR_2$ is the same, showing the consistency of the gold standard. The ASR reflected robustness against $l_2$ attack is weaker than of $l_\infty$. 

Given the established robustness evaluation by ASR, we then compare the ROBY metrics with their corresponding attack success rate in the norm form, as $FSA_\infty$, $FSD_\infty$, $ROBY_\infty$ with $ASR_\infty$, and $FSA_2$, $FSD_2$, $ROBY_2$ with $ASR_2$.

\begin{figure*}[ht]
\centering
\subfigure[]{
\includegraphics[width=0.3\textwidth]{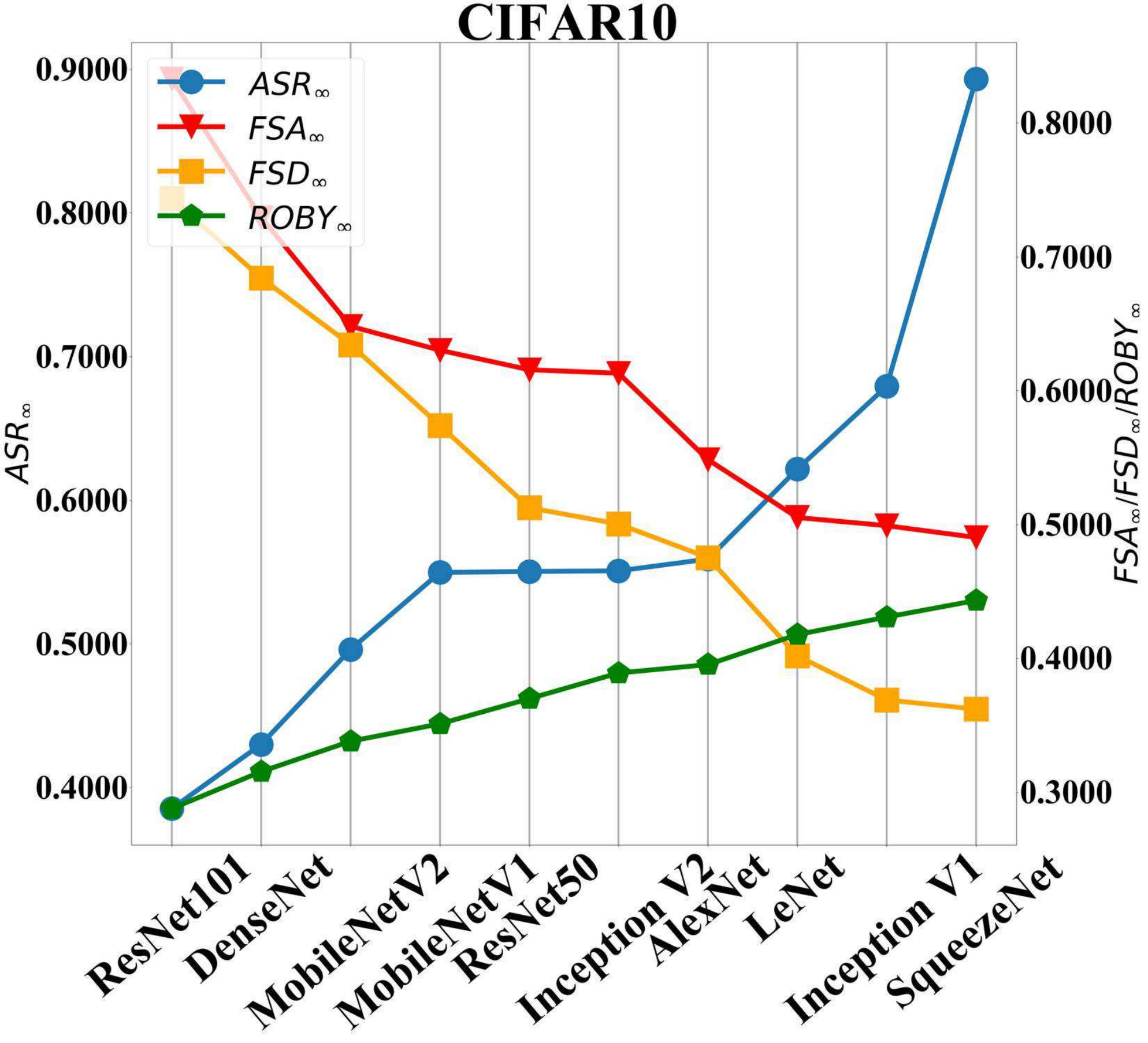} 
}
\subfigure[]{
\includegraphics[width=0.3\textwidth]{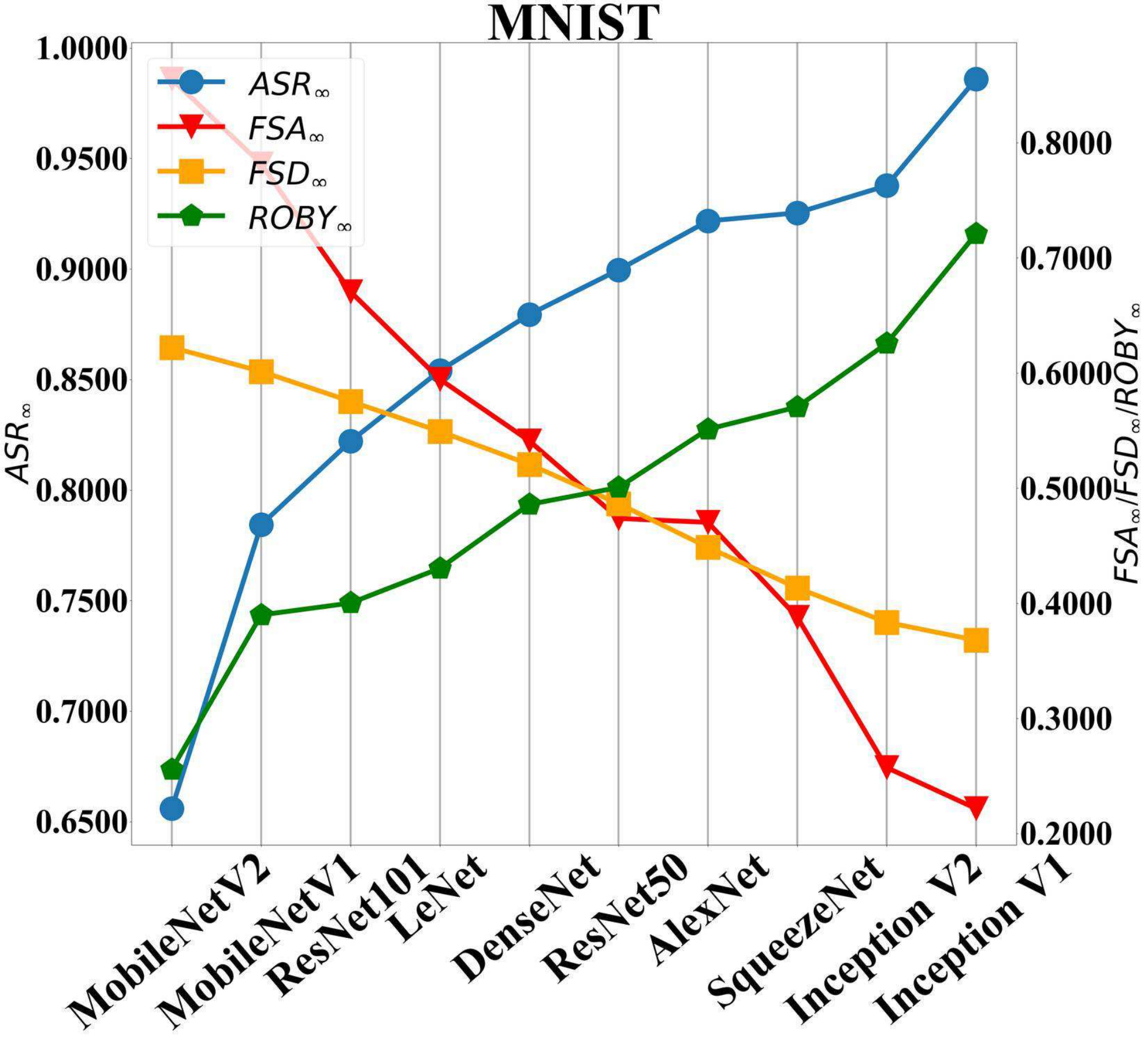}
}
\subfigure[]{
\includegraphics[width=0.3\textwidth]{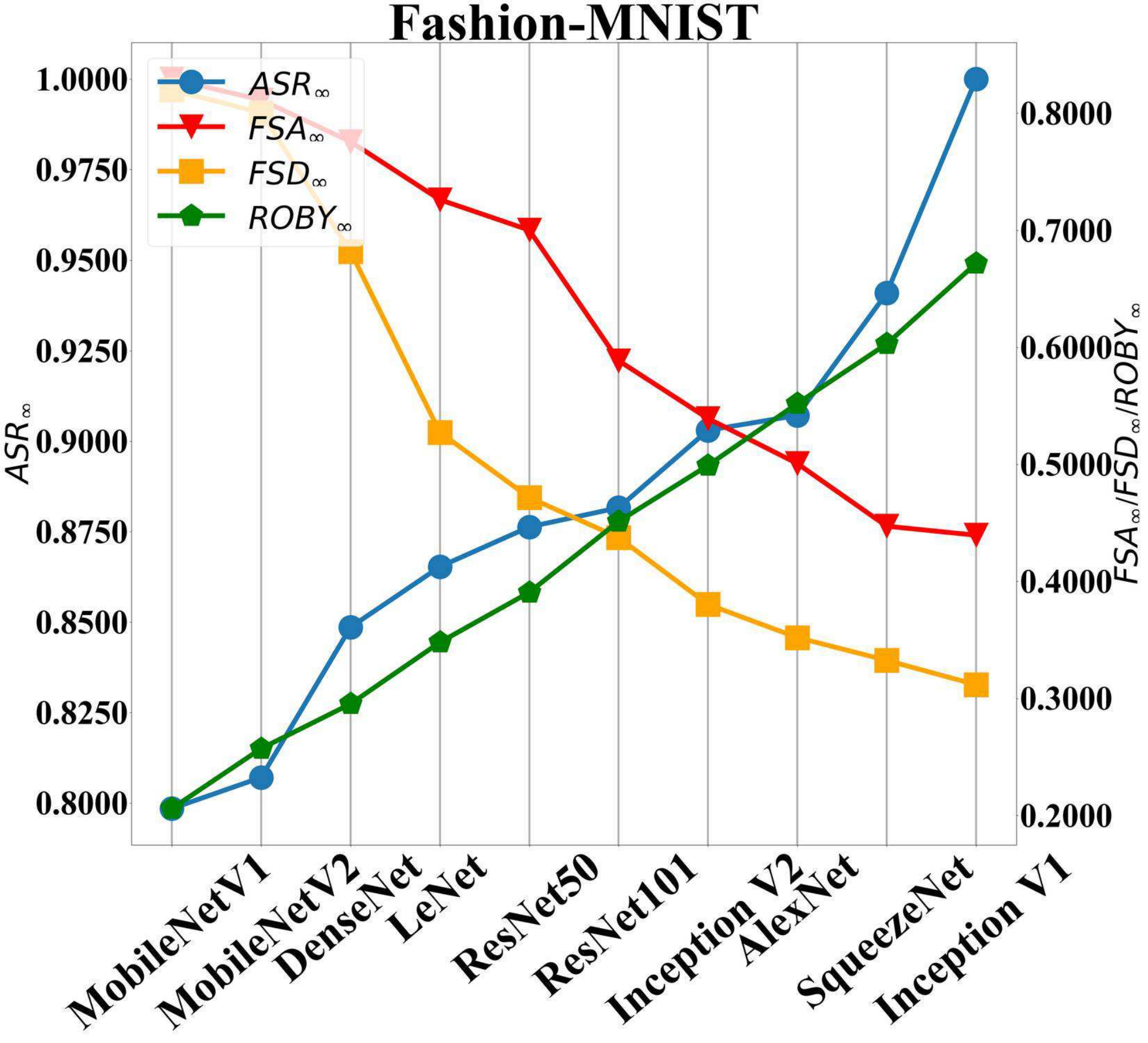}
}
\caption{The relationship between the attack success rate ($ASR_\infty$) and the $l_\infty$ form ROBY metrics.} 
\label{linfi_effectiveness}
\end{figure*}

\begin{figure*}[ht]
\centering
\subfigure[]{
\includegraphics[width=0.3\textwidth]{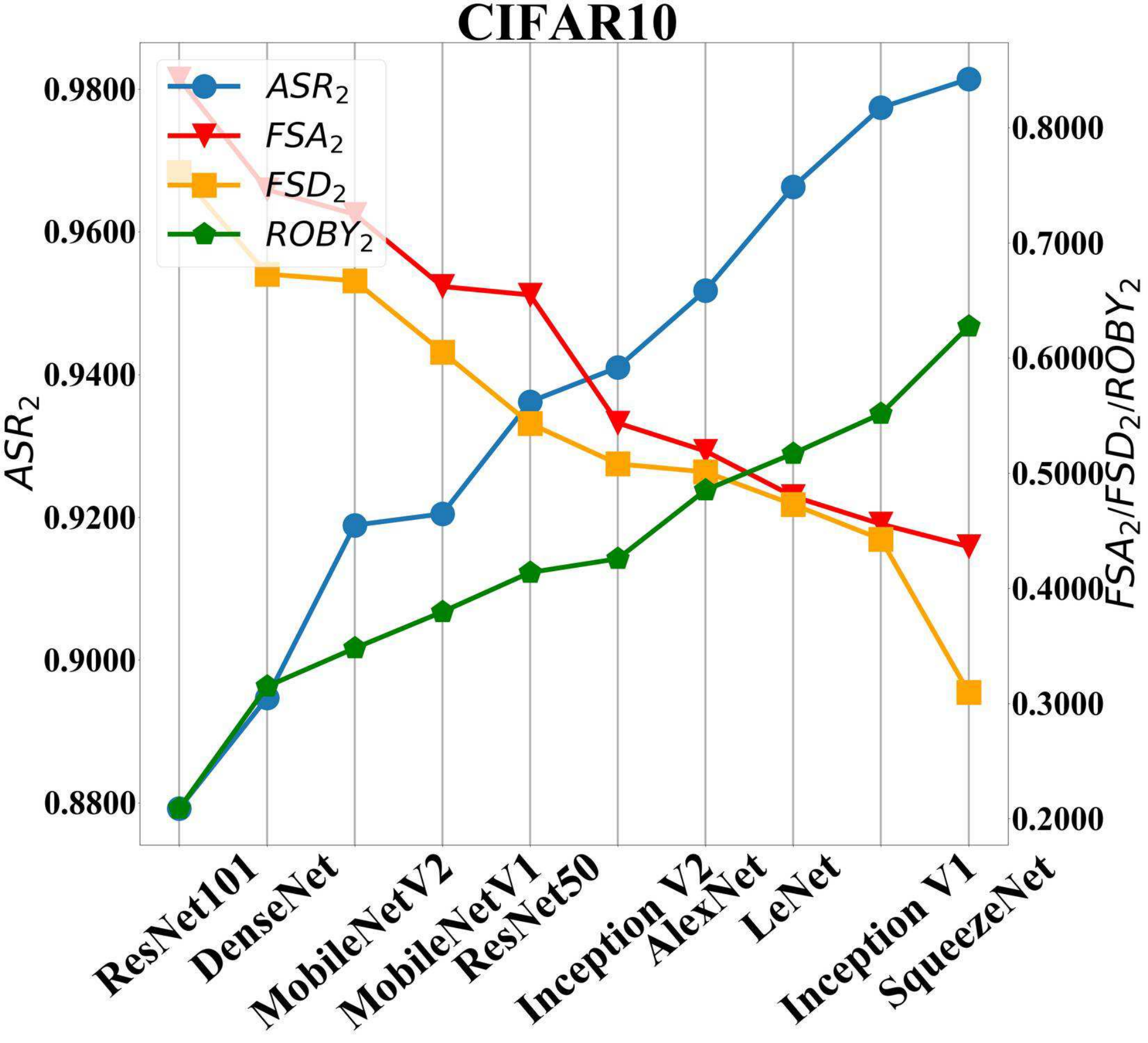} 
}
\subfigure[]{
\includegraphics[width=0.3\textwidth]{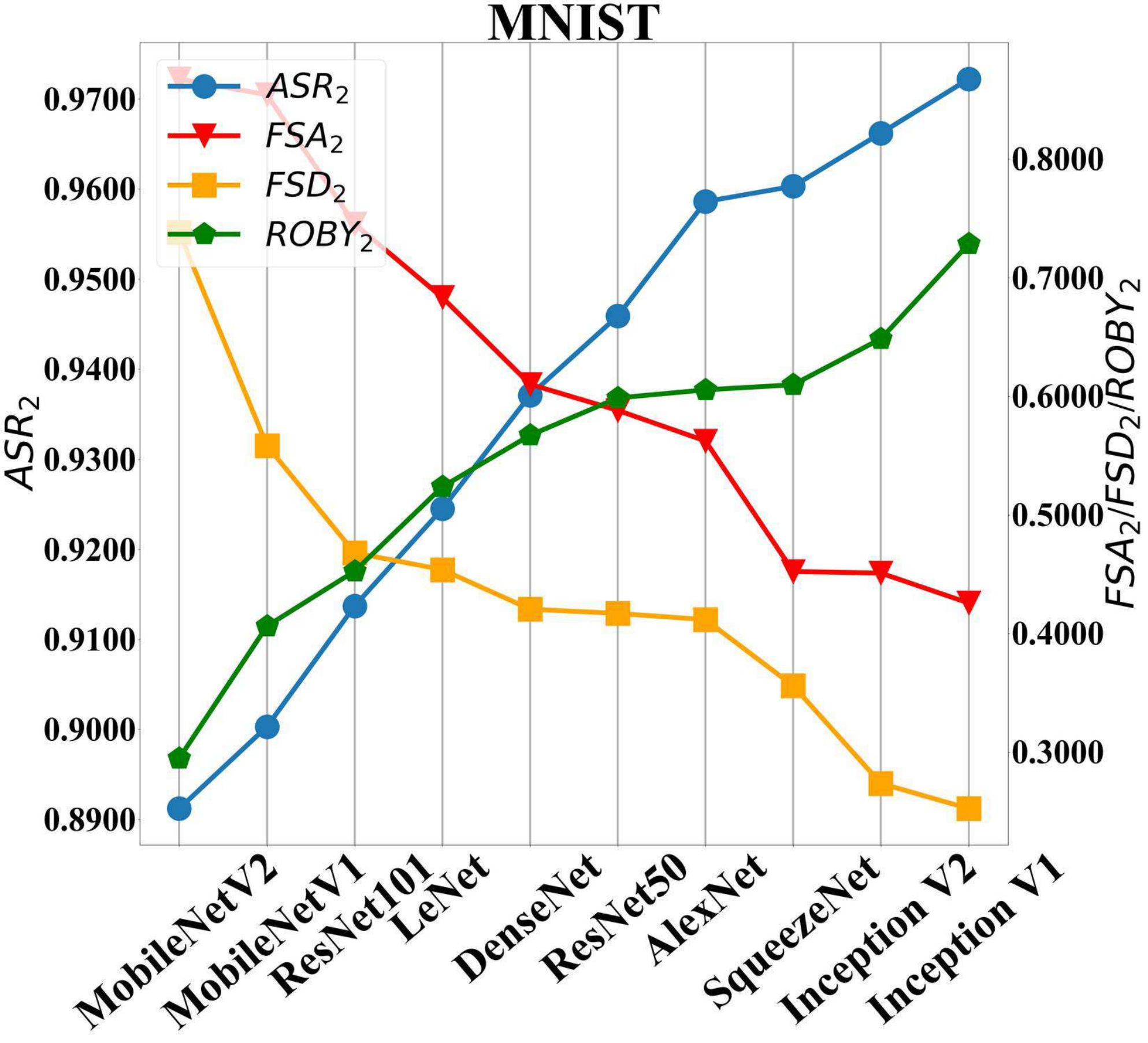}
}
\subfigure[]{
\includegraphics[width=0.3\textwidth]{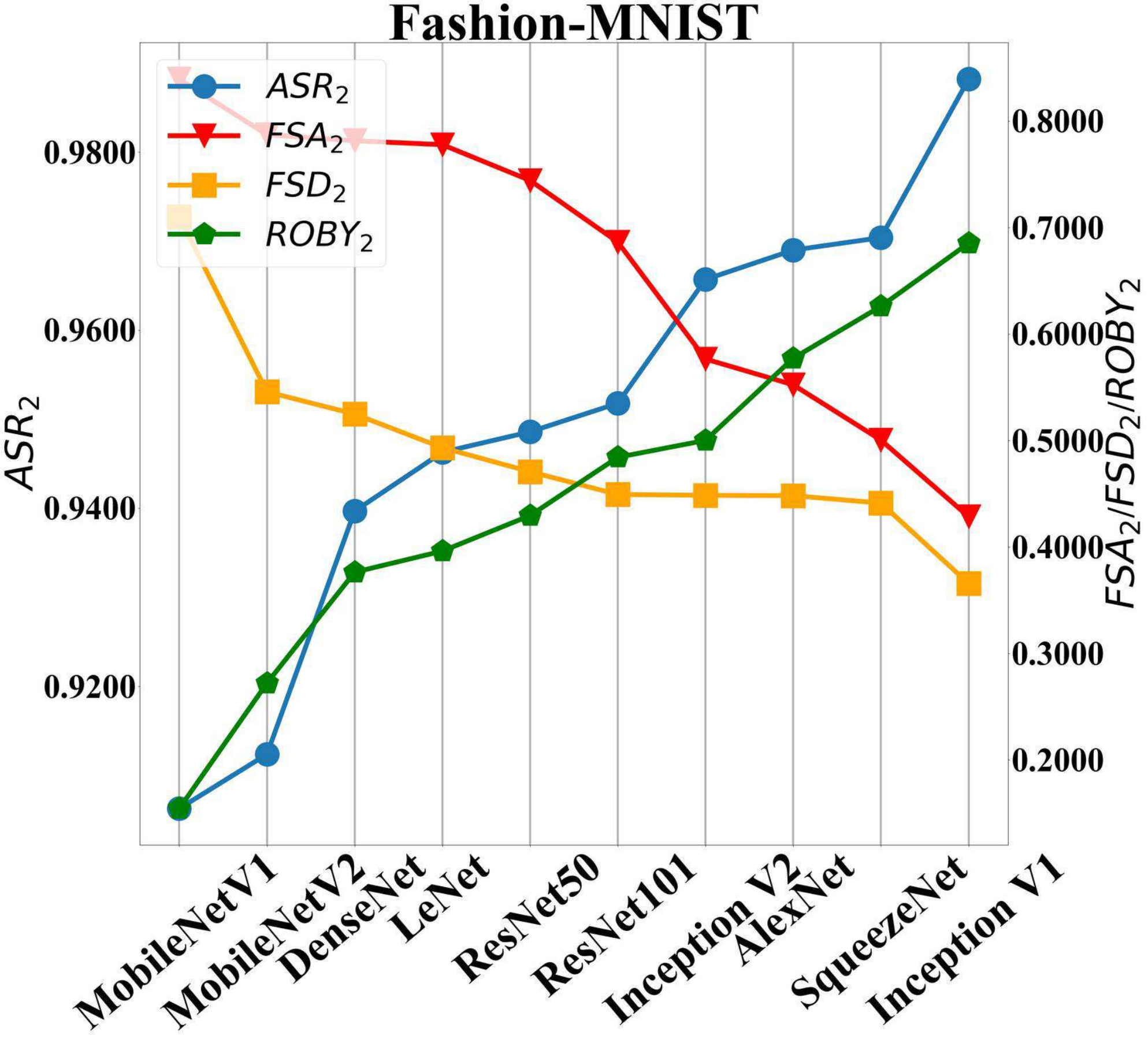}
}
\caption{The relationship between the attack success rate ($ASR_2$) and the $l_2$ form ROBY metrics. }
\label{l2_effectiveness}
\end{figure*}

The visualization of the relationship between the metrics is shown in Fig.~\ref{linfi_effectiveness} and Fig.~\ref{l2_effectiveness}. 

As expected, each model's ROBY metrics well match the attack success rates: the stronger the model's robustness against $l_\infty$ attack, the lower the $ASR_\infty$; the larger the $FSA_\infty$ and $FSD_\infty$, the smaller the $ROBY_\infty$ value.

The stronger the model's robustness against the $l_2$ attack, the lower the $ASR_2$, the larger the $FSA_2$ and $FSD_2$, the smaller the $ROBY_2$ value. We observe the same trend on all three data sets.

We then use  the Pearson correlation coefficient to measure the strength of correlation between the ROBY metrics and the ASR. The calculation goes as :

$$p=\frac{\sum_{i=1}^{N}(x-\bar{x})(y-\bar{y})}{[\sum_{i=1}^{N}(x-\bar{x})^{2}\sum_{i=1}^{N}(x-\bar{x})^{2}]^{\frac{1}{2}}}$$
where $x$ and $y$ are the two variables whose correlations are measured. $\bar{x}$ represents the mean value of the variable $x$, and $\bar{y}$ represents the mean value of $y$.

TABLE~\ref{corrinfi} and TABLE~\ref{corrl2} present the Pearson correlation coefficient of the ROBY metrics and ASR value. The ROBY metrics and model robustness show similar correlation results on the three data sets. The models' robustness against the $l_\infty$ attack and the $FSA_\infty$ and $FSD_\infty$ value are positively correlated, negatively correlated with $ROBY_\infty$. Similarly, the models' robustness against $l_2$ attack and the $FSA_2$ and $FSD_2$ value are positively correlated, negatively correlated with $ROBY_2$. Meanwhile, $ROBY_\infty$ and $ROBY_2$ obtain the highest Pearson correlation coefficient value among all three datasets. Thus, we recommend the $ROBY_\infty$ and $ROBY_2$ as the best evaluation metric, and use them for further experiments.

\begin{table}[htbp]
  \centering
  \caption{Correlation of $l_\infty$ form ROBY metrics and $ASR_\infty$}
    \begin{tabular}{cccc}
    \hline\hline
    \textbf{Data Set} & \textbf{$FSA_\infty$} & \textbf{$FSD_\infty$} & \textbf{$ROBY_\infty$} \\
    \hline
    \textbf{CIFAR-10} & \textbf{-0.8400} & \textbf{-0.8700} & \textbf{\emph{0.8900}} \\
    \hline
    \textbf{MNIST} & \textbf{-0.9300} & \textbf{-0.9000} & \textbf{\emph{0.9600}} \\
    \hline
    \textbf{Fashion-MNIST} & \textbf{-0.9300} & \textbf{-0.9100} & \textbf{\emph{0.9700}} \\
    \hline\hline
    \end{tabular}%
  \label{corrinfi}%
\end{table}%

\begin{table}[htbp]
  \centering
  \caption{Correlation of $l_2$ form ROBY metrics and $ASR_2$}
    \begin{tabular}{cccc}
    \hline\hline
    \textbf{Data Set} & \textbf{$FSA_2$} & \textbf{$FSD_2$} & \textbf{$ROBY_2$} \\
    \hline
    \textbf{CIFAR-10} & \textbf{-0.9800} & \textbf{-0.9600} & \textbf{\emph{0.9800}} \\
    \hline
    \textbf{MNIST} & \textbf{-0.9900} & \textbf{-0.9200} & \textbf{\emph{0.9700}} \\
    \hline
    \textbf{Fashion-MNIST} & \textbf{-0.9100} & \textbf{-0.9100} & \textbf{\emph{0.9800}} \\
    \hline\hline
    \end{tabular}%
  \label{corrl2}%
\end{table}%

\begin{figure*}[ht]
\centering
\subfigure[]{
\includegraphics[width=0.3\textwidth]{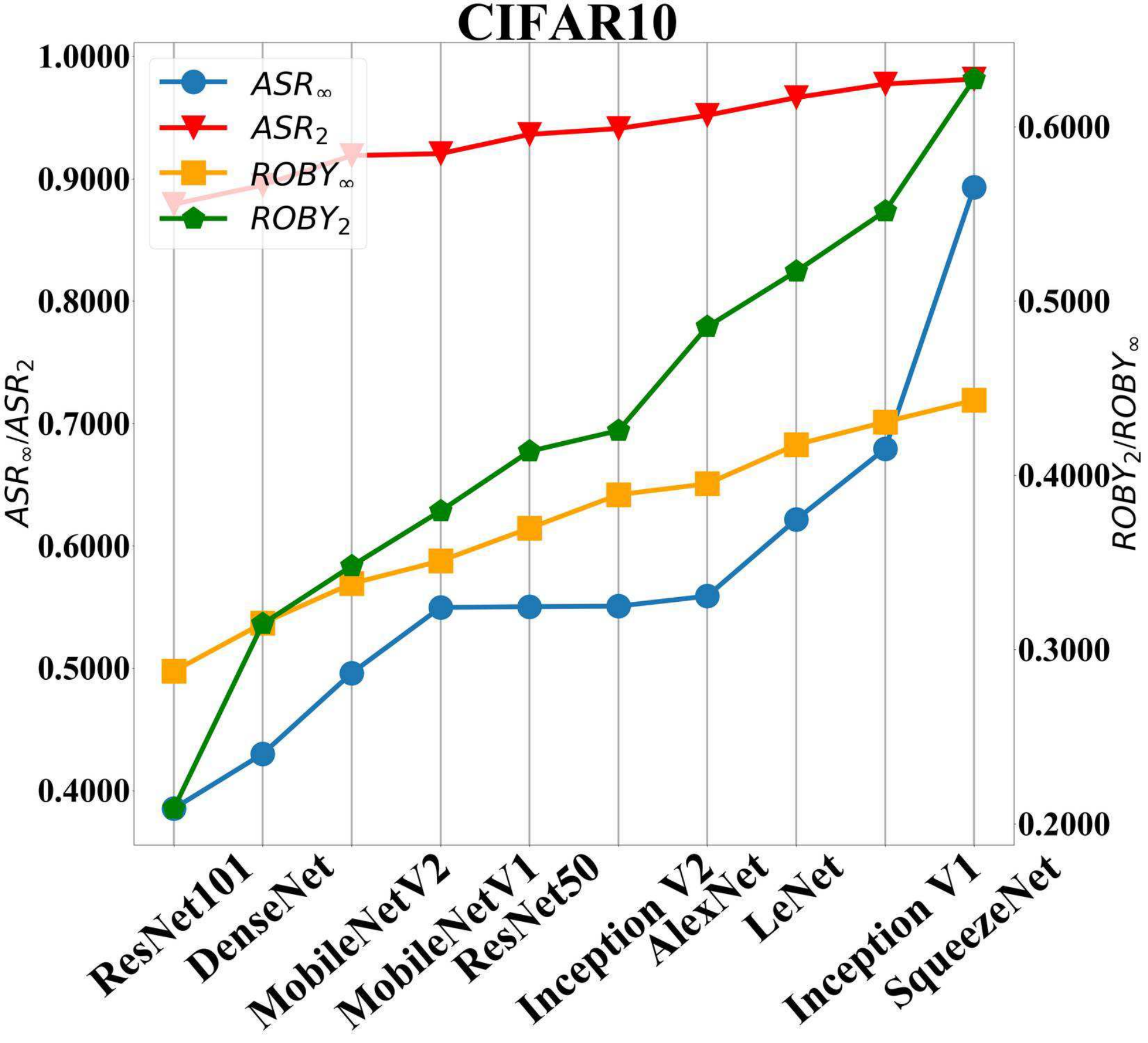} 
}
\subfigure[]{
\includegraphics[width=0.3\textwidth]{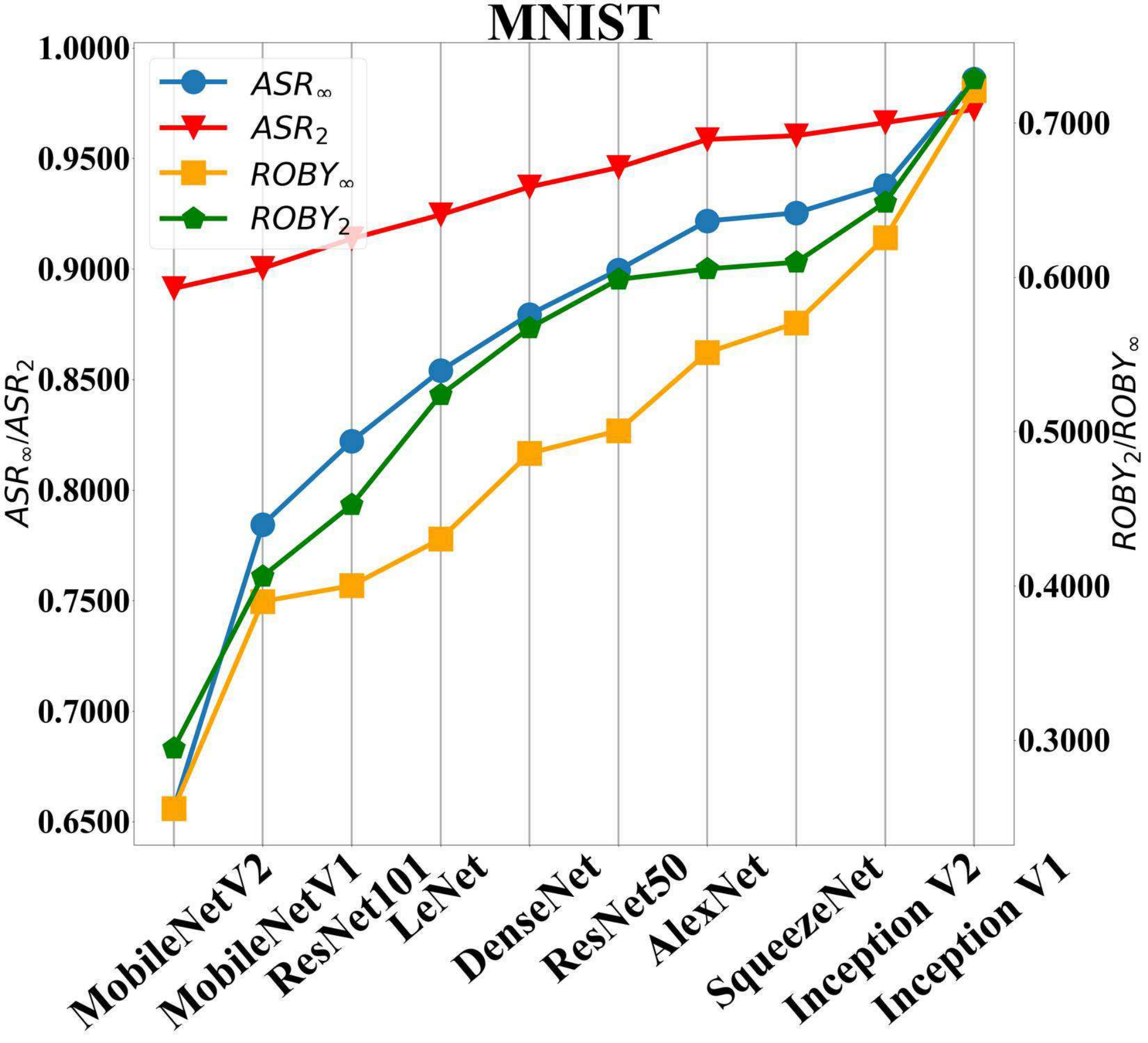}
}
\subfigure[]{
\includegraphics[width=0.3\textwidth]{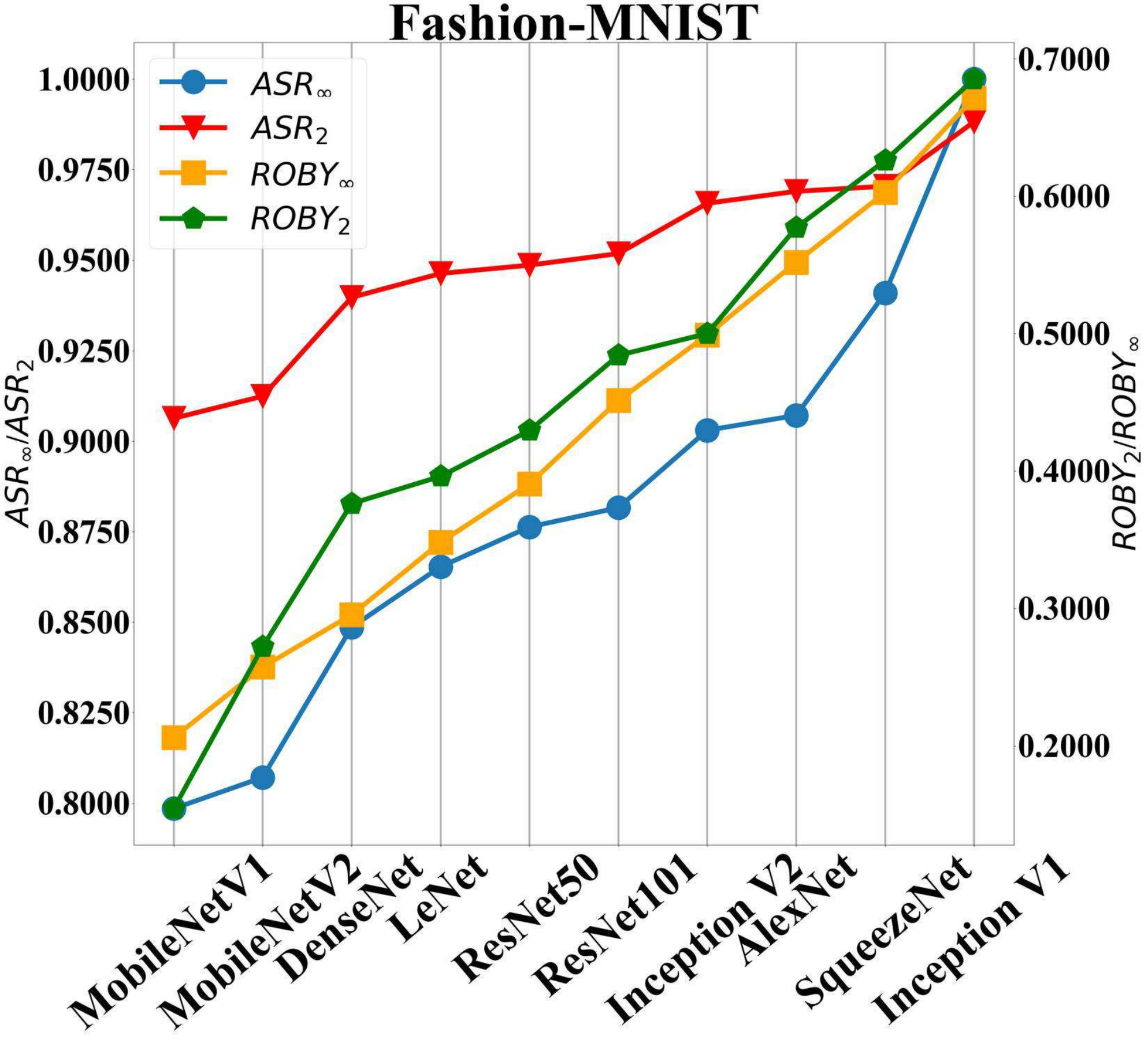}
}
\caption{The relationship between the attack success rate ($ASR_\infty$, $ASR_2$) and the $l_\infty$,$l_2$ form ROBY metrics. }
\label{both_effectiveness}
\end{figure*}

As mentioned before, the robustness ranking results based on $ASR_\infty$ or $ASR_2$ are the same. Meanwhile, ROBY metrics have verified their effectiveness in evaluating related adversarial robustness through comparison with corresponding norm-form attack success rate. Thus, we analyze their effectiveness in evaluating irrelevant adversarial robustness. That is the comparison between $ROBY_\infty$ and $ASR_2$, as well as $ROBY_2 $ and $ASR_\infty$. The visualization results of the relationship between the metrics are shown in Fig.~\ref{both_effectiveness}. These four metrics are all positively correlated with each other.

TABLE~\ref{corrboth} shows the average Pearson correlation coefficient of ROBY metrics and ASR value on three datasets. Both $ROBY_\infty$ and $ROBY_2$ obtains high Pearson correlation coefficient value with both $ASR_\infty$ and $ASR_2$ on all three data sets.

\begin{table}[htbp]
  \centering
  \caption{Correlation of $l_\infty$,$l_2$ form ROBY metrics and $ASR_\infty$,$ASR_2$}
    \begin{tabular}{ccccc}
    \hline\hline
      & \textbf{$ROBY_2$} & \textbf{$ROBY_\infty$} \\
    \midrule
    \textbf{$ASR_2$} & \textbf{0.9760} & \textbf{0.9730} \\
    \textbf{$ASR_\infty$} & \textbf{0.9600} & \textbf{0.9400} \\
    \hline\hline
    \end{tabular}%
  \label{corrboth}%
\end{table}%

It proves the effectiveness of $ROBY_\infty$ on evaluating $l_2$ robustness and $ROBY_2$ on evaluating $l_\infty $ robustness. Surprisingly, $ROBY_2$ obtains a higher Pearson correlation coefficient value with $ASR_\infty$ than $ROBY_\infty$, which, to a certain extent shows the transferability of our metrics.

\subsection{ROBY Evaluates the Robustness of Defensive Models}

Previous research \cite{mickisch2020understanding} showed that $l_\infty$-based adversarial training procedure increased the minimal $l_2$-distance of training and test samples to the decision boundary. We gain a lot of interest in such transferability, which is similar to the remarkable effectiveness of $ROBY_2$ on evaluating model robustness against $l_\infty $ adversarial attack. Correspondingly, to answer \textbf{Q2} and conduct further analysis on $ROBY_2$ and model robustness, we calculated the $ROBY_2$ on both natural and robust models with the original samples and the adversarial samples separately.

In the following experiments, we use $ROBY_2$ to represent the ROBY metrics, denote as ROBY. And use $ASR_\infty $ to represent the attack success rate, denote as ASR.
TABLE~\ref{compare_nr} compares the ROBY metric on natural models and robust defensive models.

\begin{table}[htbp]
  \centering
  \caption{Comparision of the ROBY metric on Natural Model and Robust Model}
  \setlength{\tabcolsep}{0.5mm}{
    \scriptsize 
    \begin{tabular}{ccccccccccc}
      \hline\hline
    \multirow{2}[3]{*}{\textbf{Dataset}} & \multicolumn{2}{c}{\multirow{2}[3]{*}{\textbf{Model}}} & \multicolumn{4}{c}{\textbf{ROBY-ori sample}} & \multicolumn{4}{c}{\textbf{ROBY-adv sample}} \\
\cmidrule{4-11}      & \multicolumn{2}{c}{} & \multicolumn{2}{c}{\textbf{NM}} & \multicolumn{2}{c}{\textbf{RM}} & \multicolumn{2}{c}{\textbf{NM}} & \multicolumn{2}{c}{\textbf{RM}} \\
      \hline\hline
      & \multicolumn{2}{c}{\textbf{ResNet101}} & \multicolumn{2}{c}{\textbf{0.2088 }} & \multicolumn{2}{c}{\textbf{0.2061 }} & \multicolumn{2}{c}{\textbf{0.4353 }} & \multicolumn{2}{c}{\textbf{0.2273 }} \\
      & \multicolumn{2}{c}{\textbf{DenseNet}} & \multicolumn{2}{c}{\textbf{0.3148 }} & \multicolumn{2}{c}{\textbf{0.3084 }} & \multicolumn{2}{c}{\textbf{0.4806 }} & \multicolumn{2}{c}{\textbf{0.4486 }} \\
      & \multicolumn{2}{c}{\textbf{MobileNetV2}} & \multicolumn{2}{c}{\textbf{0.3480 }} & \multicolumn{2}{c}{\textbf{0.3567 }} & \multicolumn{2}{c}{\textbf{0.4750 }} & \multicolumn{2}{c}{\textbf{0.3887 }} \\
    \textbf{CIFAR-10} & \multicolumn{2}{c}{\textbf{MobileNetV1}} & \multicolumn{2}{c}{\textbf{0.3796 }} & \multicolumn{2}{c}{\textbf{0.3415 }} & \multicolumn{2}{c}{\textbf{0.3076 }} & \multicolumn{2}{c}{\textbf{0.2872 }} \\
      & \multicolumn{2}{c}{\textbf{ResNet50}} & \multicolumn{2}{c}{\textbf{0.4138 }} & \multicolumn{2}{c}{\textbf{0.4133 }} & \multicolumn{2}{c}{\textbf{0.3971 }} & \multicolumn{2}{c}{\textbf{0.3476 }} \\
      & \multicolumn{2}{c}{\textbf{Inception V2}} & \multicolumn{2}{c}{\textbf{0.4257 }} & \multicolumn{2}{c}{\textbf{0.4119 }} & \multicolumn{2}{c}{\textbf{0.4835 }} & \multicolumn{2}{c}{\textbf{0.4789 }} \\
      & \multicolumn{2}{c}{\textbf{AlexNet}} & \multicolumn{2}{c}{\textbf{0.4853 }} & \multicolumn{2}{c}{\textbf{0.4072 }} & \multicolumn{2}{c}{\textbf{0.5942 }} & \multicolumn{2}{c}{\textbf{0.4538 }} \\
      & \multicolumn{2}{c}{\textbf{LeNet}} & \multicolumn{2}{c}{\textbf{0.5171 }} & \multicolumn{2}{c}{\textbf{0.4908 }} & \multicolumn{2}{c}{\textbf{0.4674 }} & \multicolumn{2}{c}{\textbf{0.4099 }} \\
      & \multicolumn{2}{c}{\textbf{Inception V1}} & \multicolumn{2}{c}{\textbf{0.5516 }} & \multicolumn{2}{c}{\textbf{0.5497 }} & \multicolumn{2}{c}{\textbf{0.6159 }} & \multicolumn{2}{c}{\textbf{0.5577 }} \\
      & \multicolumn{2}{c}{\textbf{SqueezeNet}} & \multicolumn{2}{c}{\textbf{0.6274 }} & \multicolumn{2}{c}{\textbf{0.6003 }} & \multicolumn{2}{c}{\textbf{0.4787 }} & \multicolumn{2}{c}{\textbf{0.4149 }} \\
    \midrule
      & \multicolumn{2}{c}{\textbf{MobileNetV2}} & \multicolumn{2}{c}{\textbf{0.2948 }} & \multicolumn{2}{c}{\textbf{0.2908 }} & \multicolumn{2}{c}{\textbf{0.6051 }} & \multicolumn{2}{c}{\textbf{0.3891 }} \\
      & \multicolumn{2}{c}{\textbf{MobileNetV1}} & \multicolumn{2}{c}{\textbf{0.4062 }} & \multicolumn{2}{c}{\textbf{0.3746 }} & \multicolumn{2}{c}{\textbf{0.6159 }} & \multicolumn{2}{c}{\textbf{0.5577 }} \\
      & \multicolumn{2}{c}{\textbf{ResNet101}} & \multicolumn{2}{c}{\textbf{0.4525 }} & \multicolumn{2}{c}{\textbf{0.4452 }} & \multicolumn{2}{c}{\textbf{0.4754 }} & \multicolumn{2}{c}{\textbf{0.2937 }} \\
    \textbf{MNIST} & \multicolumn{2}{c}{\textbf{LeNet}} & \multicolumn{2}{c}{\textbf{0.5237 }} & \multicolumn{2}{c}{\textbf{0.4901 }} & \multicolumn{2}{c}{\textbf{0.6126 }} & \multicolumn{2}{c}{\textbf{0.5318 }} \\
      & \multicolumn{2}{c}{\textbf{DenseNet}} & \multicolumn{2}{c}{\textbf{0.5671 }} & \multicolumn{2}{c}{\textbf{0.5607 }} & \multicolumn{2}{c}{\textbf{0.5480 }} & \multicolumn{2}{c}{\textbf{0.5098 }} \\
      & \multicolumn{2}{c}{\textbf{ResNet50}} & \multicolumn{2}{c}{\textbf{0.5986 }} & \multicolumn{2}{c}{\textbf{0.5423 }} & \multicolumn{2}{c}{\textbf{0.5686 }} & \multicolumn{2}{c}{\textbf{0.3588 }} \\
      & \multicolumn{2}{c}{\textbf{AlexNet}} & \multicolumn{2}{c}{\textbf{0.6054 }} & \multicolumn{2}{c}{\textbf{0.5909 }} & \multicolumn{2}{c}{\textbf{0.5938 }} & \multicolumn{2}{c}{\textbf{0.4595 }} \\
      & \multicolumn{2}{c}{\textbf{SqueezeNet}} & \multicolumn{2}{c}{\textbf{0.6097 }} & \multicolumn{2}{c}{\textbf{0.5115 }} & \multicolumn{2}{c}{\textbf{0.6499 }} & \multicolumn{2}{c}{\textbf{0.2937 }} \\
      & \multicolumn{2}{c}{\textbf{Inception V2}} & \multicolumn{2}{c}{\textbf{0.6485 }} & \multicolumn{2}{c}{\textbf{0.6324 }} & \multicolumn{2}{c}{\textbf{0.6324 }} & \multicolumn{2}{c}{\textbf{0.5815 }} \\
      & \multicolumn{2}{c}{\textbf{Inception V1}} & \multicolumn{2}{c}{\textbf{0.7284 }} & \multicolumn{2}{c}{\textbf{0.5690 }} & \multicolumn{2}{c}{\textbf{0.5664 }} & \multicolumn{2}{c}{\textbf{0.4159 }} \\
    \midrule
      & \multicolumn{2}{c}{\textbf{MobileNetV1}} & \multicolumn{2}{c}{\textbf{0.1543 }} & \multicolumn{2}{c}{\textbf{0.1507 }} & \multicolumn{2}{c}{\textbf{0.4160 }} & \multicolumn{2}{c}{\textbf{0.3703 }} \\
      & \multicolumn{2}{c}{\textbf{MobileNetV2}} & \multicolumn{2}{c}{\textbf{0.2721 }} & \multicolumn{2}{c}{\textbf{0.2716 }} & \multicolumn{2}{c}{\textbf{0.3821 }} & \multicolumn{2}{c}{\textbf{0.2857 }} \\
      & \multicolumn{2}{c}{\textbf{DenseNet}} & \multicolumn{2}{c}{\textbf{0.3763 }} & \multicolumn{2}{c}{\textbf{0.3743 }} & \multicolumn{2}{c}{\textbf{0.4904 }} & \multicolumn{2}{c}{\textbf{0.4449 }} \\
    \textbf{Fashion-MNIST} & \multicolumn{2}{c}{\textbf{LeNet}} & \multicolumn{2}{c}{\textbf{0.3961 }} & \multicolumn{2}{c}{\textbf{0.3671 }} & \multicolumn{2}{c}{\textbf{0.5215 }} & \multicolumn{2}{c}{\textbf{0.5207 }} \\
      & \multicolumn{2}{c}{\textbf{ResNet50}} & \multicolumn{2}{c}{\textbf{0.4295 }} & \multicolumn{2}{c}{\textbf{0.4077 }} & \multicolumn{2}{c}{\textbf{0.4193 }} & \multicolumn{2}{c}{\textbf{0.3936 }} \\
      & \multicolumn{2}{c}{\textbf{ResNet101}} & \multicolumn{2}{c}{\textbf{0.4842 }} & \multicolumn{2}{c}{\textbf{0.3422 }} & \multicolumn{2}{c}{\textbf{0.3804 }} & \multicolumn{2}{c}{\textbf{0.3489 }} \\
      & \multicolumn{2}{c}{\textbf{Inception V2}} & \multicolumn{2}{c}{\textbf{0.5001 }} & \multicolumn{2}{c}{\textbf{0.4916 }} & \multicolumn{2}{c}{\textbf{0.5326 }} & \multicolumn{2}{c}{\textbf{0.4886 }} \\
      & \multicolumn{2}{c}{\textbf{AlexNet}} & \multicolumn{2}{c}{\textbf{0.5772 }} & \multicolumn{2}{c}{\textbf{0.5428 }} & \multicolumn{2}{c}{\textbf{0.6909 }} & \multicolumn{2}{c}{\textbf{0.6729 }} \\
      & \multicolumn{2}{c}{\textbf{SqueezeNet}} & \multicolumn{2}{c}{\textbf{0.6260 }} & \multicolumn{2}{c}{\textbf{0.4390 }} & \multicolumn{2}{c}{\textbf{0.6358 }} & \multicolumn{2}{c}{\textbf{0.5182 }} \\
      & \multicolumn{2}{c}{\textbf{Inception V1}} & \multicolumn{2}{c}{\textbf{0.6853 }} & \multicolumn{2}{c}{\textbf{0.4779 }} & \multicolumn{2}{c}{\textbf{0.4982 }} & \multicolumn{2}{c}{\textbf{0.4752 }} \\
      \hline\hline
    \end{tabular}}%
  \label{compare_nr}%
\end{table}%

As for the general ROBY metric calculated with original samples, the robust models have a smaller ROBY value than the natural models on all three datasets. This result verifies ROBY metric's effectiveness on robustness evaluation as the $l_\infty $-norm PGD adversarial training a promising defense mechanism. 
Meanwhile, another interesting result is that, as for ROBY metric calculated with adversarial samples, the robust model also has a smaller ROBY value than the natural model and obtain a more significant decline than general ROBY.

\begin{figure*}[htb]
\centering
\subfigure[CIFAR-10]{
\includegraphics[width=0.3\textwidth]{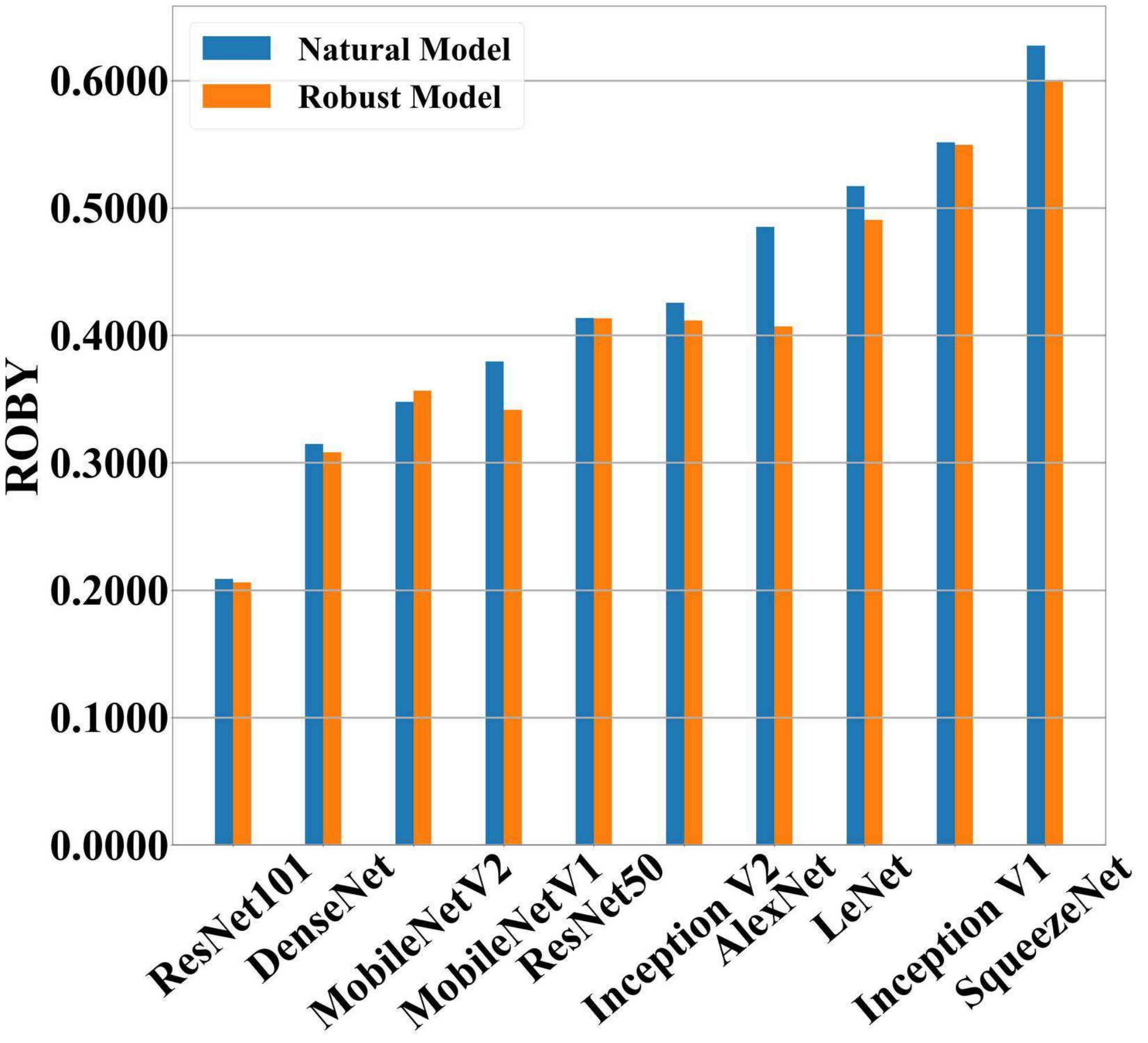} 
}
\subfigure[MNIST]{
\includegraphics[width=0.3\textwidth]{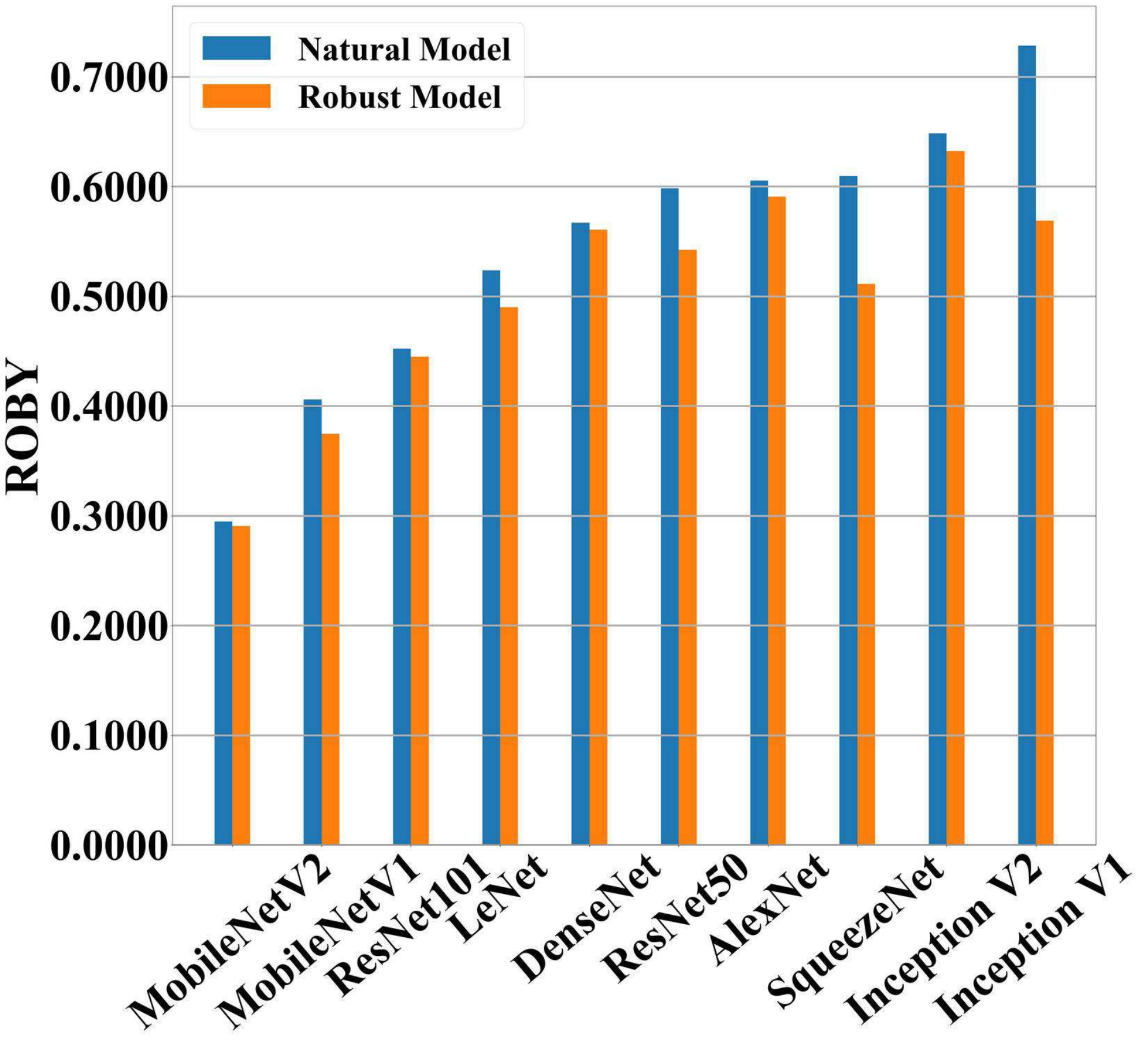}
}
\subfigure[Fashion-MNIST]{
\includegraphics[width=0.3\textwidth]{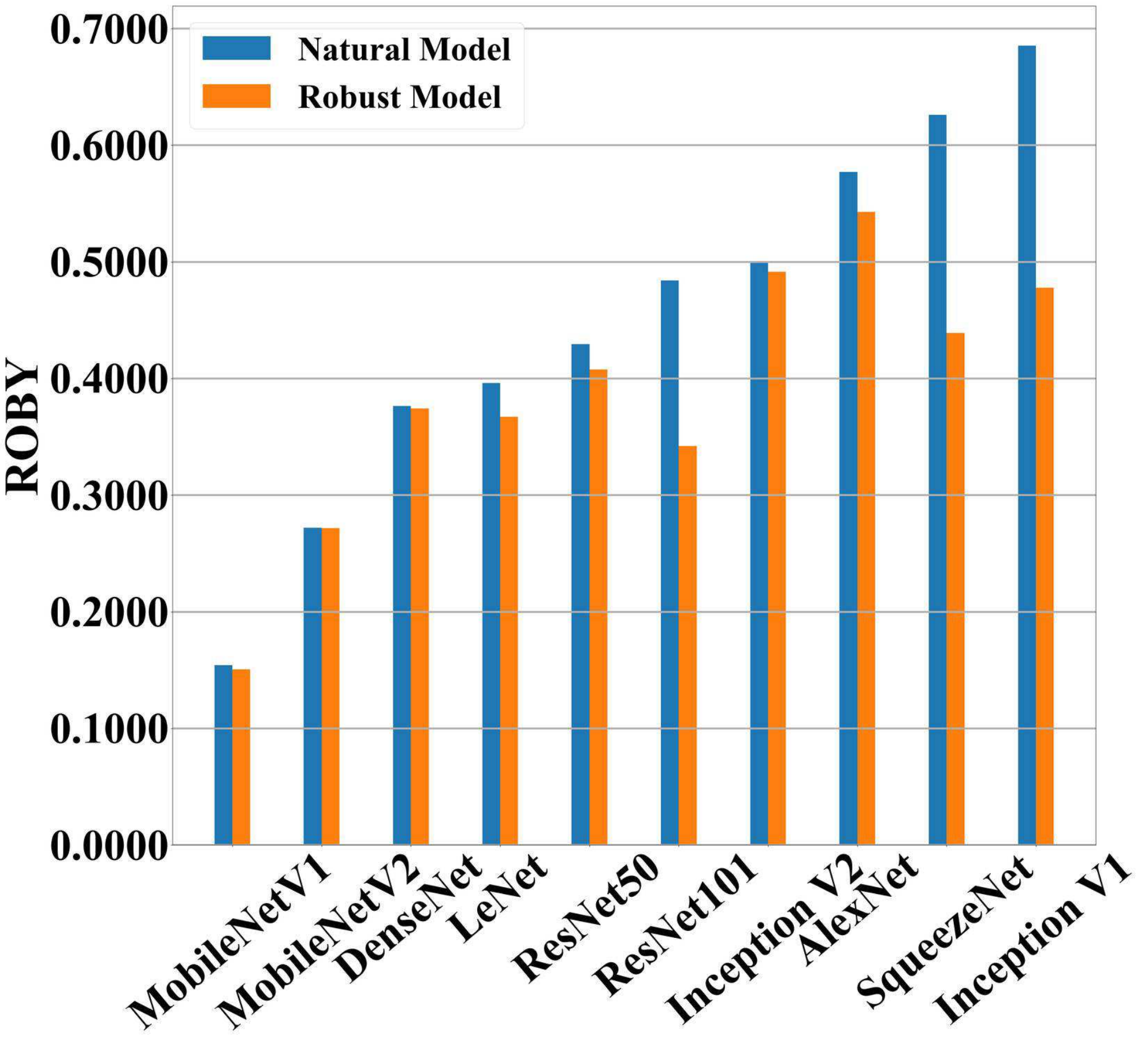}
}
\caption{The comparison of ROBY metric between natural model and robust model on original samples.}
\label{compare_nr_ori}
\end{figure*}

\begin{figure*}[htb]
\centering
\subfigure[CIFAR-10]{
\includegraphics[width=0.3\textwidth]{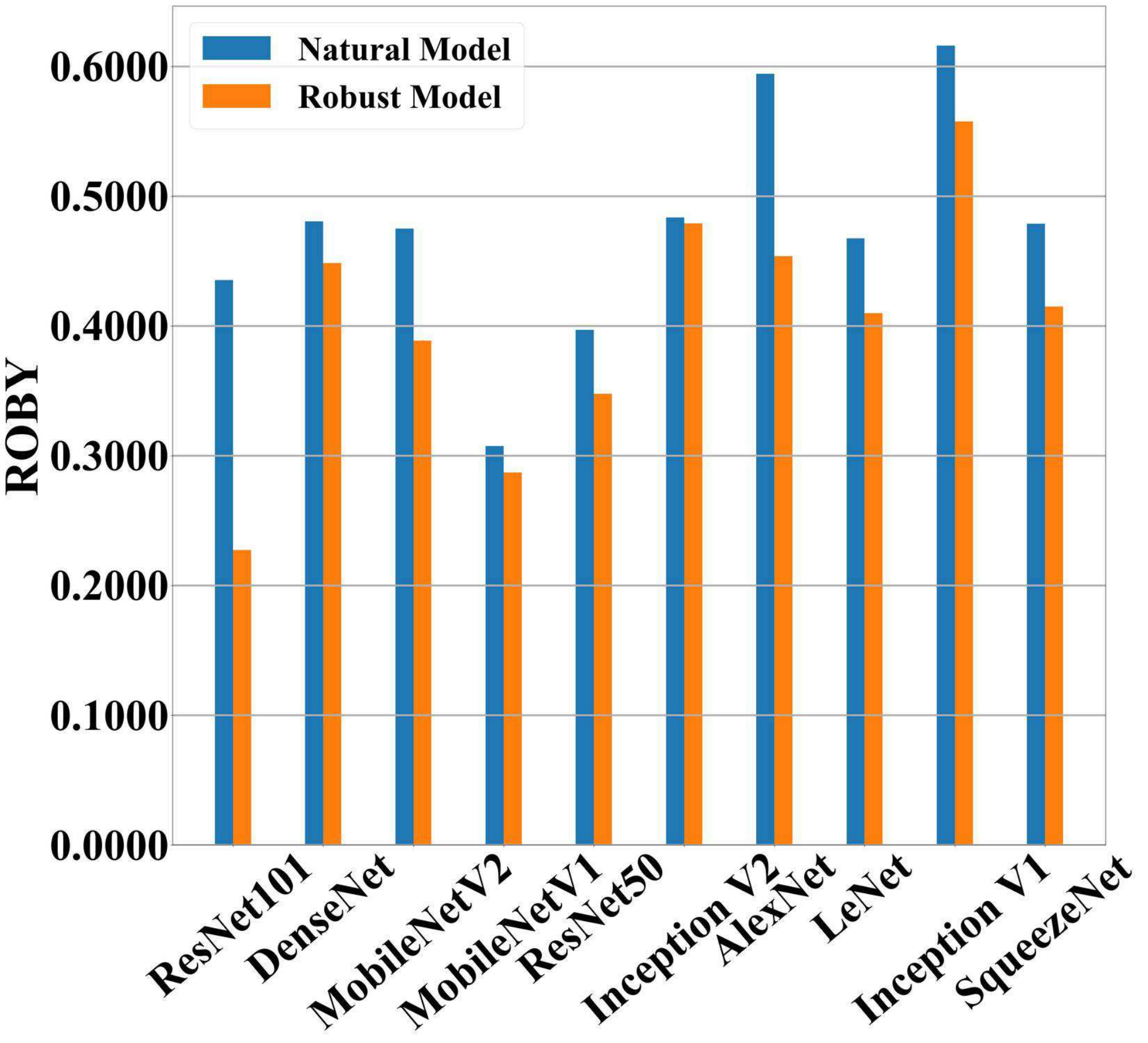} 
}
\subfigure[MNIST]{
\includegraphics[width=0.3\textwidth]{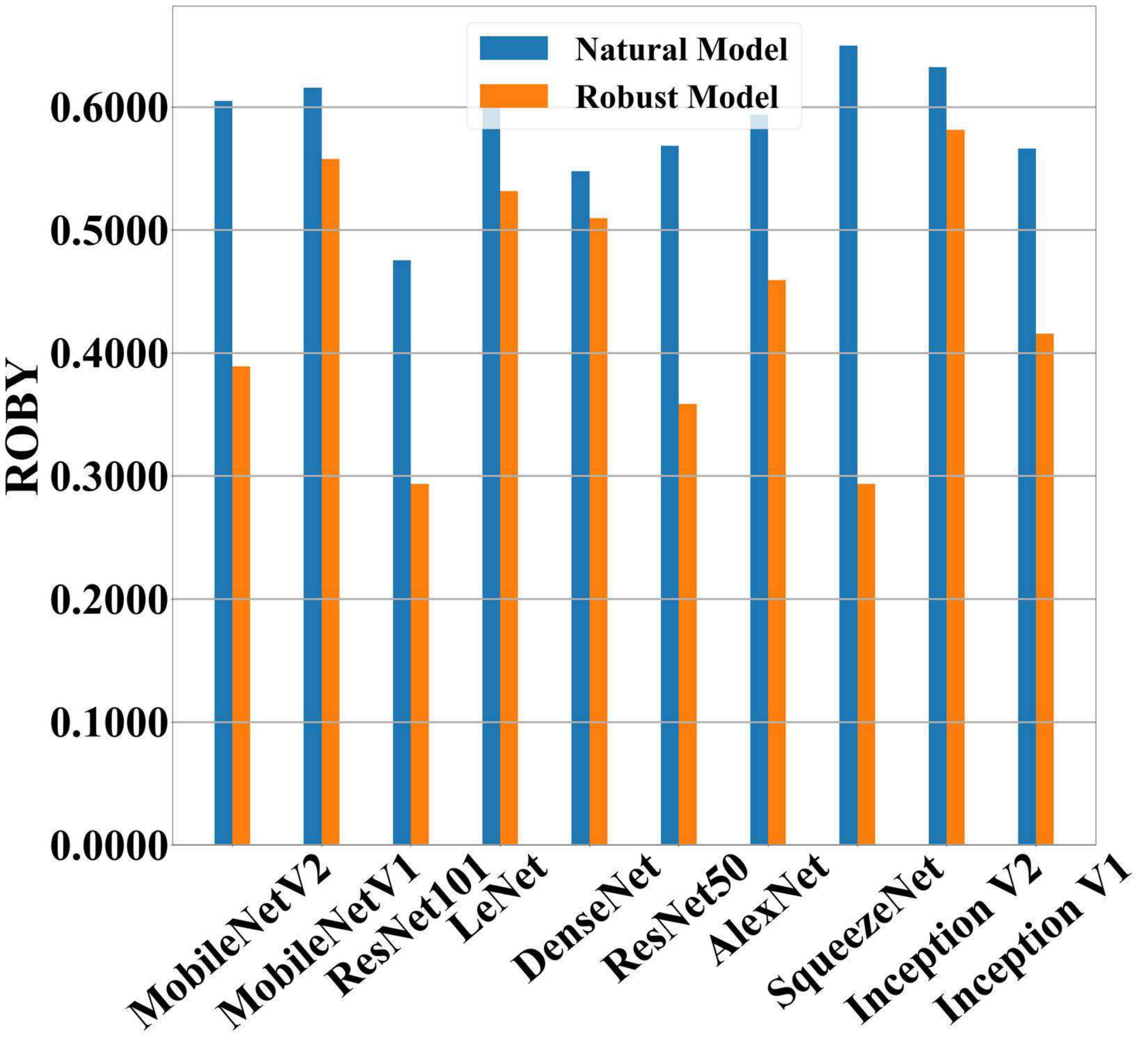}
}
\subfigure[Fashion-MNIST]{
\includegraphics[width=0.3\textwidth]{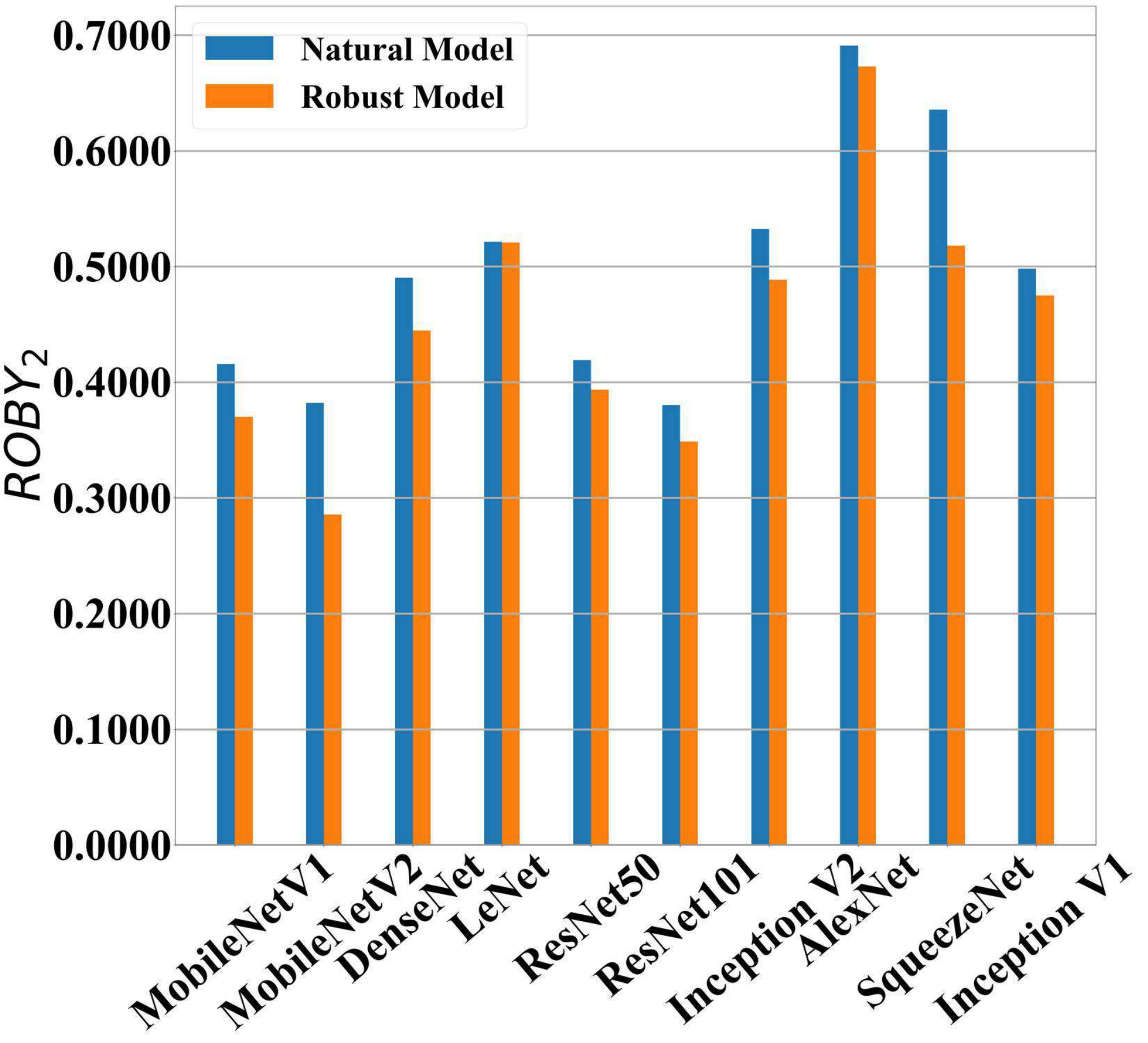}
}
\caption{The comparison of ROBY metric between natural model and robust model on adversarial samples.}
\label{compare_nr_adv}
\end{figure*}

\begin{figure*}[ht]
\centering
\subfigure[MobileNetV2, ROBY=0.2948]{
\includegraphics[width=0.23\textwidth]{pic6a.pdf} 
}
\subfigure[ResNet101, ROBY=0.4525]{
\includegraphics[width=0.23\textwidth]{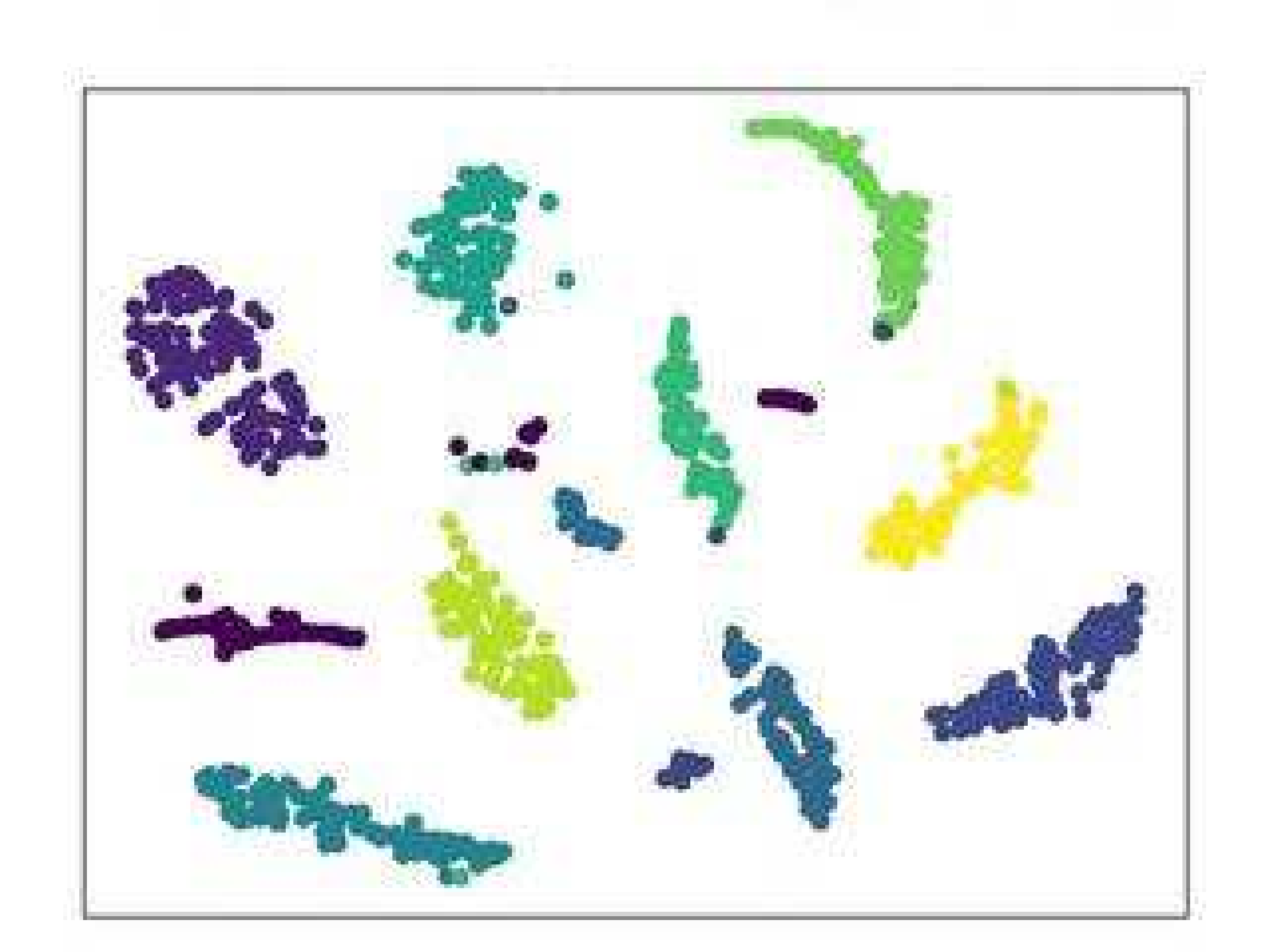}
}
\subfigure[LeNet, ROBY=0.5237]{
\includegraphics[width=0.23\textwidth]{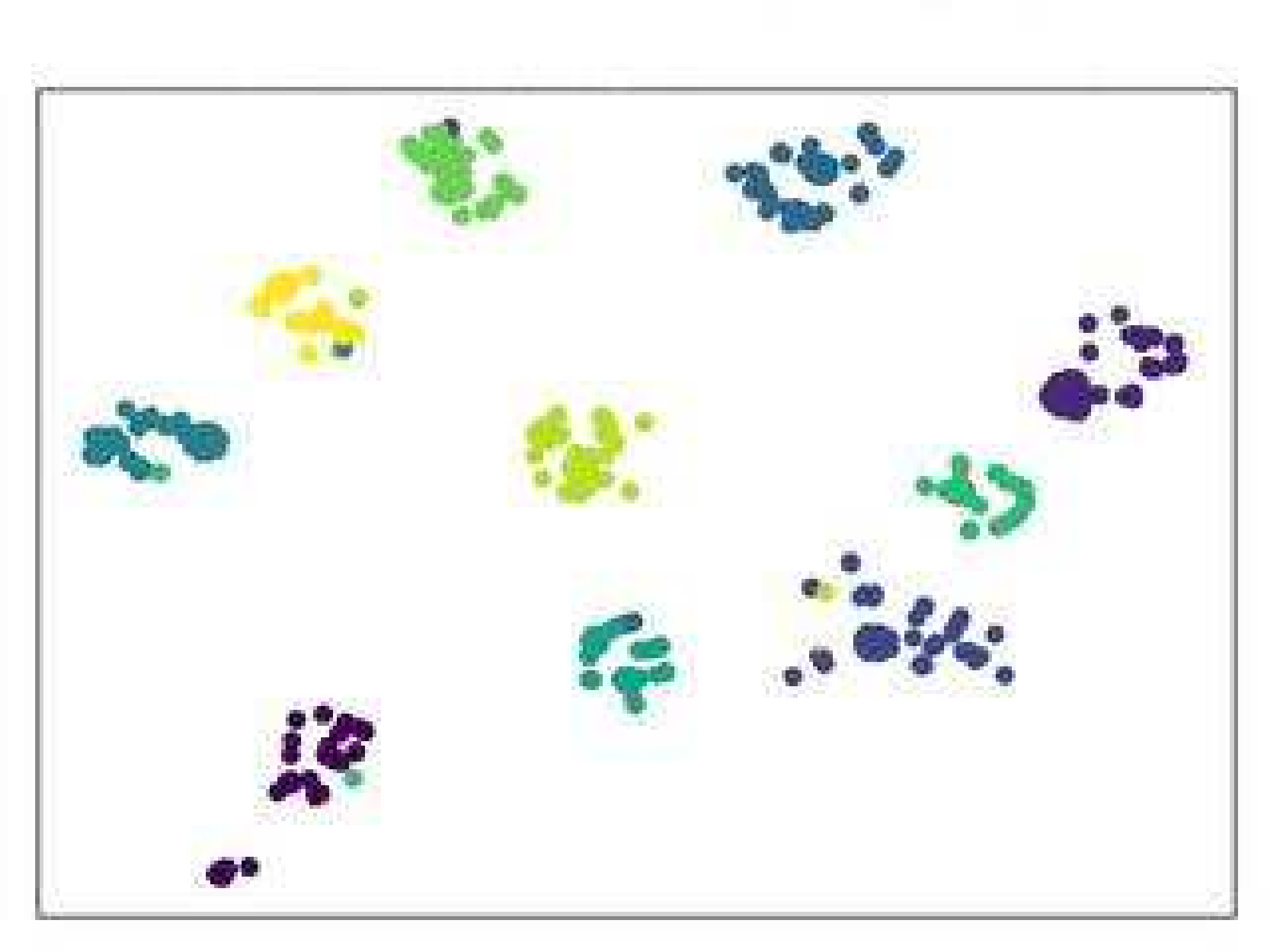}
}
\subfigure[InceptionV2, ROBY=0.6485]{
\includegraphics[width=0.23\textwidth]{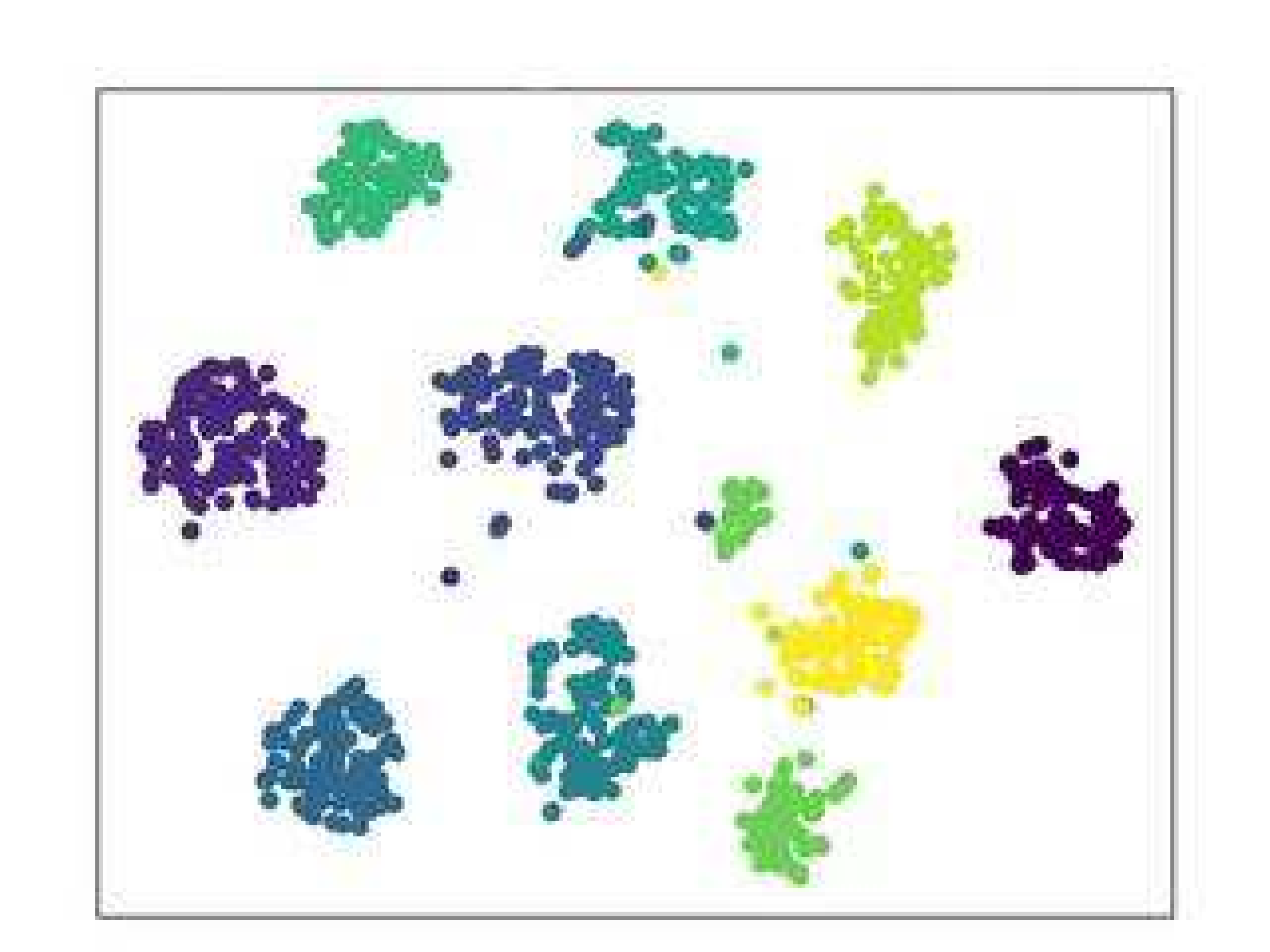} 
}
\caption{The t-SNE embedding of different natural models with original samples on MNIST.}
\label{tsne_no}
\end{figure*}

\begin{figure*}[ht]
\centering
\subfigure[MobileNetV2, ROBY=0.6051]{
\includegraphics[width=0.23\textwidth]{pic7a.pdf} 
}
\subfigure[ResNet101, ROBY=0.4754]{
\includegraphics[width=0.23\textwidth]{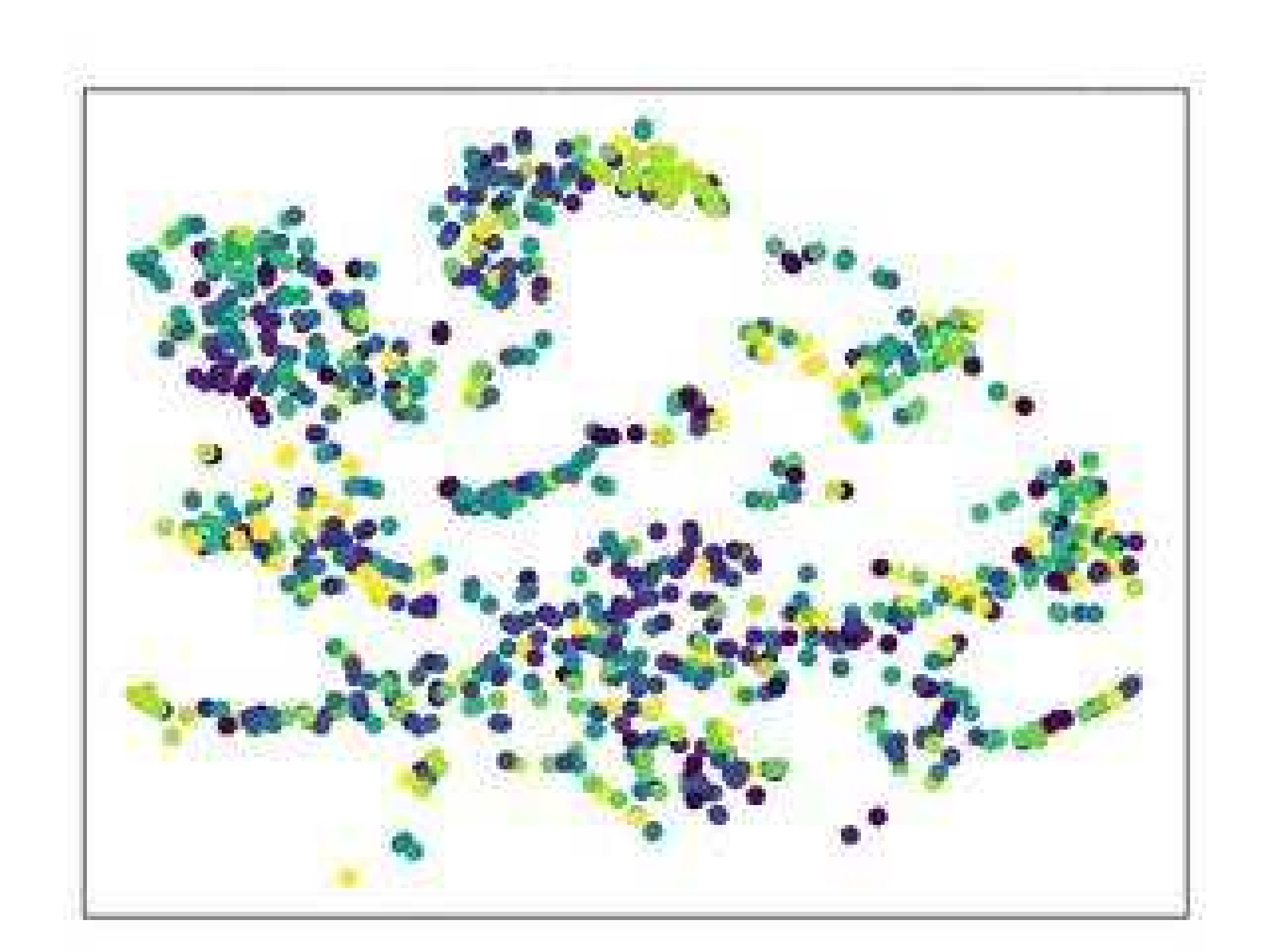}
}
\subfigure[LeNet, ROBY=0.6126]{
\includegraphics[width=0.23\textwidth]{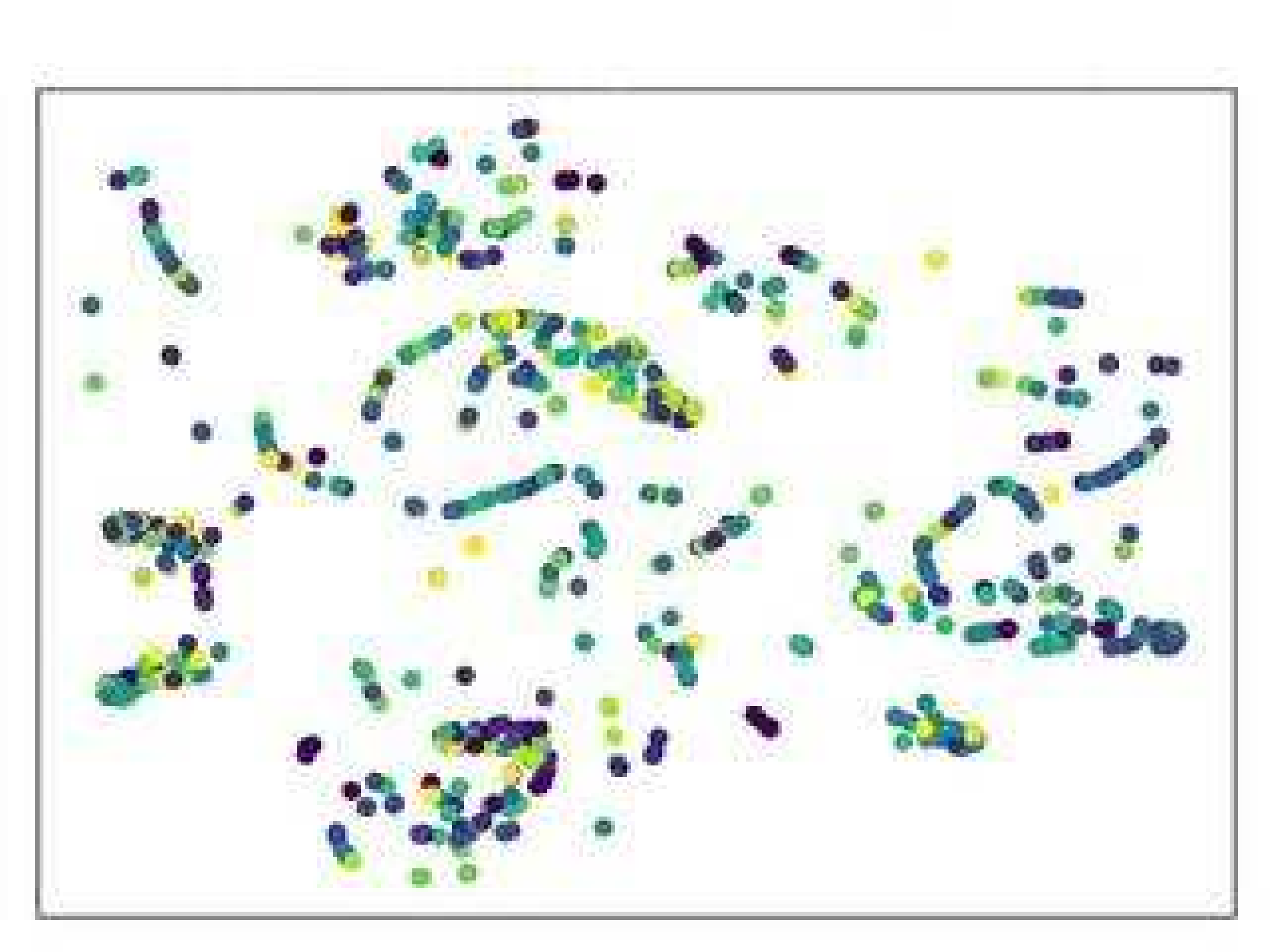}
}
\subfigure[InceptionV2, ROBY=0.6324]{
\includegraphics[width=0.23\textwidth]{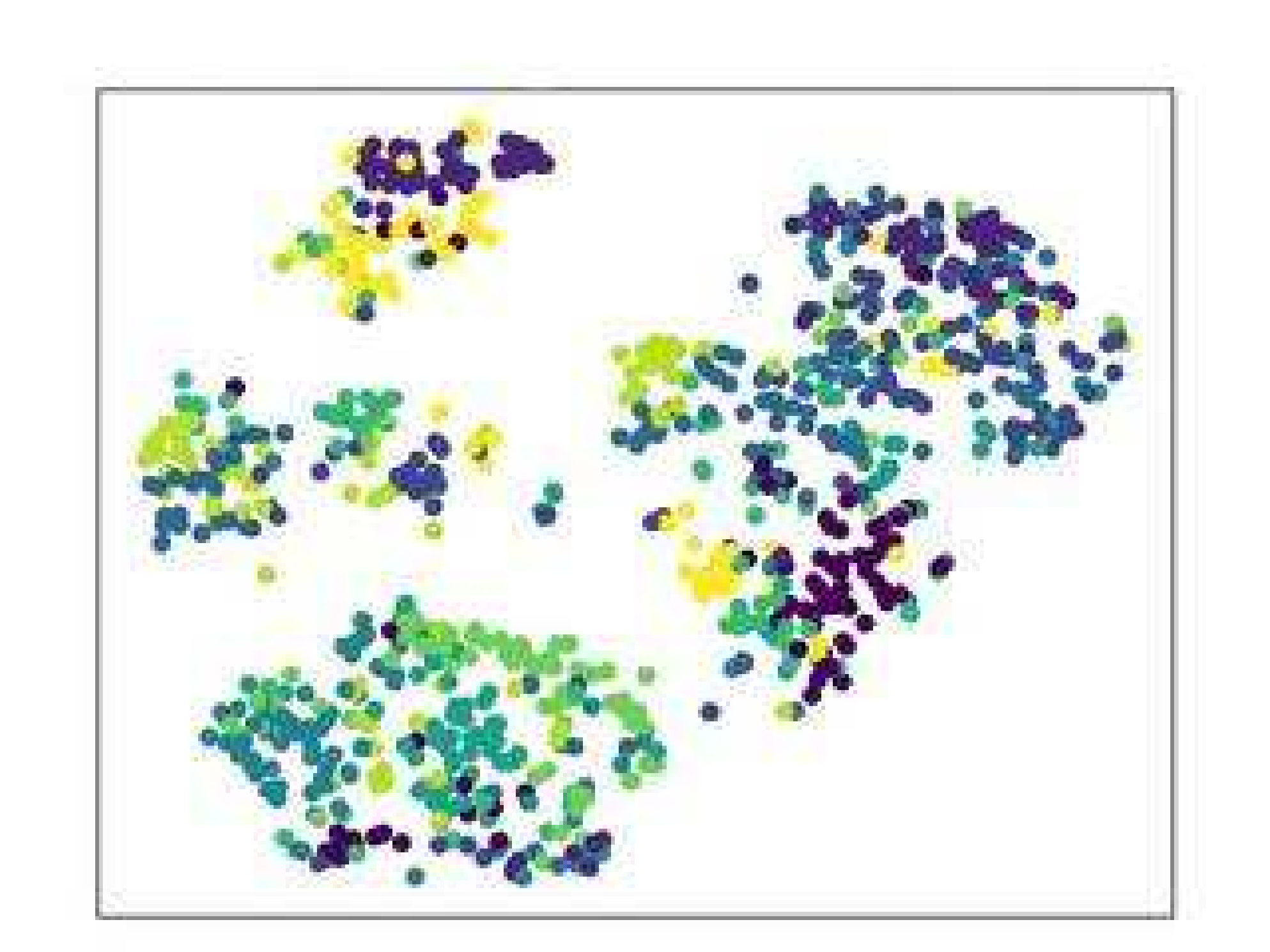} 
}
\caption{The t-SNE embedding of different natural models with adversarial samples on MNIST.}
\label{tsne_na}
\end{figure*}

\begin{figure*}[ht]
\centering
\subfigure[MobileNetV2, ROBY=0.3891]{
\includegraphics[width=0.23\textwidth]{pic8a.pdf} 
}
\subfigure[ResNet101, ROBY=0.2937]{
\includegraphics[width=0.23\textwidth]{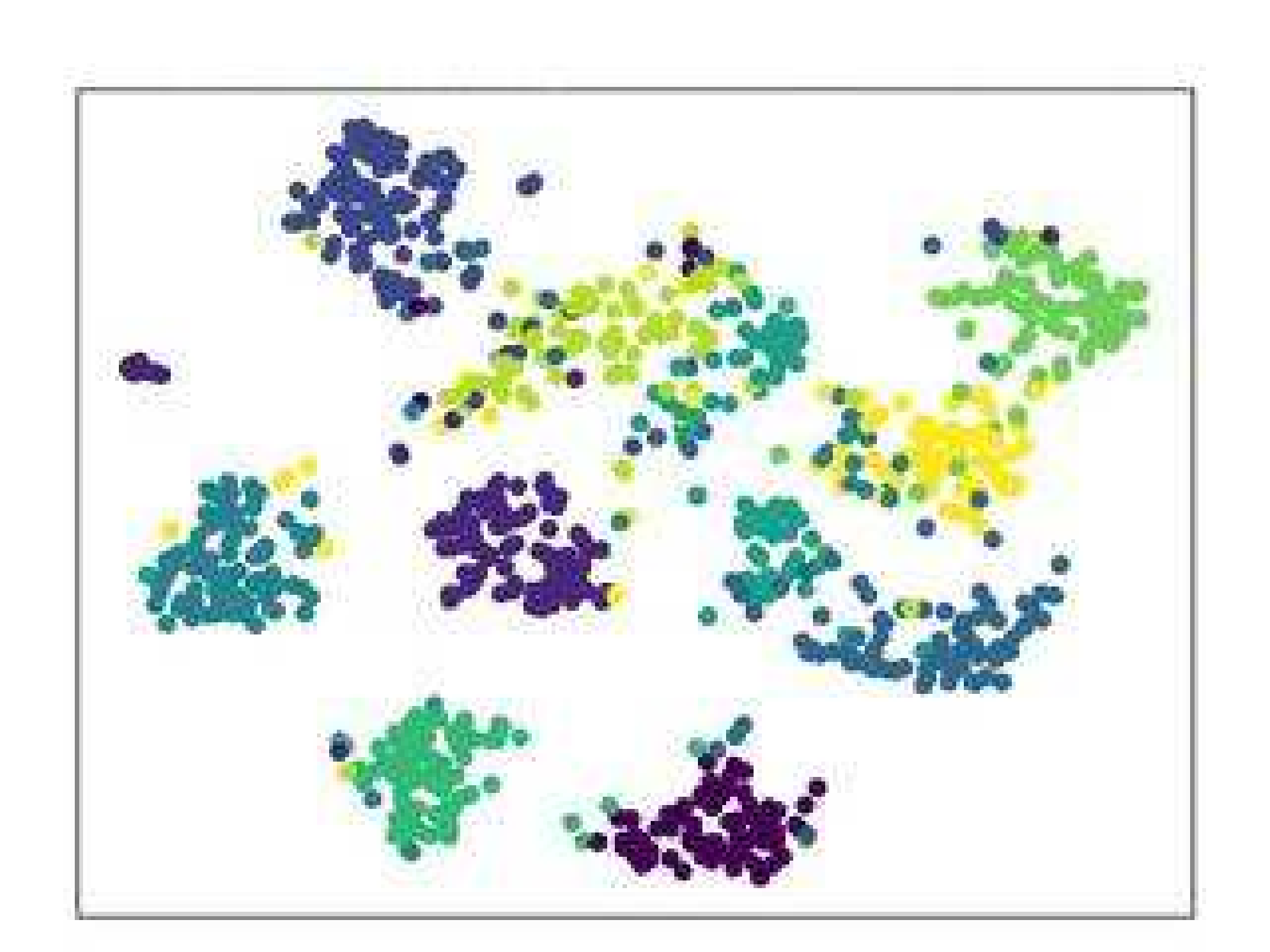}
}
\subfigure[LeNet, ROBY=0.5318]{
\includegraphics[width=0.23\textwidth]{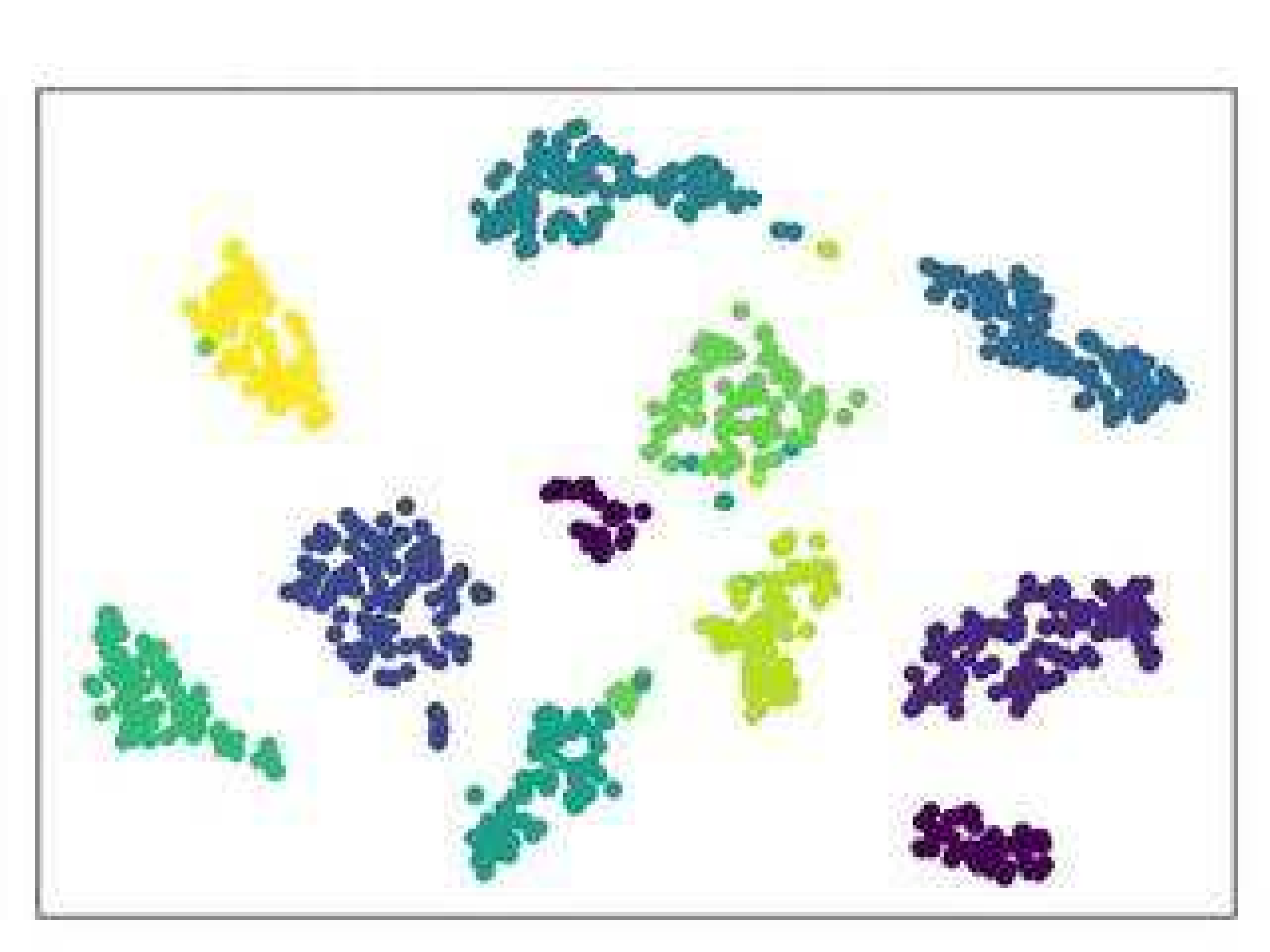}
}
\subfigure[InceptionV2, ROBY=0.5815]{
\includegraphics[width=0.23\textwidth]{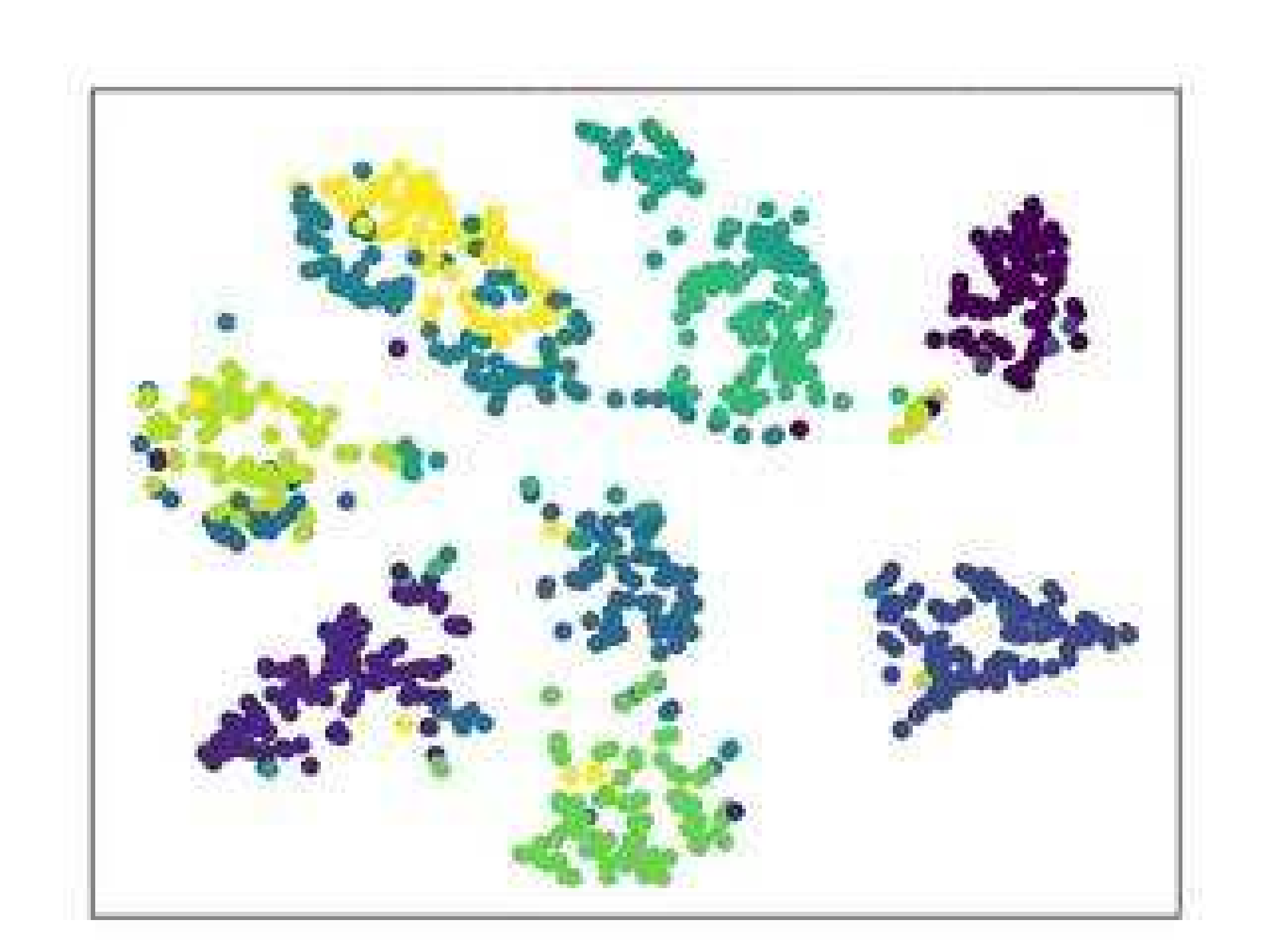} 
}
\caption{The t-SNE embedding of different robusts model with adversarial samples on MNIST.}
\label{tsne_ra}
\end{figure*}

The visualization result of the comparison is shown in Fig.~\ref{compare_nr_ori} and Fig.~\ref{compare_nr_adv}. The robust models have a smaller ROBY value than the natural models on both original samples and adversarial samples.
We applied t-SNE embedding on different natural models and robust models to draw the decision boundary by giving each classification output a two-dimensional location.  Fig.~\ref{tsne_no}, Fig.~\ref{tsne_na}, and Fig.~\ref{tsne_ra} show the diversity of decision boundaries of different models. As expected, the different natural and robust models have different decision boundaries, and the decision boundaries of robust models are more complicated and clear than natural models on adversarial samples. Furthermore, the robust model obtains higher compactness of data in the same class and more significant distinction across classes, which results in less overlapping among all the classes and smaller ROBY value.

\subsection{ROBY Evaluates the Robustness of Varying Model Capacity}
Now we turn to \textbf{Q3} about the robustness of models with varying architecture.  We analyze the relationship between ROBY metric and adversarial robustness on models with different complexity and model capacity. 

For the model capacity, we start with a simple, fully connected network and study how its robustness, as manifested by ROBY and ASR, changes against adversarial samples as we keep increasing the network's size by adding the number of neurons and the size of the fully connected layer. We used a simple convolutional network for the model type and study how robustness, as manifested by ROBY and ASR changes against adversarial samples as we change the model structure with different layers. We conduct experiments on MNIST data set by comparing the ROBY metric and related $l_\infty$-norm PGD attack success rate ASR.

\begin{figure*}[ht]
\centering
\subfigure[Robustness of models with different numbers of neurons]{
\includegraphics[width=0.48\textwidth]{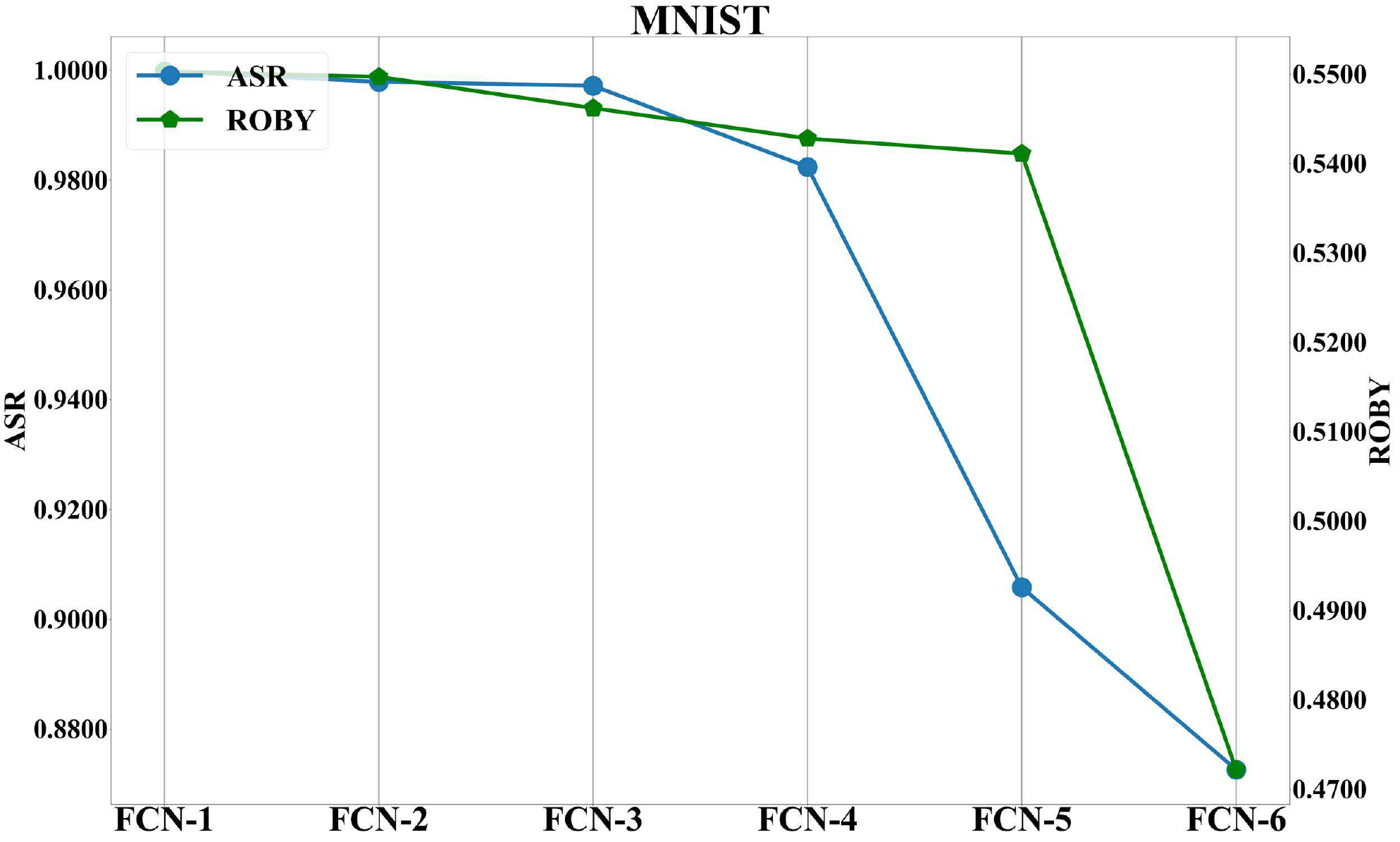} 
}
\subfigure[Robustness of models with different numbers of layers]{
\includegraphics[width=0.48\textwidth]{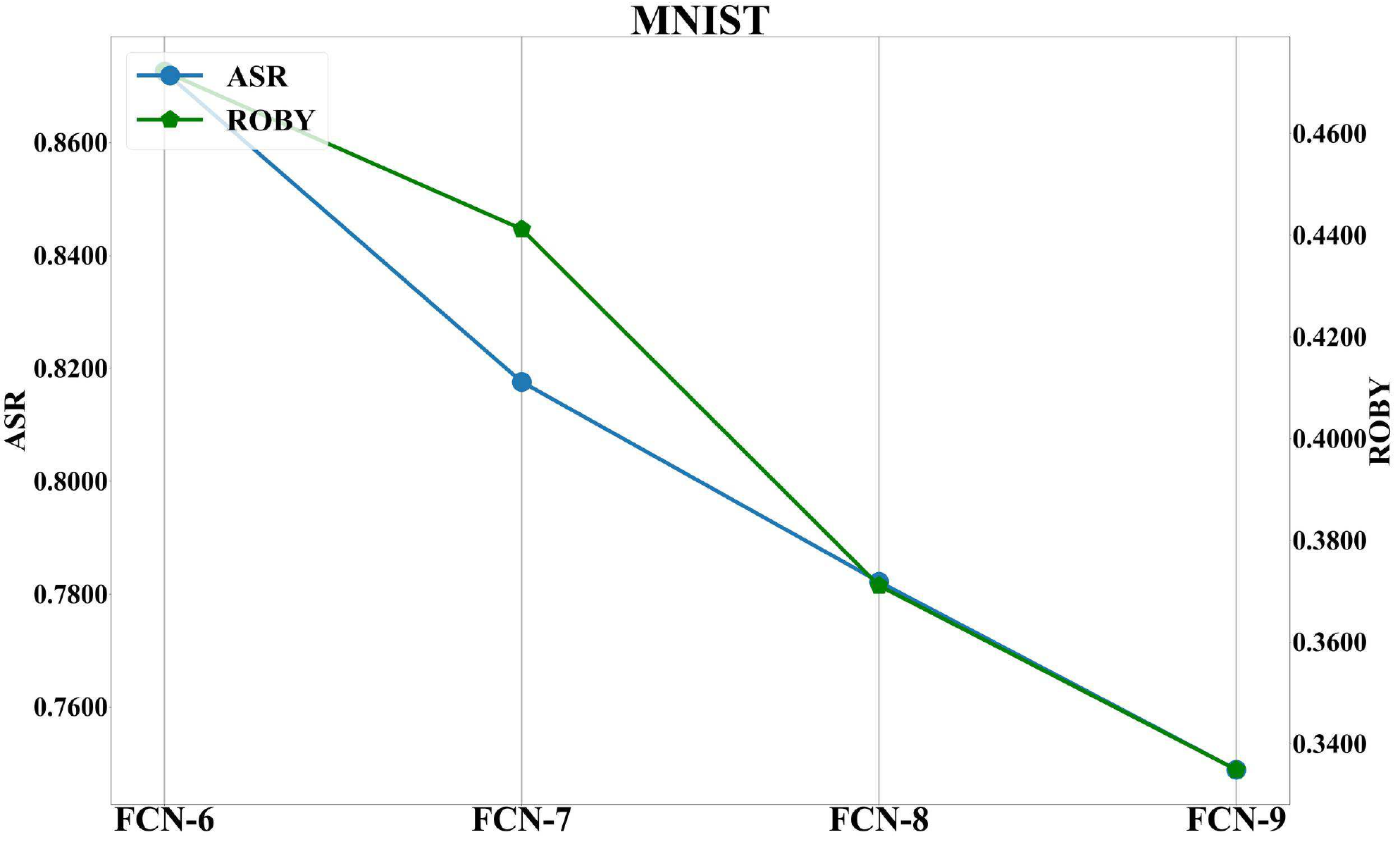}
}
\caption{Robustness evaluation of models with different model size.}
\label{size}
\end{figure*}

TABLE~\ref{neurons} shows the adversarial robustness of models with different numbers of neurons.
\begin{table}[htbp]
\centering
\caption{Robustness Evaluation of Models with Different Numbers of Neurons}
\setlength{\tabcolsep}{1mm}{
\scriptsize 
\begin{tabular}{ccccccc}
    \hline\hline
    \textbf{Data Set} & \textbf{Model} & \textbf{Layer} & \textbf{Neurons } & \textbf{ACC} & \textbf{ASR} & \textbf{ROBY} \\
    \hline\hline
      & \textbf{FCN-1} & \textbf{2} & \textbf{100} & \textbf{0.9790 } & \textbf{0.9998 } &   \textbf{0.5503 } \\
      & \textbf{FCN-2} & \textbf{2} & \textbf{500} & \textbf{0.9821 } & \textbf{0.9979 } &  \textbf{0.5497 } \\
      & \textbf{FCN-3} & \textbf{2} & \textbf{1000} & \textbf{0.9797 } & \textbf{0.9972 } &   \textbf{0.5462 } \\
    \textbf{MNIST} & \textbf{FCN-4} & \textbf{2} & \textbf{2000} & \textbf{0.9761 } & \textbf{0.9824 }  &  \textbf{0.5428 } \\
      & \textbf{FCN-5} & \textbf{2} & \textbf{3000} & \textbf{0.9758 } & \textbf{0.9058 }  &  \textbf{0.5411 } \\
      & \textbf{FCN-6} & \textbf{2} & \textbf{4000} & \textbf{0.9734 } & \textbf{0.8726 }  &  \textbf{0.4722 } \\
    \hline\hline
\end{tabular}}%
  \label{neurons}%
\end{table}%
Each deep learning model is composed of a fully connected network. Each model has the same number of layers, and each layer has a different number of neurons. After training, each network’s classification success rate reaches approximately 97\%. On this basis, the robustness and ROBY metric of each model is calculated and analyzed. The visualization results are shown in Fig.~\ref{size}. The ROBY metric of the deep model negatively correlates with the number of model neurons and matches model robustness. The more the number of neurons, the smaller the ASR, the smaller the ROBY value, the safer the model. 

\begin{table}[htbp]
  \centering
  \caption{Robustness Evaluation of Models with Different Numbers of Layers}
\setlength{\tabcolsep}{1mm}{
\scriptsize 
    \begin{tabular}{ccccccc}
    \hline\hline
    \textbf{Data Set} & \textbf{Model} & \textbf{Layer} & \textbf{Neurons} & \textbf{ACC} & \textbf{ASR} & \textbf{ROBY} \\
    \hline\hline
      & \textbf{FCN-6} & \textbf{2} & \textbf{4000} & \textbf{0.9734 } & \textbf{0.8726 }  &  \textbf{0.4722 } \\
    \textbf{MNIST} & \textbf{FCN-7} & \textbf{3} & \textbf{4000} & \textbf{0.9796 } & \textbf{0.8176 }  &  \textbf{0.4412 } \\
      & \textbf{FCN-8} & \textbf{4} & \textbf{4000} & \textbf{0.9627 } & \textbf{0.7822 } &   \textbf{0.3711 } \\
      & \textbf{FCN-9} & \textbf{5} & \textbf{4000} & \textbf{0.9800 } & \textbf{0.7489 }  &  \textbf{0.3349 } \\
    \hline\hline
    \end{tabular}}%
  \label{lay}%
\end{table}%
TABLE~\ref{lay} shows the adversarial robustness of models with different numbers of layers.
Each deep learning model comprises of a fully connected network with the same number of neurons in each layer and a different number of layers. After training, each network's classification accuracy (ACC) reaches approximately 96\%. On this basis, the robustness and ROBY metric of each model is calculated and analyzed. The deep model's ROBY metric and the number of model layers are negatively correlated and match the model robustness in ASR. The more the number of layers, the smaller the ASR, the smaller the ROBY value, the safer the model. The trend is visualized in Fig.~\ref{size}. Both ASR and ROBY drop as the number of layers increase, meaning that the models' robustness is well captured.

We design a simple convolutional neural network for the model type, including convolutional layers, pooling layers, fully connected layers. We then study how its behavior changes against adversarial samples as we change the model structure. TABLE~\ref{structure} shows the structure of different models. TABLE~\ref{robustness_type} shows the adversarial robustness of models with varying types of model.

\begin{table}[htbp]
  \centering
  \caption{Model Structure of Simple Convolutional Neural Networks}
    \begin{tabular}{ccccc}
    \hline\hline
    \textbf{Model} & \textbf{Conv} & \textbf{Pool} & \textbf{Dropout} & \textbf{FC} \\
    \hline\hline
    \textbf{CNN-1} & \textbf{\checkmark} & \textbf{\checkmark} & \textbf{\checkmark} & \textbf{\checkmark} \\
    \textbf{CNN-2} & \textbf{\checkmark} & \textbf{—} & \textbf{\checkmark} & \textbf{\checkmark} \\
    \textbf{CNN-3} & \textbf{\checkmark} & \textbf{\checkmark} & \textbf{—} & \textbf{\checkmark} \\
    \textbf{CNN-4} & \textbf{\checkmark} & \textbf{—} & \textbf{—} & \textbf{\checkmark} \\
    \hline\hline
    \end{tabular}%
  \label{structure}%
\end{table}%

\begin{table}[htbp]
  \centering
  \caption{Robustness Evaluation of Models with Different Model Types}
\setlength{\tabcolsep}{1mm}{
\scriptsize 
    \begin{tabular}{ccccc}
    \hline\hline
    \textbf{Dataset} & \textbf{Model} & \textbf{ACC} &  \textbf{ASR} & \textbf{ROBY} \\
    \hline\hline
      & \textbf{CNN-3} & \textbf{0.5739 } & \textbf{0.9398 } &  \textbf{0.4113 } \\
    \textbf{CIFAR-10} & \textbf{CNN-1} & \textbf{0.6151 } & \textbf{0.9432 }  &  \textbf{0.4581 } \\
      & \textbf{CNN-2} & \textbf{0.6040 } & \textbf{0.9437 }  & \textbf{0.4879 } \\
      & \textbf{CNN-4} & \textbf{0.5568 } & \textbf{0.9602 } & \textbf{0.5674 } \\
    \hline
      & \textbf{CNN-3} & \textbf{0.9918 } & \textbf{0.7832 }  & \textbf{0.5004 } \\
    \textbf{MNIST} & \textbf{CNN-1} & \textbf{0.9929 } & \textbf{0.9250 }  & \textbf{0.5735 } \\
      & \textbf{CNN-4} & \textbf{0.9892 } & \textbf{0.9670 }  & \textbf{0.5857 } \\
      & \textbf{CNN-2} & \textbf{0.9705 } & \textbf{0.9693 }  & \textbf{0.5955 } \\
    \hline
      & \textbf{CNN-3} & \textbf{0.9178 } & \textbf{0.9388 } & \textbf{0.4807 } \\
    \textbf{Fashion-MNIST} & \textbf{CNN-1} & \textbf{0.9105 } & \textbf{0.9670 } &  \textbf{0.4955 } \\
      & \textbf{CNN-4} & \textbf{0.9095 } & \textbf{0.9810 } & \textbf{0.5433 } \\
      & \textbf{CNN-2} & \textbf{0.9176 } & \textbf{0.9934 } & \textbf{0.5575 } \\
    \hline\hline
    \end{tabular}}%
  \label{robustness_type}%
\end{table}%

\begin{figure*}[ht]
\centering
\subfigure[]{
\includegraphics[width=0.3\textwidth]{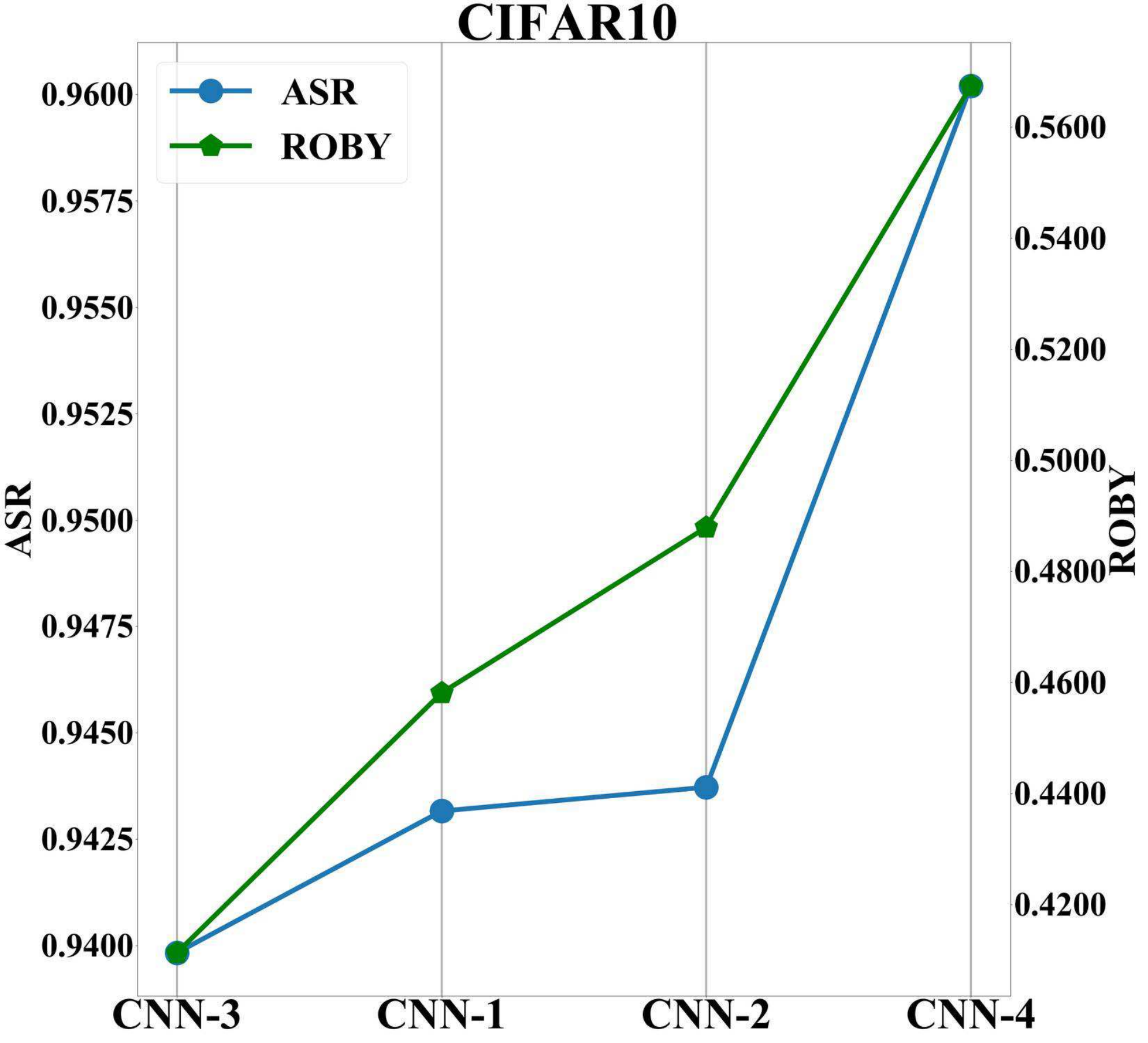} 
}
\subfigure[]{
\includegraphics[width=0.3\textwidth]{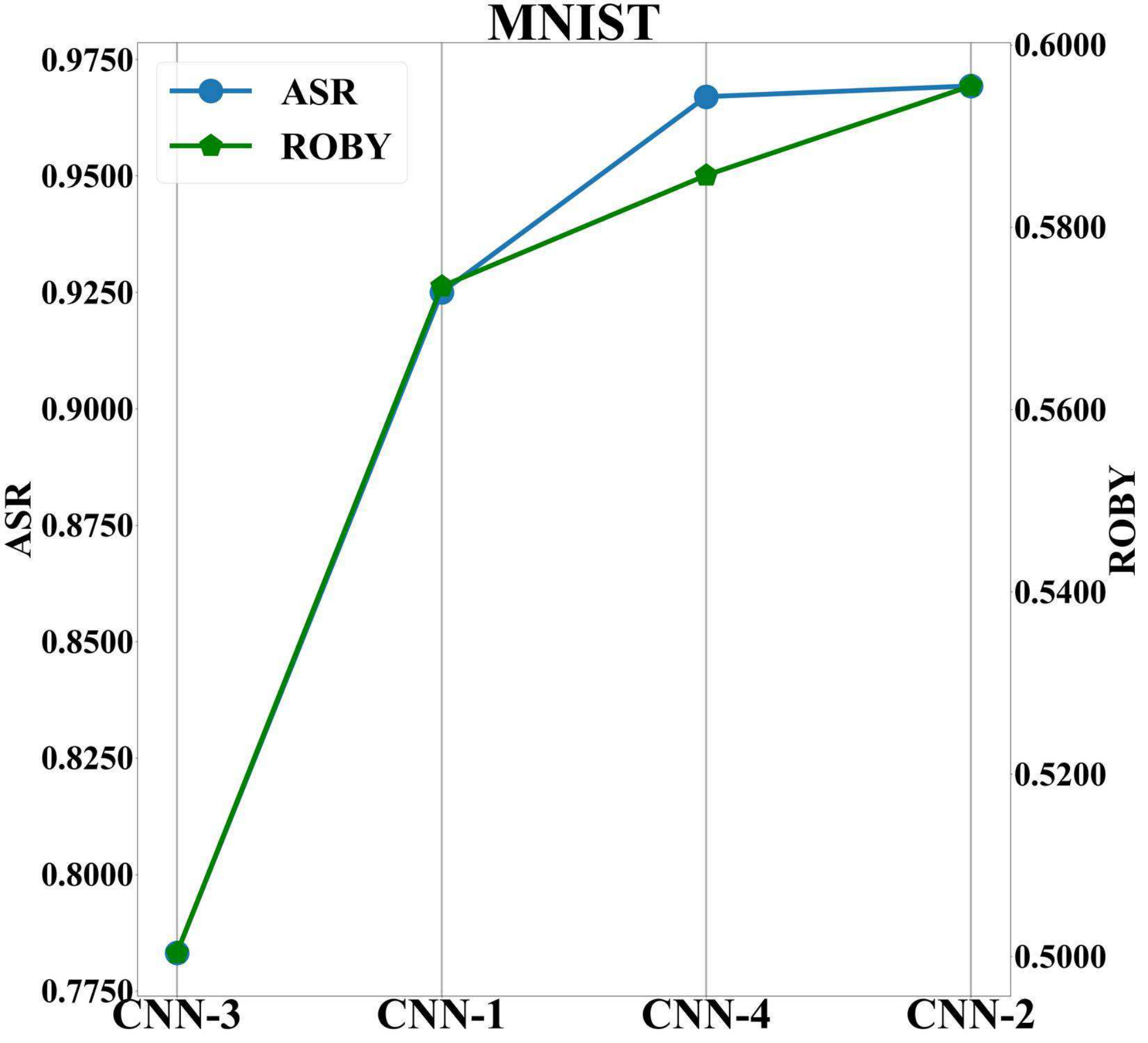}
}
\subfigure[]{
\includegraphics[width=0.3\textwidth]{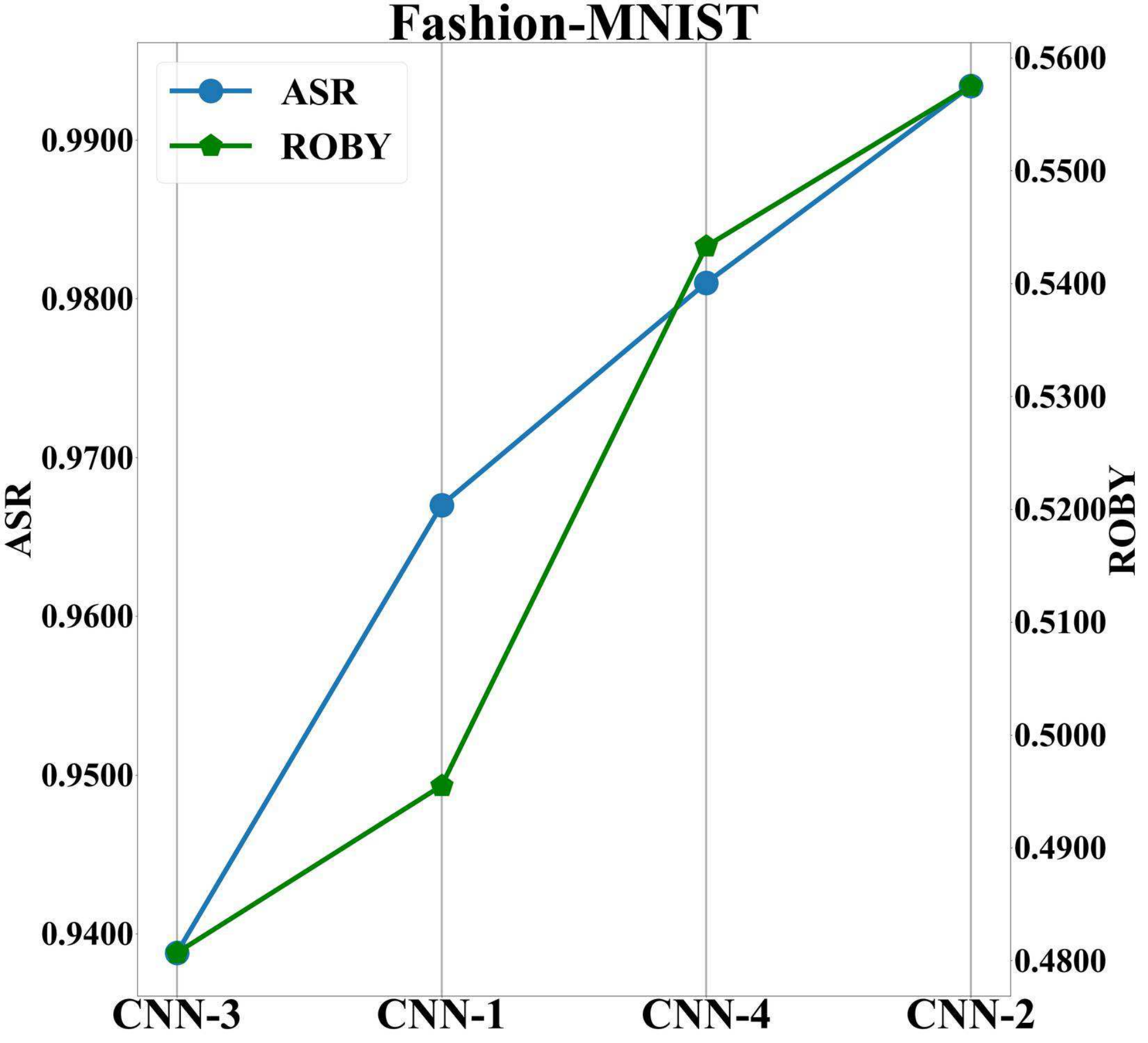}
}
\caption{Robustness evaluation of models with different model types.}
\label{fig_type}
\end{figure*}

After training, each network's classification accuracy reaches a similar level. Calculating and analyzing the ASR and ROBY metric of each model shows that deep learning model's ROBY metric has a linear relationship with the model layer type and matches the model robustness. The visualization results are shown in Fig.~\ref{fig_type}. The lower the ASR, the smaller the ROBY value, and the safer the model. And the three data sets all show similar model robustness comparison results. The model without the dropout layer has better robustness, as the model has more neurons and a more complicated structure.

Relating to the adversarial perspective, for a fixed set of possible perturbations, the model robustness is mainly dependent on the architecture of the classifier. Consequently, the architectural capacity of the model becomes a significant factor affecting its overall performance. At a higher level, classifying adversarial examples requires a more robust classifier, which changes the decision boundary of the resulting model to a more complicated one. Generally, the adversarial robustness requires a significantly larger architectural capacity of the networks. 

Overall, we find from the empirical results that the ROBY metric matches the model robustness on models with different complexity and model capacity.

\subsection{Cost Analysis of Robustness Evaluation}
The ROBY metrics is attack-independent: it can be calculated with only neural network models and the original samples without the adversarial samples. It saves significant cost incurred from generating adversarial samples than attack-based evaluation method.

TABLE~\ref{cc} shows the computational cost of ROBY metrics and PGD attack based ASR on different data sets. The time shown in the table is calculated by averaging the running time of the evaluation process of 10 neural network models on a dataset.

\begin{table}[htbp]
  \centering
  \caption{Comparison of the Cost of ROBY and the Adversarial Attack based Robustness Evaluation}
    \begin{tabular}{ccc}
    \hline\hline
    \textbf{Data Set} & \textbf{ROBY metrics} & \textbf{PGD Attack} \\
    \hline
    \textbf{CIFAR-10} & \textbf{202ms} & \textbf{31481ms} \\
    \hline
    \textbf{MNIST} & \textbf{189ms} & \textbf{16001ms} \\
    \hline
    \textbf{Fashion-MNIST} & \textbf{195ms} & \textbf{16968ms} \\
    \hline\hline
    \end{tabular}%
  \label{cc}%
\end{table}%

The computational cost of ROBY metrics is significantly lower than the PGD attack. It takes less than a second to evaluate a model's robustness, which is practical in real-world applications.

\section{Conclusion}
In this paper, we propose ROBY, a novel generic attack-independent robustness measure based on the models' decision boundary. ROBY depicts the decision boundaries by the inter-class and intra-class statistic features and evaluates the robustness of target neural network classifiers without adversarial examples. Our extensive exper¬iments show that the ROBY robustness evaluation metric matches the attack-based robustness indicator ASR on a wide range of natural and defended networks. ROBY applies to a wide range of state-of-the-art neural network classifiers and requires little computation cost compared to the existing robustness evaluation approaches.

To improve the robustness of deep models, we would suggest actions along two lines: to include adversarial samples in model training and to design larger architectural capacity and higher complexity, including neuron number, layer size, and model structure, when there is sufficient training data.

%

 \section*{Acknowledgments}
This research was supported by the National Natural Science Foundation of China under Grant No. 62072406, the Natural Science Foundation of Zhejiang Provincial under Grant No. LY19F020025, the Major Special Funding for ``Science and Technology Innovation 2025'' in Ningbo under Grant No. 2018B10063,
the Key Laboratory of the Public Security Ministry Open Project in 2020 under Grant No. 2020DSJSYS001.

\ifCLASSOPTIONcaptionsoff
  \newpage
\fi



%

\bibliographystyle{IEEEtran}
\bibliography{ref} 

%
\begin{IEEEbiography}[{\includegraphics[width=1in,height=1.2in,clip,keepaspectratio]{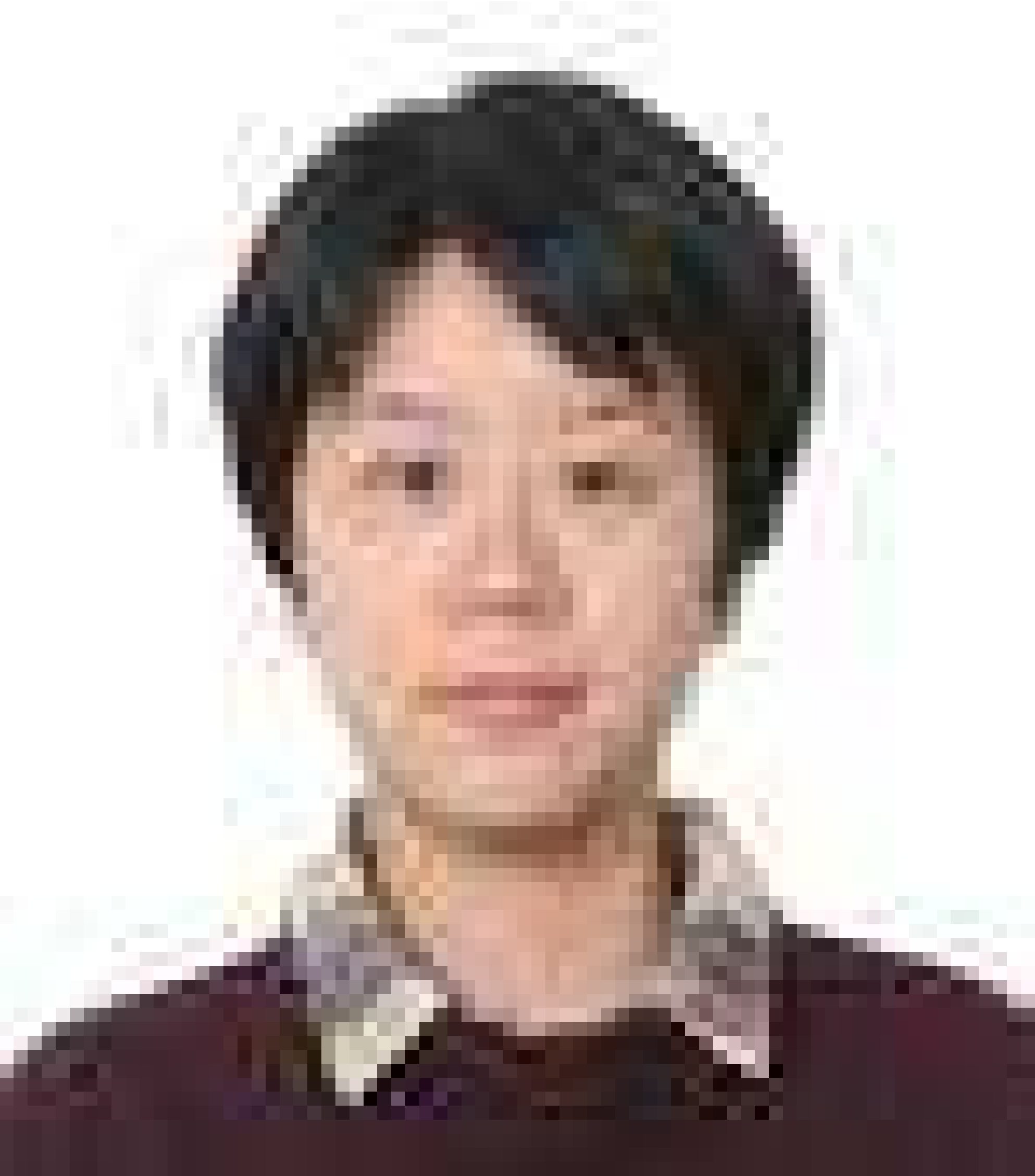}}]{Jinyin Chen}
Jinyin Chen received BS and Ph.D. degrees from Zhejiang University of Technology, Hangzhou, China, in 2004 and 2009, respectively. She studied evolutionary computing in Ashikaga Institute of Technology, Japan in 2005 and 2006. She is currently an Associate Professor with the Zhejiang University of Technology, Hangzhou, China. Her research interests include artificial intelligence security, graph data mining, and evolutionary computing. 
\end{IEEEbiography}

\begin{IEEEbiography}[{\includegraphics[width=1in,height=1.2in,clip,keepaspectratio]{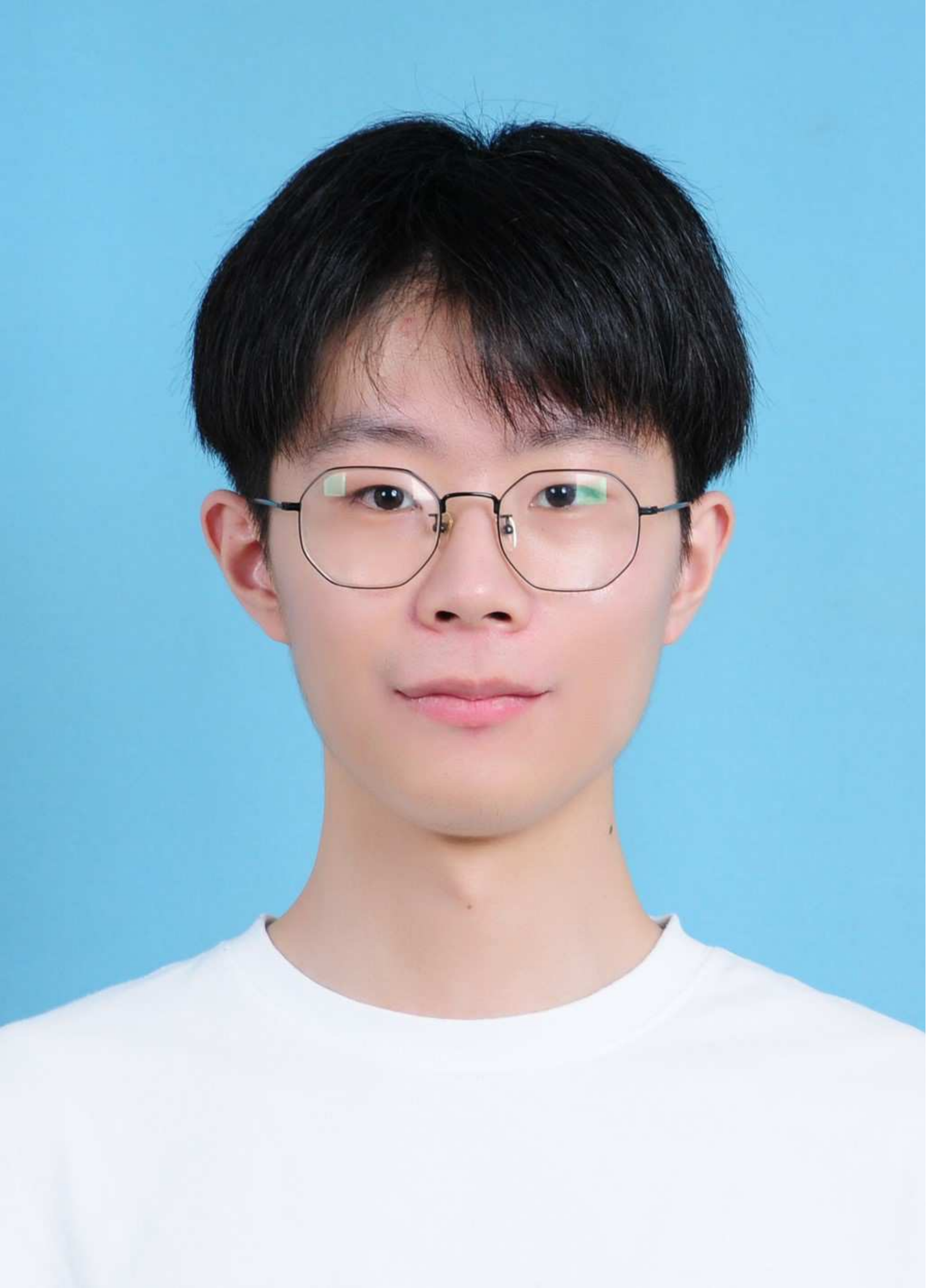}}]{Zhen Wang}
Zhen Wang received BS degrees in electronic information engineering from Zhejiang University of Technology, Hangzhou, China, in 2020. He has been working toward the MS degree in software engineering at Zhejiang Univeristy, Hangzhou, China, in 2020. His research interests include machine learning and artificial intelligence security.
\end{IEEEbiography}

\begin{IEEEbiography}[{\includegraphics[width=1in,height=1.2in,clip,keepaspectratio]{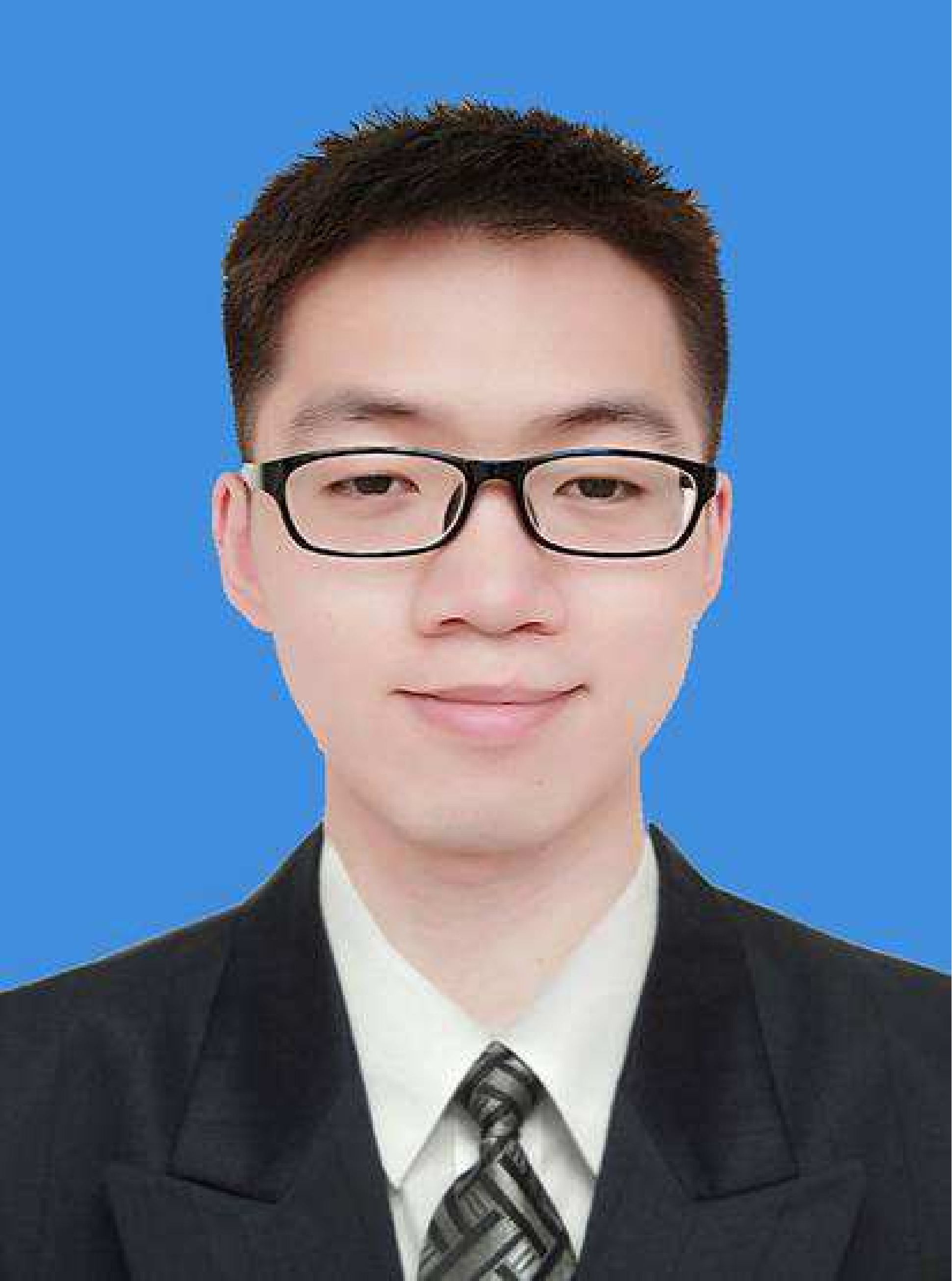}}]{Haibin Zheng}
Haibin Zheng is a PhD student at the college of Information Engineering, Zhejiang University of Technology. He received his bachelor degree from Zhejiang University of Technology in 2017. His research interests include deep learning, artificial intelligence, and adversarial attack and defense.
\end{IEEEbiography}

\begin{IEEEbiography}[{\includegraphics[width=1in,height=1.2in,clip,keepaspectratio]{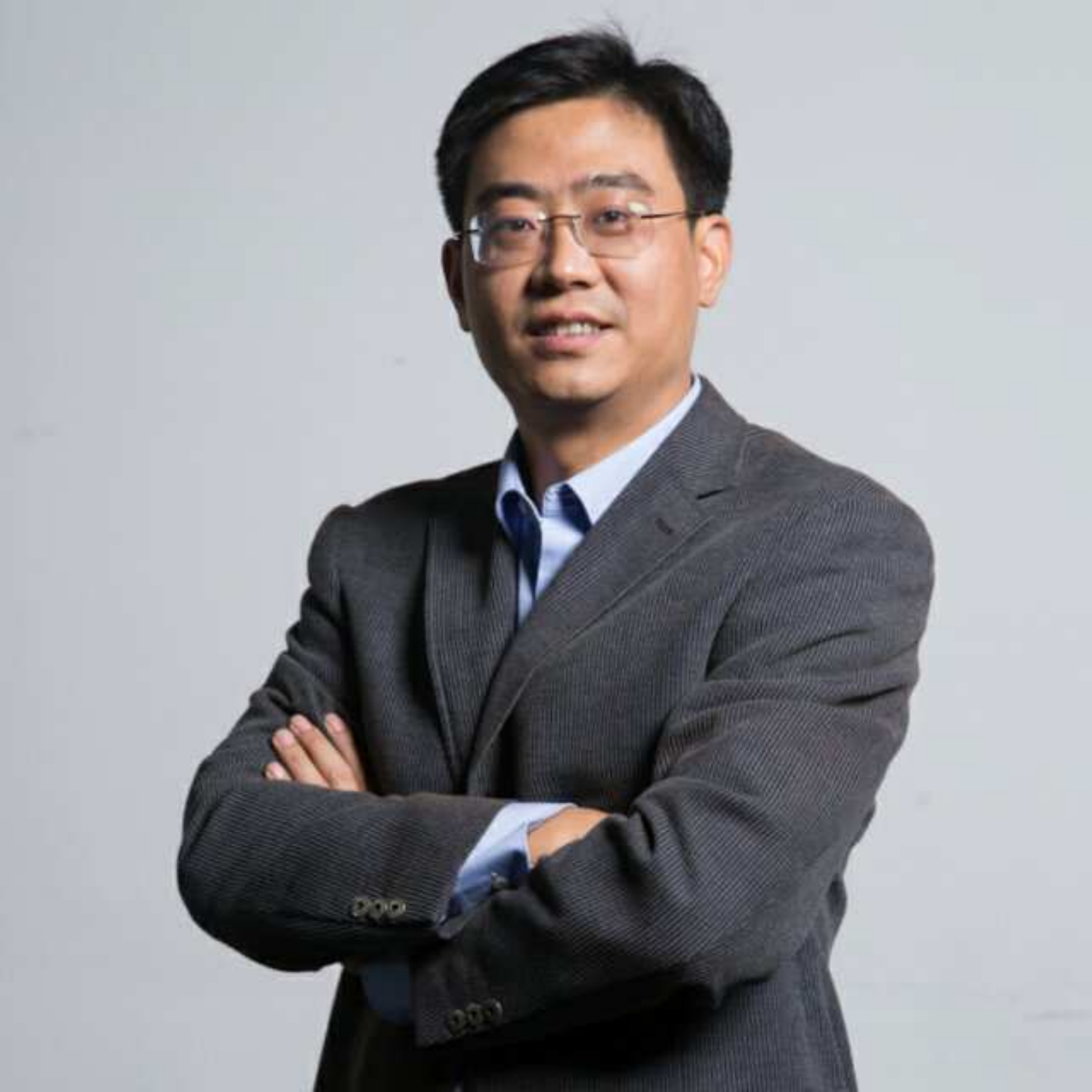}}]{Jun Xiao}
Jun Xiao received the Ph.D. degree in computer science and technology
from the College of Computer Science, Zhejiang University, Hangzhou,
China, in 2007. He is currently a professor with the College of Computer
Science, Zhejiang University. His current research interests include computer
vision, multimedia retrieval, and machine learning.
\end{IEEEbiography}


\begin{IEEEbiography}[{\includegraphics[width=1in,height=1.2in,clip,keepaspectratio]{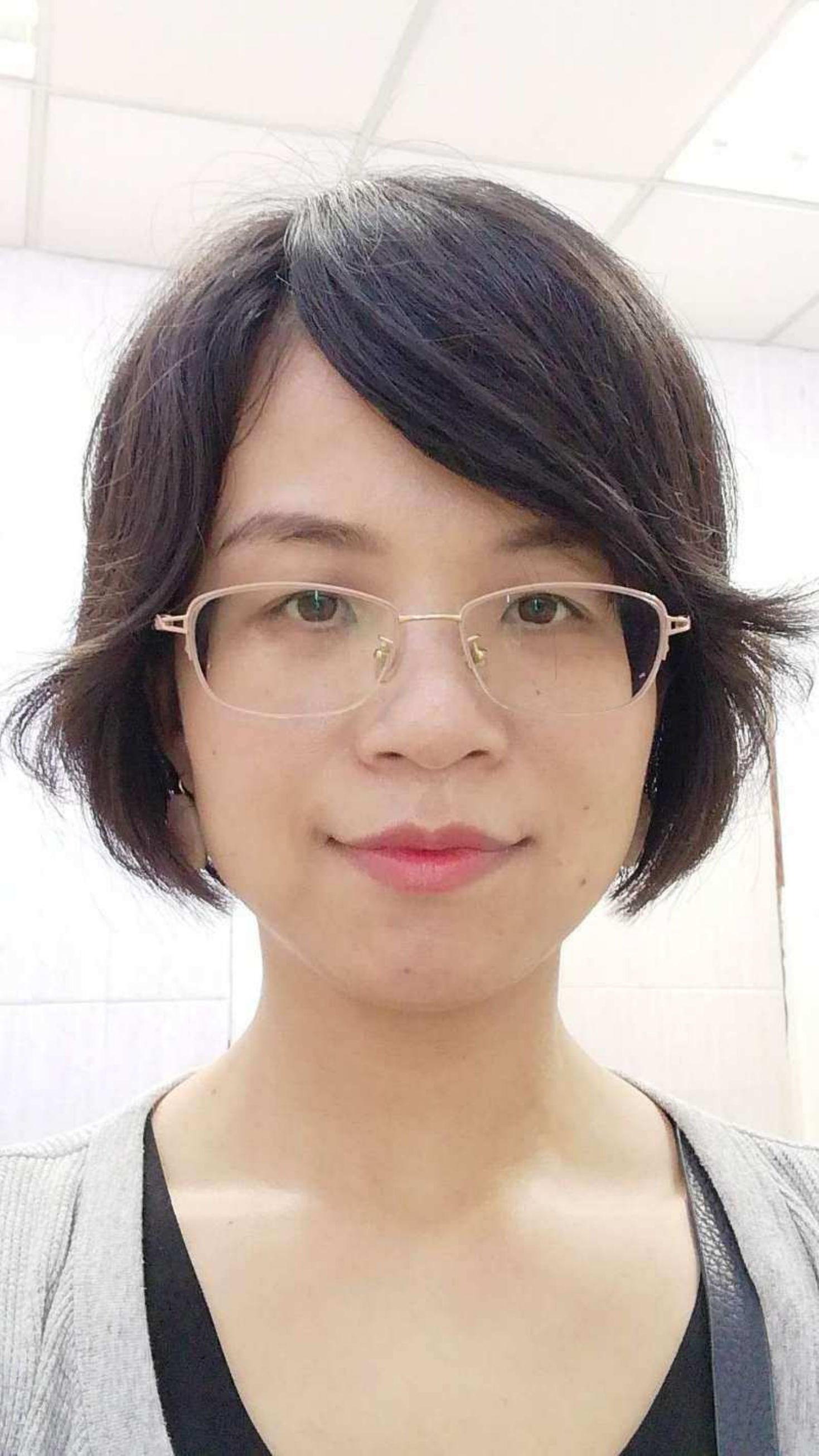}}]{Zhaoyan Ming}
Zhaoyan Ming is a researcher at the Institute of Innovative Computing, Zhejiang University.  Her expertise lies in the intersection of Multimedia Knowledge Graph and AI.  Dr. Ming holds a Bachelor degree in Information Science \& Electronic Engineering at Zhejiang University and a Ph.D. degree in Computer Science at National University of Singapore.
\end{IEEEbiography}

\end{document}